\documentclass[journal]{IEEEtran}

\ifCLASSINFOpdf
\else
\fi

\usepackage{graphics} 
\usepackage{epsfig} 
\usepackage{mathptmx} 
\usepackage{times} 
\usepackage{amsmath} 
\usepackage{amssymb}  
\usepackage{graphicx}
\usepackage{hyperref}
\usepackage{makecell}
\usepackage{graphics,graphicx,amssymb,amsmath,verbatim}
\usepackage{mathrsfs}
\usepackage{amsfonts}
\usepackage{epstopdf}
\usepackage{xcolor}
\usepackage{color}
\usepackage{subfigure}
\usepackage{hyperref}
\usepackage{booktabs}
\usepackage{threeparttable}

\newtheorem{theorem}{Theorem}

\usepackage{algorithm}
\usepackage{algorithmic}
\usepackage{multirow}
\input{tcilatex}

\hypersetup{
colorlinks=true,
linkcolor=black,
citecolor=black,
urlcolor=black
}
\newenvironment{proof of Proposition 1}[1][Proof of Proposition 1]{\textbf{\noindent{\textit{#1.}}} }{\ \rule{0.5em}{0.5em}}

\newenvironment{Proof}[1][Proof]{{\textit{#1.}} }{\ \rule{0.5em}{0.5em}}
\newenvironment{Remark 1}[1][Remark 1]{{\textit{#1.}} }{\ }
\newenvironment{proof of Theorem 2}[1][Proof of Theorem 2]{\textbf{\noindent{\textit{#1.}}} }{\ \rule{0.5em}{0.5em}}
\newenvironment{proof of Theorem 3}[1][Proof of Theorem 3]{\textbf{\noindent{\textit{#1.}}} }{\ \rule{0.5em}{0.5em}}
\newenvironment{proof of Theorem 4}[1][Proof of
Theorem 4]{\textbf{\noindent{\textit{#1.}}} }{\ \rule{0.5em}{0.5em}}
\newenvironment{proof of Theorem 5}[1][Proof of Theorem 5]{\textbf{\noindent{\textit{#1.}}} }{\ \rule{0.5em}{0.5em}}
\newenvironment{evaluation metrics}[1][Evaluation Metrics]{\textbf{\noindent{\textit{#1.}}} }{}
\title{\LARGE \bf
 SLAM-based Joint Calibration of Multiple Asynchronous Microphone Arrays and Sound Source Localization
}

\author{Jiang Wang, Yuanzheng He, Daobilige Su, Katsutoshi Itoyama, Kazuhiro Nakadai, Junfeng Wu, Shoudong Huang, \\
	Youfu Li, and He Kong
  \thanks{This paper was accepted to and going to appear in \textit{the IEEE Transactions on Robotics}.}
		\thanks{Jiang Wang, Yuanzheng He, and He Kong (corresponding author) are with the Shenzhen Key Laboratory of Control Theory and Intelligent Systems, Southern University of Science and Technology, No. 1088 Xueyuan Avenue, Shenzhen, China; Email: 12132297@mail.sustech.edu.cn; 12132259@mail.sustech.edu.cn; kongh@sustech.edu.cn. Daobilige Su is with the College of Engineering, China Agricultural University, Beijing, China; Email: sudao@cau.edu.cn. Katsutoshi Itoyama and Kazuhiro Nakadai are with the Department of Systems and Control Engineering, Tokyo Institute of
Technology, Tokyo, Japan; Email: {itoyama;nakadai}@ra.sc.e.titech.ac.jp. Junfeng Wu is with the School of Data Science,  The Chinese University of Hong Kong, Shenzhen, Shenzhen, China; Email: junfengwu@cuhk.edu.cn. Shoudong Huang is with  the Robotics Institute,
University of Technology Sydney, Sydney, Australia; Email: shoudong.huang@uts.edu.au. Youfu Li is with the Department of Mechanical Engineering, City University of Hong Kong, Hong Kong SAR, China; Email: meyfli@cityu.edu.hk.
		}
	}

\begin{document}

\maketitle

\begin{abstract}
Robot audition systems with multiple microphone arrays have many applications in practice. However, accurate calibration of multiple microphone arrays remains challenging because there are many unknown parameters to be identified, including {the relative transforms (i.e., orientation, translation) and asynchronous factors (i.e., initial time offset and sampling clock difference) between microphone arrays.} To tackle these challenges, in this paper, we adopt batch simultaneous localization and mapping (SLAM) for joint calibration of multiple asynchronous microphone arrays and sound source localization. Using the Fisher information matrix (FIM) approach, we first conduct the observability analysis (i.e., parameter identifiability) of the above-mentioned calibration problem and establish necessary/sufficient conditions under which the FIM and the Jacobian matrix have full column rank, which implies the identifiability of the unknown parameters. We also discover several scenarios where the unknown parameters are not uniquely identifiable. Subsequently, we propose an effective framework to initialize the unknown parameters, which is used as the initial guess in batch SLAM for multiple microphone arrays calibration,  aiming to further enhance optimization accuracy and convergence.
 Extensive numerical simulations and real experiments have been conducted to verify the performance of the proposed method. The experiment results show that the proposed pipeline achieves higher accuracy with fast convergence in comparison to methods that use the noise-corrupted ground truth of the unknown parameters as the initial guess in the optimization and other existing frameworks. 

\end{abstract}


\section{INTRODUCTION}

Microphone array-based robotic auditory systems have many applications such as sound source localization and human-robot interaction \cite{Grondin2022}-\cite{AnTRO}. As with other sensing modalities
\cite{ZHANG2022}-\cite{Jiao2019}, precise calibration of robotic auditory system parameters is crucial for achieving satisfactory sound source localization and tracking performance \cite{Plinge2016}. Hence, the calibration of robotic auditory systems made of single or multiple microphone arrays has received significant attention recently. 

Of particular interest in this paper is the parameter calibration of robotic auditory systems that are made of multiple microphone arrays. Compared to single microphone array-based audition systems, there are more parameters to be calibrated for systems with multiple microphone arrays, including the {relative transforms (i.e., orientation, translation)} and the asynchronous offsets among the arrays. In the following, we first give a brief overview of the relevant literature on calibration of single microphone array-based systems, and then discuss the existing calibration methods for systems with multiple arrays.

\subsection{{Related Work}}
In \cite{Perrodin 2012}, based on the time difference of arrival (TDOA) between each pair of microphones, a calibration algorithm was developed to estimate the positions of microphones within a single microphone array. In \cite{Crocco 2011}, a bilinear calibration method based on time of flight (TOF) between each sensor source pair was proposed to estimate the microphone and source positions in 3D under the condition that the transmitting time is known. In \cite{Kuang 2013}, based on time of arrival (TOA) measurements and assuming knowledge of the distances between the sources and the microphones, a method for joint calibration of the positions of multiple microphones and sound source localization was proposed. In \cite{Burgess 2015}, a calibration method using TOA measurements was proposed for the scenario with a planar microphone array and a sound source moving in 3D.

Note that the applicability of the above-mentioned methods is limited in that they all rely on hardware synchronization between microphone channels,
which is challenging to implement for robotic platforms in practice due to spatial and cost constraints \cite{Plinge2016}. Recently, in \cite{Su2015}-\cite{Kong2021},
a general framework using batch simultaneous localization and mapping has been developed for joint sound source localization and
calibration of a single microphone array with asynchronous effects
(i.e., clock difference and initial time offset).

Compared to single microphone array-based systems, the calibration of systems with multiple arrays has gained more recent attention. For example, the proposed approach in \cite{Plinge2014} utilizes direction of arrival (DOA) measurements to determine the sound source location and inter-array TDOA measurements to obtain the microphone array location through exhaustive grid search.  The work \cite{Plinge2017} employs evolutionary algorithms to improve the accuracy and real-time performance of the approach in \cite{Plinge2014}. Based on DOA and inter-array TDOA measurements, another calibration framework for multiple microphone arrays
is proposed in \cite{Yin 2021} using distributed damped Newton optimization. Note that the above-mentioned methods focus on the 2D case. 

For the more general 3D case, there are only a few existing works. In \cite{Wang 2021}, an artificial bee colony algorithm was employed to calibrate the positions and orientation of microphone arrays in 3D. Nevertheless, this method assumes that the sound source position at different moments is partially known and the clocks of the arrays are synchronized using hardware. Simultaneous calibration
of positions, orientations, and time offsets of multiple microphone arrays
and sound source positions in 3D was explored in \cite{Wozniak 2019} and \cite{Nakadai2023}. 
\subsection{{Motivation}}
For spatially distributed microphone arrays, it is necessary
to consider both the initial time offsets and the sampling clock differences
between the arrays \cite{Wang2016}, especially in the case of asynchronized scenarios based on the USB protocol and wireless acoustic sensor networks. In the above situations, each microphone array captures acoustic signals through its own microprocessor-controlled analog-to-digital converter and has a unique sampling clock source. Therefore, when launching multiple microphone arrays, differences in initialization result in varying initial time offsets between arrays. Moreover, the microprocessors in these microphone arrays often have limited performance, and the oscillators/crystals used to generate clock signals typically drift around their nominal frequencies. As a result, differences in sampling clocks accumulate over time. Not properly handling 
the above issue will significantly degrade the performance of sound source localization/tracking algorithms embedded in the arrays \cite{Su2020}.

\begin{figure}[t]
\centering
\includegraphics[width=0.8\columnwidth]{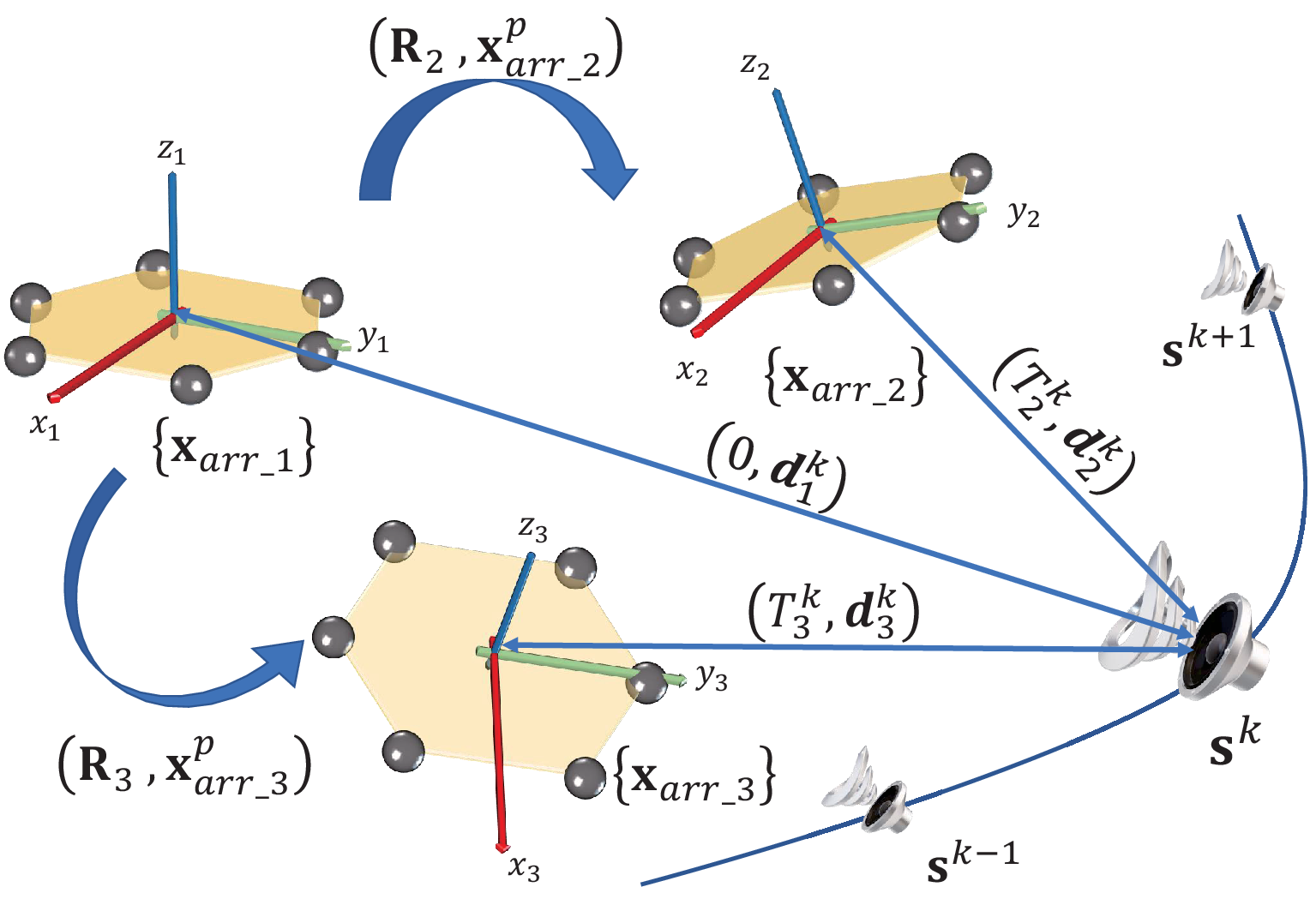}
\caption{Geometry of the problem setup and batch SLAM-based framework for
        multiple microphone arrays calibration and sound source localization.}
\label{TDOA}
\end{figure}

To the best of our knowledge, there is no work that has addressed the simultaneous calibration of positions, orientations,
time offsets and sampling clock differences of multiple microphone arrays and sound source
positions in 3D. In fact, as for single microphone array, calibration of multiple
microphone arrays can be considered as a SLAM problem \cite{Thrun2006}-\cite{Moradi2021}, where microphone
arrays and the moving sound source serve as landmarks in the environment and the robot, respectively. As illustrated in Fig. \ref{TDOA}, the acoustic measurements from the microphone arrays and the motion measurements from the robot are utilized in the optimization process, with landmark-robot constraints and robot relative pose constraints enforced, similar to the approach used in full information estimation and batch SLAM
\cite{Thrun2006}-\cite{Kong2018b}. Then, two important questions arise. 

{Firstly,} it is critical to assess whether the information
contained in the measurements is sufficient to estimate the unknown
parameters of microphone arrays and sound source locations. This is the so-called observability
problem in the SLAM literature \cite{Dissanayake2008}-\cite{Huang2016}.
Although there exist works on observability analysis of SLAM-based calibration of single microphone arrays, in-depth analysis for the case with multiple microphone arrays is lacking.

Secondly, the selection of initial values is crucial because the considered calibration is a nonlinear least squares (NLS)
problem, similar to batch SLAM \cite{Thrun2006}, \cite{Nasiri2018}. Many existing algorithms for solving such NLS problems employ the Gauss-Newton method or its variants. These methods typically require reasonable initial guesses; otherwise, the algorithms may converge toward local minima, or in extreme cases, diverge. For some specific problems, novel algorithms with certifiable convergence properties have been proposed in \cite{Carlone2019}-\cite{Carlone2023}.

\subsection{{Contributions}}
Motivated by the above observations, in this paper, we adopt batch SLAM as a general framework for the simultaneous calibration of translations, orientations, time offsets and sampling clock differences of multiple microphone arrays, and sound source positions in 3D.  Our contributions are two-fold. 

Firstly, we concentrate on the parameter identifiability
of the corresponding SLAM problem. As discussed in existing works \cite{Dissanayake2008}-\cite{Huang2016}, SLAM is not observable from a control theoretical perspective. Hence, in the SLAM literature, the observability problem of SLAM has been tackled from an information-theoretic perspective, where all the parameters to be identified are taken to be constant but unknown. From the information-theoretic perspective, Fisher information quantifies the amount of information contained in a set of observations about a set of unknown parameters \cite{Huang2016}. Following the above line of argument, when the multiple microphone array calibration problem is formulated as an NLS parameter estimation problem, the full rankness of the associated FIM determines the parameter identifiability or observability of the calibration problem.

Hence, in this paper, by leveraging the FIM approach, we thoroughly
investigate the identifiability of the unknown parameters, including
translations, orientations, and asynchronous factors between the microphone
arrays and the sound source positions. We establish necessary/sufficient
conditions under which the FIM and the Jacobian matrix have full column
rank, which implies the identifiability of the unknown parameters. Furthermore, we identify several scenarios where the unknown parameters are not uniquely identifiable.

Secondly, we propose an effective framework to initialize the unknown parameters from the measurements, which is used as the initial guess in batch SLAM. Specifically, the initialization procedure
is composed of the following major steps: (i) estimation of the sound
source position by triangulation; (ii) estimation of distance between
the sound source and microphone arrays using 3D geometry; (iii) estimation
of microphone array poses using the iterative closest point (ICP) method; (iv)
estimation of the asynchronous factors using linear least squares (LLS). As to be explained later in the paper (see Section IV. A), the microphone array pose estimation problem addressed in step (iii) mentioned above is conceptually a point-to-point registration problem, and hence can be tackled effectively using ICP \cite{Pomerleau F}. To validate the effectiveness and robustness of the proposed initialization framework, we have conducted extensive numerical simulations and real experiments. Overall, the proposed pipeline achieves higher accuracy with fast convergence, in comparison to methods that use the noise-corrupted ground truth of the unknown parameters as the initial guess in the optimization, and other state-of-the-art methods in the literature \cite{Plinge2017}, \cite{Wozniak 2019}.

Compared to existing frameworks, the proposed calibration method requires less prior information. More specifically,  the knowledge of the source's position required in \cite{Wang 2021}-\cite{Wozniak 2019},  or the distance between the signal source and the microphones needed in \cite{Kuang 2013} is not required in the proposed framework in this paper. It should also be noted that our previous works documented in \cite{Su2015}-\cite{Kong2021} primarily focused on calibrating individual microphones within a single array while in this paper we address the more challenging problem of calibrating multiple microphone arrays.

Finally, we remark that the observability analysis reported in Section III has been previously reported in our conference paper \cite{SII2023}. However, the results of \cite{SII2023} are only applicable for the case where the time interval between consecutive sound source events is fixed. In the current paper, we generalize the results in \cite{SII2023} from the scenario of fixed-interval sound source emissions to arbitrary time intervals
(i.e., the interval between every two consecutive sound events can be asynchronous and time-varying). 
More importantly, we have proposed an effective framework for estimating the initial values of the parameters and conducted extensive simulation studies and real experiments to validate the entire calibration pipeline. All the codes and 
multimodal dataset used in this paper are publicly available
at \href{https://github.com/AISLAB-sustech/Calibration_of_Multi_Mic_Arrays}{https://github.com/AISLAB-sustech/Calibration\_of\_Multi\_Mic\_Arrays}.


\textbf{Notation}: Denote $x$, $\mathbf{x}$, and $\mathbf{X}$ as
scalars, vectors, and matrices, respectively. $\mathbf{X}^{\mathrm{T}}$ represents the transpose of matrix $\mathbf{X}$. $\mathbf{I}_{n}$ stands for the identity matrix of $n$ dimensions. $\mathbb{R}^{n}$ denotes the $n$-dimensional
Euclidean space. $[a_{1};\cdots;a_{n}]$ denotes $[a_{1}^{\mathrm{T}},\cdots,a_{n}^{\mathrm{T}}]^{\mathrm{T}}$,
where $a_{1},\cdots,a_{n}$ are scalars/vectors/matrices with proper
dimensions. $diag_{n}(\mathbf{A})$ denotes a block diagonal matrix with $\mathbf{A}$ as block diagonal entries for $n$ times; $diag(\mathbf{A},\mathbf{B})$
denotes a block diagonal matrix with $\mathbf{A}$ and $\mathbf{B}$ as
its block diagonal entries; and $\mathbf{0}$ as a matrix
of appropriate dimensions with its all entries as 0. $\mathbf{X}>0$
means that $\mathbf{X}$ is a positive definite matrix. We denote $\left\Vert \mathbf{x}\right\Vert _{\mathbf{P}}^{2}=\mathbf{x}^{%
\mathrm{T}}\mathbf{Px}$. Vectors/matrices, with dimensions not explicitly stated, are assumed to be algebraically compatible.

\section{PROBLEM FORMULATION}
In a calibration
scene containing $N$ microphone arrays, as shown in Fig. \ref{TDOA} (with $N=3$ as an example), the arrays capture $K$ consecutive
acoustic signals emitted by a single sound source at several spatial
positions. $\mathbf{x}_{arr\_i}^{p}$ represents
the position of the $i\raisebox{0mm}{-}th$ microphone array in the global reference frame and any two arrays are in different positions.
We assume that there is a local reference frame $\left\{ \mathrm{\mathbf{x}_{\mathit{arr\_i}}}\right\} $
attached to every microphone array; we choose $\left\{ \mathrm{\mathbf{x}_{\mathit{arr\_\mathrm{1}}}}\right\} $
as the global reference frame; $\mathbf{R}_{i}$ is the rotation matrix
of reference frame $\left\{ \mathrm{\mathbf{x}}_{arr\_1}\right\} $
to the frame $\left\{ \mathrm{\mathbf{x}}_{arr\_i}\right\} $ with
the ZYX Euler angles vector $\mathbf{x}_{arr\_i}^{\theta}$; $\mathbf{s}^{k}$
is the sound source position at time instance $t^{k},$ $k=1,\ldots,K$, with
respect to (w.r.t.) $\left\{ \mathrm{\mathbf{x}_{\mathit{arr\_\mathrm{1}}}}\right\} $,
where $K$ is the total number of time steps. In the calibration
process, the arrays remain static while the sound source moves around.

Here we consider the most general scenario with initial time offset and sampling clock difference among microphone arrays (we assume that the configuration of each microphone array itself, including its geometry, is known). When the sound source sends the $k\raisebox{0mm}{-}th$ acoustic signal,
the DOA information, i.e., the  direction vector of the sound source in the $i\raisebox{0mm}{-}th$ microphone array frame $\left\{ \mathrm{\mathbf{x}}_{arr\_i}\right\} $
is obtained as: 
\begin{equation}
\mathbf{d}_{i}^{k}=\mathbf{R_{\mathit{\mathrm{\mathit{i}}}}^{\mathrm{\mathit{\mathrm{T}}}}}\frac{\mathbf{s}^{k}-\mathbf{x_{\mathit{arr\_i}}^{\mathit{p}}}}{d_{i}^{k}}.\label{expression_DOA}
\end{equation}
Note that the Euclidean norm of $\mathbf{d}_{i}^{k}$ is 1, i.e., $\mathbf{d}_{i}^{k}$ is a unit vector.
Denote $d_{i}^{k}$, for $i=1,2,\ldots{,}N$, as the distance between
the $i\raisebox{0mm}{-}th$ microphone array and the sound source
at the $k\raisebox{0mm}{-}th$ time instant. The inter-array TDOA  information
between the $i\raisebox{0mm}{-}th$ \textsuperscript{}and the first
microphone arrays can be expressed as follows: 
\begin{equation}
    T_{i}^{k}=\frac{d_{i}^{k}}{c}-\frac{d_{1}^{k}}{c}+x_{arr\_i}^{\tau}+{\Delta_{k}}x_{arr\_i}^{\delta}\label{eq:TDOA}
    \end{equation}
for $i=1,2,\ldots{,}N$, where $c$ represents the sound speed in
the air; the scalar (unknown) constant variables $x_{arr\_i}^{\tau}$
and $x_{arr\_i}^{\delta}$ represent the initial time offset and
the sampling clock difference per second of each microphone array, respectively; {
 $\Delta_{k}$ is the time interval from the beginning to the $k\raisebox{0mm}{-}th$ sound signal.} Since the first microphone array is used as the reference,
then 
\begin{equation}
\mathbf{x}_{arr\_1}^{p}=\mathbf{0},\text{ }\mathbf{x}_{arr\_1}^{\theta}=\mathbf{0},\text{ }x_{arr\_1}^{\tau}=0,\text{ }x_{arr\_1}^{\delta}=0.\label{reference}
\end{equation}

The positions and orientation of the $i\raisebox{0mm}{-}th$
microphone array (where $i=2,\ldots{,}N$), i.e., $\mathbf{x}_{arr\_i}^{p}$
and $\mathbf{x}_{arr\_i}^{\theta}$, are: 
\begin{equation}
\begin{array}{c}
\mathbf{x}_{arr\_i}^{p}=\left[x_{arr\_i}^{x};x_{arr\_i}^{y};x_{arr\_i}^{z}\right],\text{ }\mathbf{x}_{arr\_i}^{\theta}=\left[\theta_{arr\_i}^{x};\theta_{arr\_i}^{y};\theta_{arr\_i}^{z}\right],\end{array}\label{eq:X}
\end{equation}
respectively, where $\theta_{arr\_i}^{x},\theta_{arr\_i}^{y}$, and
$\theta_{arr\_i}^{z}$ take values in the range of $[-\pi,\pi],[-\frac{\pi}{2},\frac{\pi}{2}]$,
and $[-\pi,\pi],$ respectively. Denote the unknown parameters w.r.t.
the $i\raisebox{0mm}{-}th$ microphone array as: 
\begin{equation}
\mathbf{x}_{arr\_i}=\left[\mathbf{x}_{arr\_i}^{p};\mathbf{x}_{arr\_i}^{\theta};x_{arr\_i}^{\tau};x_{arr\_i}^{\delta}\right].\label{eq_mic_state_vector_3d}
\end{equation}
All the unknown parameters w.r.t. microphone arrays are: 
\begin{equation}
\mathbf{x}_{arr}=\left[\mathbf{x}_{arr\_2};\ldots;\mathbf{x}_{arr\_N}\right].\label{eq_mic_state_vector}
\end{equation}
Denote the sound source position at time $t^{k},$ $k=1,\ldots,K$
as: 
\begin{equation}
\mathbf{s}^{k}=\left[s_{x}^{k};s_{y}^{k};s_{z}^{k}\right].\label{eq:Source location}
\end{equation}
Thus, all unknown parameters to be identified are: 
\begin{equation}
\mathbf{x}=\left[\mathbf{x}_{arr};\mathbf{s}^{1};\ldots;\mathbf{s}^{K}\right].\label{eq_state_vector}
\end{equation}
We denote the ideal inter-array TDOA and DOA measurements at the $k\raisebox{0mm}{-}th$\textsuperscript{}
time instance as:
\begin{equation}
\mathbf{m}^{k}=\left[\mathbf{d}_{1}^{k};T_{2}^{k};\mathbf{d}_{2}^{k};T_{3}^{k};\mathbf{d}_{3}^{k};\ldots;T_{N}^{k};\mathbf{d}_{N}^{k}\right]\in\mathbf{\mathbb{R}}^{4N-1}.\label{eq_p_l_ob}
\end{equation}
The measurements of DOA and inter-array TDOA at time $k$ are subject to Gaussian noises as follows: 
\begin{equation}
\mathbf{y}^{k}=\mathbf{m}^{k}+\mathbf{v}^{k}\label{measurements}
\end{equation}
where $\mathbf{m}^{k}$ is defined in (\ref{eq_p_l_ob}), $\mathbf{v}^{k}\sim\mathcal{N}(\mathbf{0},\mathbf{P})$,
with $\mathbf{P}=diag(\Lambda,diag_{N-1}(\lambda,\Lambda))$, where
$\ensuremath{\lambda>0}$ is a positive scalar, $\ensuremath{\Lambda>\mathbf{0},}$
and $\Lambda\in\mathbf{\mathbb{R}}^{3\times3}$. Assume that the sound
source relative position between two consecutive time steps can be
measured with Gaussian noise, i.e., 
\begin{equation}
\mathbf{s}_{\Delta}^{k}=\mathbf{s}^{k+1}-\mathbf{s}^{k}+\mathbf{w}^{k}\label{random_walk}
\end{equation}
where $k=1,...,K-1$, $\mathbf{w}^{k}\sim\mathcal{N}(\mathbf{0},\mathbf{Q})$,
with $\mathbf{Q}>\mathbf{0}\in\mathbf{\mathbb{R}}^{3\times3}$. We
combine the relative position measurements, the TDOA, and DOA measurements
as: 
\begin{equation}
\mathbf{z}=\left[\mathbf{y}^{1};\mathbf{s}_{\Delta}^{1};\mathbf{y}^{2};\mathbf{s}_{\Delta}^{2};\ldots;\mathbf{y}^{K-1};\mathbf{s}_{\Delta}^{K-1};\mathbf{y}^{K}\right].\label{eq_ob_M-1}
\end{equation}
The models in (\ref{measurements})-(\ref{random_walk}) can be rewritten
compactly as: 
\begin{equation}
\mathbf{z}=\mathbf{g}(\mathbf{x})+{\gamma}\label{eq_ob_M_2-1}
\end{equation}
where $\mathbf{g}(\mathbf{x})$ is the combined observation model,
and ${\gamma}\sim\mathcal{N}(\mathbf{0},\mathbf{W})$ is the
noise of combined observations with 
\begin{equation}
\mathbf{W}=diag(diag_{K-1}(\mathbf{P,Q),P}).\label{eq_ob_W-1}
\end{equation}

As shown in Fig. \ref{TDOA}, the batch SLAM framework is a
feasible solution to the above problem by treating the moving sound
source as a robot and the multiple microphone arrays as landmarks
\cite{Thrun2006}. As in \cite{Su2015}-\cite{Su2020}, the problem
of joint calibration of multiple asynchronous microphone arrays and
sound source localization can be treated as the following NLS using
batch SLAM: 
\begin{equation}
\noindent \min\limits _{{\mathbf{x}}}\left\Vert \mathbf{g}({\mathbf{x}})\mathbf{-z}\right\Vert _{\mathbf{W}^{-1}}^{2}\label{eq_ls}
\end{equation}
The measurements obtained by microphone arrays and robots constitute
the spatial constraints and can be included in (\ref{eq_ls}) to improve
estimation accuracy.

Given the problem formulation described above, our main objective
is (1) to determine the identifiability of the unknown parameters (microphone
arrays positions, orientations, time offsets, sampling clock differences, and sound
source positions) based on the available measurements (DOAs, inter-array TDOAs,
and relative position measurements), and (2) to develop an efficient algorithm pipeline
for solving the corresponding NLS in (\ref{eq_ls}).

\section{\label{observability}OBSERVABILITY ANALYSIS}

In this section, by utilizing the FIM method, the observability analysis of the batch SLAM framework for the above calibration problem is performed. More specifically, we have established necessary/sufficient conditions under which the FIM and Jacobian matrix have full column rank (which implies the identifiability of the
unknown parameters, including the microphone array positions, orientations, time offsets, sampling clock differences, and sound source positions). In addition, we also discover some scenarios where the FIM and Jacobian matrix cannot have full column rank (in this case, the unknown parameters could not be uniquely identified). 

\subsection{The Fisher Information Matrix and the Jacobian}
The covariance matrix  $\mathbf{C}_{\hat{x}}$ of the estimation error corresponding to the estimated values $\hat{\mathbf{x}}$  and the true values $\check{\mathbf{x}}$ of unknown parameters  in the observation model in (\ref{eq_ob_M_2-1}) can be calculated by
\begin{equation}
\mathbf{C}_{\hat{x}}=E\left[(\hat{\mathbf{x}}-\check{\mathbf{x}})(\hat{\mathbf{x}}-\check{\mathbf{x}})^{\mathit{\mathrm{T}}}\right].\label{eq:covariance}
\end{equation}
For nonrandom vector parameter estimation, the FIM of an unbiased
estimator is defined as:
\begin{equation}
\mathbf{I}_{FIM}=E\left[(\nabla_{x}\ln p(\mathbf{z}|\mathbf{x}))(\nabla_{x}\ln p(\mathbf{z}|\mathbf{x}))^{\mathrm{T}}\right],\label{eq:FIM}
\end{equation}
where {$\nabla_{x}$ is the gradient operator w.r.t. the vector $\mathbf{x}$,} $p(\mathbf{z}|\mathbf{x})$ is the probability distribution
function, and the derivatives are calculated at the true value $\check{\mathbf{x}}$ \cite[chap. 2]{Bar-Shalom2004}.
It can be shown that the covariance matrix of any unbiased
estimator $\hat{\mathbf{x}}$ satisfies
\begin{equation}
\mathbf{C}_{\hat{x}}-\mathbf{I}_{FIM}^{-1}\geq\mathbf{0},
\end{equation}
i.e., when the $\mathbf{I}_{FIM}$ is singular, the Cram$\acute{\mathrm{e}}$r-Rao 
lower bound will not exist \cite[pp. 165]{Bar-Shalom2004}, one or more parameters
will be unobservable. As in \cite{Dissanayake2008}, the {Fisher information matrix} for the models described in (\ref{eq:FIM}) can be formulated as:
\begin{equation}
\mathbf{I}_{FIM}=\mathbf{J^{\mathrm{T}}W^{\mathrm{-1}}J},
\end{equation}
where $\mathbf{J}$ is the Jacobian of the observation model $\mathbf{g}(\cdot)$
in (\ref{eq_ob_M_2-1}), and its explicit expressions will be given
in (\ref{eq:Jacobi}). When $\mathbf{W}>\mathbf{0},$ one has that 
\begin{equation}
rank(\mathbf{I}_{FIM})=rank(\mathbf{J}).\label{eq_ob_rank_J}
\end{equation}
Since the first microphone array is viewed as the reference array, its
corresponding parameters are all set to zero. The remaining state
vectors contain only $\left(N-1\right)$ microphone arrays parameters
$\mathbf{x}_{arr}$ and the sound source position $\mathbf{s}^{k}$
at all $K$ time steps. 
From the definition of the Jacobian matrix \cite[pp. 569]{Siciliano2009},
we know that $\mathbf{J}\in\mathbb{R}^{g_{1}\times g_{2}}$, where
\[
g_{1}=4(N-1)K+3(K-1),\text{ }g_{2}=8(N-1)+3K.
\]
From (\ref{eq:FIM})-(\ref{eq_ob_rank_J}), a necessary and sufficient
condition for $\mathbf{I}_{FIM}$ to be nonsingular is that $\mathbf{J}$
has full column rank. For $\mathbf{J}$ to be of full column rank,
it is necessary that 
\begin{equation}
\begin{array}{l}
4(N-1)K+3(K-1)\geq8(N-1)+3K\\
\implies K\geqslant\left\lceil 2+\dfrac{3}{4(N-1)}\right\rceil ,
\end{array}\label{eq:neccessary_condion}
\end{equation}
where $\left\lceil \cdot\right\rceil$ stands for the ceiling operation generating the least integer not
less than the number within the operator. We then have the following
results.

\textit{Proposition 1:} The Jacobian $\mathbf{J}$ can be written as
\begin{equation}
\mathbf{J}={\left[\begin{array}{c}
\mathbf{L}^{1}\\
\mathbf{0}\\
\mathbf{L}^{2}\\
\mathbf{0}\\
\vdots\\
\mathbf{L}^{K-1}\\
\mathbf{0}\\
\mathbf{L}^{K}
\end{array}\right.}{\left.\begin{array}{ccccc}
\mathbf{T}^{1} & \mathbf{0} & \cdots & \mathbf{0} & \mathbf{0}\\
-\mathbf{I}_{3} & \mathbf{I}_{3} & \cdots & \mathbf{0} & \mathbf{0}\\
\mathbf{0} & \mathbf{T}^{2} & \cdots & \mathbf{0} & \mathbf{0}\\
\mathbf{0} & -\mathbf{I}_{3} & \cdots & \mathbf{0} & \mathbf{0}\\
\vdots & \vdots & \ddots & \vdots & \vdots\\
\mathbf{0} & \mathbf{0} & \cdots & \mathbf{T}^{K-1} & \mathbf{0}\\
\mathbf{0} & \mathbf{0} & \cdots & -\mathbf{I}_{3} & \mathbf{I}_{3}\\
\mathbf{0} & \mathbf{0} & \cdots & \mathbf{0} & \mathbf{T}^{K}
\end{array}\right]}\label{eq:Jacobi}
\end{equation}
{where $\mathbf{L}^{k}=
\frac{\partial\mathbf{y}^{k}(\mathbf{x}_{arr},\mathbf{s}^{k})}{{\partial}\mathbf{x}_{arr}}$, $\mathbf{T}^{k}=\frac{\partial\mathbf{y}^{k}(\mathbf{x}_{arr},\mathbf{s}^{k})}{\partial\mathbf{s}^{k}}$
with $\mathbf{y}^{k}(\mathbf{x}_{arr},\mathbf{s}^{k})$ being the inter-array TDOA and DOA observation model at the $k\raisebox{0mm}{-}th$ time instant, $k=1,...,K$ (expression of $\mathbf{y}^{k}(\mathbf{x}_{arr},\mathbf{s}^{k})$ can be found in (\ref{measurements}); the detailed expressions of $\mathbf{L}^{k}$ and $\mathbf{T}^{k}$ can be found in 
(\ref{eq:L}) and (\ref{eq:part TK}) in Appendix A, respectively).}

\begin{Proof} See Appendix A.
\end{Proof}

{Given the equivalence of full rankness between the FIM and the Jacobian, in the following, we will focus on investigating conditions under which the Jacobian derived in (\ref{eq:Jacobi}) can or can not be of full column rank.}

\subsection{Main Results of Observability}

{We firstly have the following results regarding the equivalence of full column rank between the Jacobian (\ref{eq:Jacobi}) and matrix $\mathbf{F}$ in (\ref{eq:L,T}) which has a much simpler structure.}

\begin{theorem} The Jacobian matrix $\mathbf{J}$ is of full column
rank if and only if the following matrix 
\begin{equation}
\mathbf{F}=\underset{\mathbf{L}}{\underbrace{\left[\begin{array}{c}
\mathbf{L}^{1}\\
\mathbf{L}^{2}\\
\vdots\\
\mathbf{L}^{K}
\end{array}\right.}}\underset{\mathbf{T}}{\underbrace{\left.\begin{array}{c}
\mathbf{T}^{1}\\
\mathbf{T}^{2}\\
\vdots\\
\mathbf{T}^{K}
\end{array}\right]}}\label{eq:L,T}
\end{equation}
is of full column rank. \end{theorem}

\begin{Proof} The proof is similar to that of \cite[Theorem 1]{Kong2021} and is skipped here. 
\end{Proof}

{We next present a necessary condition (Theorem 2) and a sufficient condition (Theorem 3) under which matrix $\mathbf{F}$ in (\ref{eq:L,T}) is of full column rank.}

\begin{theorem} {The Jacobian matrix $\mathbf{J}$ is of full column rank only if matrices $\mathbf{\bar{T}}$ and $\mathbf{\bar{L}}_{i}$, for $i=2,\ldots,N,$ are of full column rank, respectively\footnote{As shown in the full proof in Appendix A, submatrices $\mathbf{\bar{T}}$ and $\mathbf{\bar{L}}_{i}$ are obtained from the matrices after applying elementary transformations to $\mathbf{T}$ and $\mathbf{L}$ (both defined in (\ref{eq:L,T})), respectively.},
 where 
\begin{equation}
\mathbf{\bar{T}}=\left[\mathbf{0};\Psi;\mathbf{0}\right],\text{ }\mathbf{\bar{L}}_{i}=\left[\begin{array}{cc}
\mathbf{I}_{2} & \mathbf{0}\\
\mathbf{0} & \Phi_i
\end{array}\right],\label{eq:easyTLbar}
\end{equation}
with $\Psi$ and $\Phi_i$ being defined in (\ref{eq:T-BAR}) and (\ref{eq: L-BAR}), respectively.}

\begin{Proof} See Appendix A.
\end{Proof}

\end{theorem} \begin{theorem} The Jacobian matrix $\mathbf{J}$
is of full column rank if the following statements hold concurrently:

(i) Any matrix  resulting from the horizontal concatenation of $\mathbf{\bar{L}}_{j}$ and $\mathbf{\bar{T}}$
is of full column rank, $2\leq j\leq N$.

(ii) All matrices $\mathbf{\bar{L}}_{i}$, $i=2,\ldots,N$ and $i\neq j$
are of full column rank.\end{theorem}

\begin{Proof} See Appendix A.
\end{Proof}


\subsection{Special Cases When Observability is Impossible}
{It can be seen from Proposition 1 and Theorems 1-3 that observability
of the considered identification question is determined both by the configuration of microphone arrays (i.e., the relative transforms, namely, orientation and translation) and the sound source positions.} This raises the question of under what
conditions on the microphone array configuration and the sound
source trajectory, the necessary conditions
in Theorem 2 cannot hold. In this section, we will focus on this
question and discover some special cases where observability
is impossible. Our major result is stated in Theorems 4-5.

\begin{theorem}The matrix $\mathbf{\bar{T}}$ is not of full column
rank if one or more of the following conditions hold.

(i) For all microphone arrays, there exists fewer than five time steps information (i.e., the value of $K$ in (\ref{eq:L,T}) is less than 5).

(ii) The sound source positions at all moments are collinear with the origin of the global frame $\left\{ \mathrm{\mathbf{x}}_{arr\_1}\right\} $, i.e., $\mathbf{\mathbf{s}}^{k}={\lambda}_{k-1}\mathbf{s}^{k-1}$ always holds, where $k=2, \ldots, K$, and ${\lambda}_{k-1}$ is an arbitrary non-zero scalar (${\lambda}_{k-1}$ might take different values at different time steps).


(iii) The sound source lies on any Euclidean plane of $x+\alpha y=0$,
$x+\beta z=0$, and $y+\gamma z=0$  within the three-dimensional $x$-$y$-$z$ Cartesian coordinate frame $\left\{ \mathrm{\mathbf{x}}_{arr\_1}\right\} $, at all moments, where $\alpha,\beta,\gamma$ are arbitrary scalars.

\end{theorem} 

\begin{Proof} See Appendix A.
\end{Proof}

\begin{theorem}The matrices $\mathbf{\bar{L}}_{i}$, $i=2,3,\cdots,N$,
are not of full column rank if one or more of the following conditions
hold:

(i) The sound source positions at all moments are collinear with the origin of the frame $\left\{ \mathrm{\mathbf{x}}_{arr\_i}\right\} $, i.e., $(\mathbf{\mathbf{s}}^{k}-\mathbf{x}_{arr\_i}^{p})={\epsilon}_{k-1}(\mathbf{\mathbf{s}}^{k-1}-\mathbf{x}_{arr\_i}^{p})$ always holds, where $k=2, \ldots, K$ and ${\epsilon}_{k-1}$ is
an arbitrary non-zero scalar (${\epsilon}_{k-1}$ might take different values at different time steps).

(ii) For the $i\raisebox{0mm}{-}th$ microphone array, one of the
Euler angles satisfies $\theta_{arr\_i}^{y}=\pm \frac{\pi}{2}$.

\end{theorem}

\begin{Proof} See Appendix A.
\end{Proof}

\subsection{Discussions} 
{The observability analysis presented in the above subsections 
refers to conditions concerning the ground truth value of the sound source trajectories or the configurations of microphone arrays. Hence, the observability analysis is of theoretical interest as it can serve as guidelines when designing microphone array configurations or the sound source trajectories during the calibration process. One can also rely on the results of Section III.C to avoid the unobservable scenarios from a theoretical point of view.}

However, during real calibration processes, the measurements contain noises (i.e., the ground truth is not known a prior). Hence, the observability analysis results obtained above are not directly applicable. It is crucial to develop a reliable algorithmic pipeline that can achieve satisfactory convergence and accuracy. This will be discussed in the next section. One should note that the algorithmic pipeline presented in the sequel can also be applied to the nonobservable cases (but the calibration results will be unreliable). This is because, for these scenarios, the noisy measurements do not contain enough information to estimate the unknown parameters. This is also why the analysis in Sections III.A to III.C is valuable, as it suggests avoiding such unobservable situations when designing the microphone array configurations or the sound source trajectories.

{Based on the above arguments, to validate the theoretical analysis, we will discuss both observable and unobservable situations in the numerical simulations in Section V. In the experimental results of Section VI, we will only design experiments that correspond to observable cases.}

\section{\label{3D}BATCH SLAM BASED CALIBRATION}
In this section, we present our proposed
pipeline for batch SLAM based joint calibration of multiple microphone arrays and sound source localization. As illustrated in Fig. \ref{TDOA},
we treat the microphone arrays as landmarks and the sound source as
a mobile robot in the corresponding batch SLAM problem and utilize
Gauss–Newton iterations to solve the corresponding NLS problem. More specifically, we propose an effective framework to initialize the unknown parameters which are used as the initial guess in the Gauss–Newton iterative algorithm.

\subsection{\label{initial_method}The Proposed Initialization Procedure}
For notational simplicity, in the sequel, we use 
$\mathbf{d}_{i}^{k}$ and $T_{i}^{k}$ to denote the Gaussian noise corrupted DOA and inter-array TDOA measurements, respectively.
We use $\hat{\cdot}$ to represent the estimates of the unknown scalar/vector/matrix parameters. Our proposed initialization procedure is composed of the following main steps: (i) estimation of the sound source position by triangulation; (ii) estimation of the distance between the sound source and microphone arrays
using 3D geometry; (iii) estimation of microphone array poses using ICP; (iv) estimation
of the asynchronous factors using LLS.

\begin{figure}[t]
\centering \subfigure[]{\includegraphics[width=0.46\columnwidth]{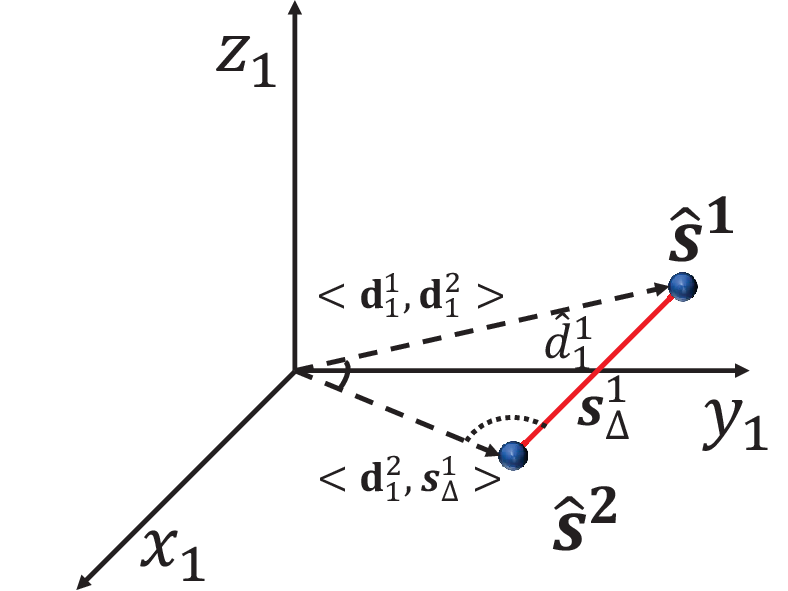}}
\centering\subfigure[]{\includegraphics[width=0.46\columnwidth]{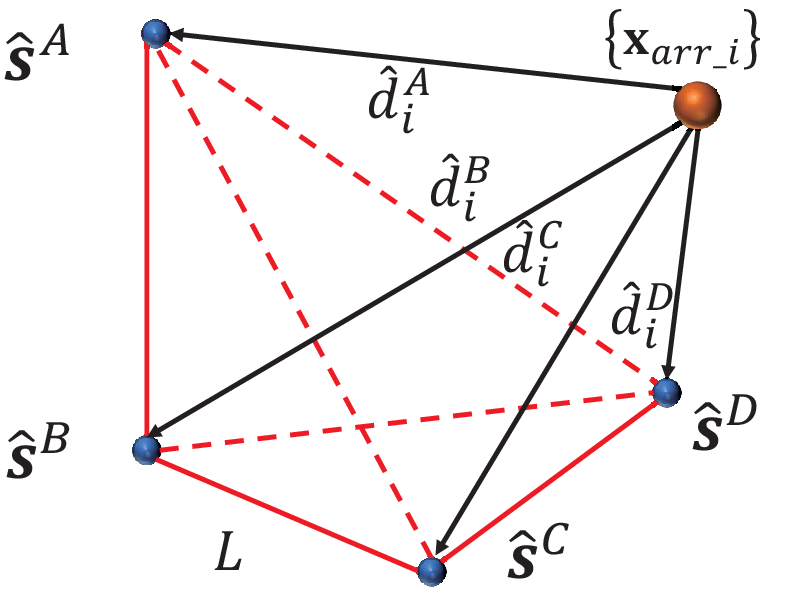}}
\centering\subfigure[]{\includegraphics[width=0.46\columnwidth]{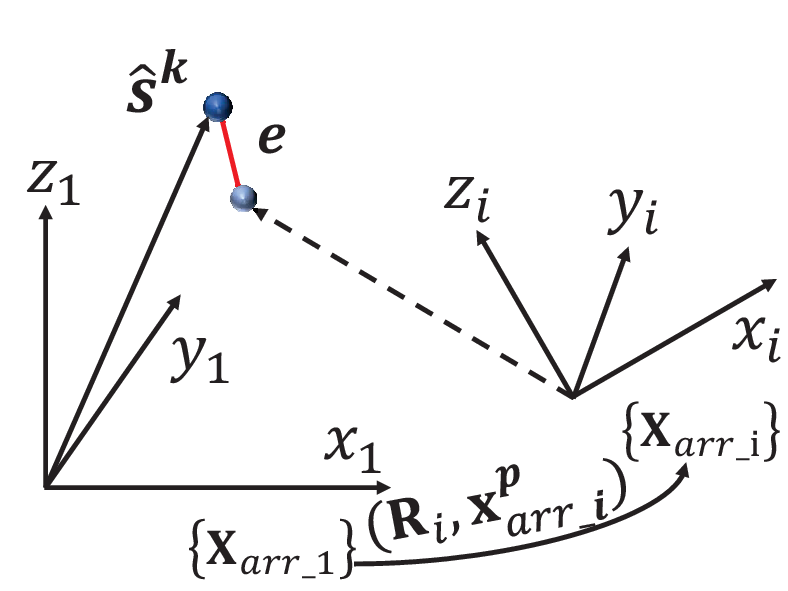}}
\centering\subfigure[]{\includegraphics[width=0.46\columnwidth]{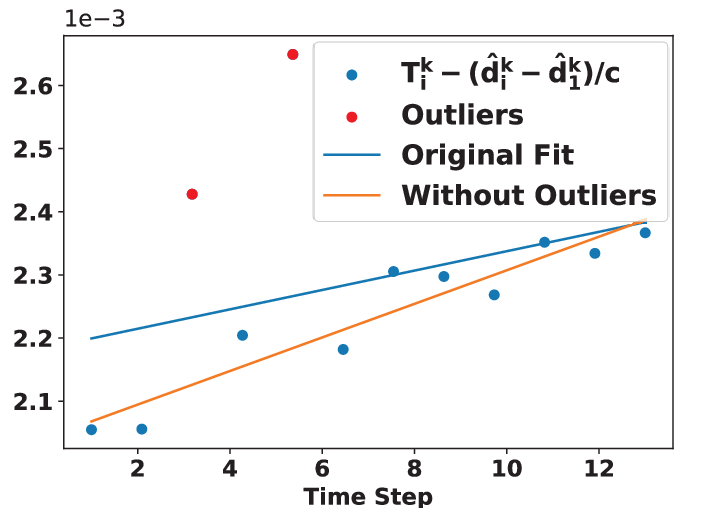}}
\caption{Initialization process of unknown parameters for the microphone arrays and sound source. (a) Estimation of the initial position of the sound source by triangulation. (b) Estimation of the distances between the sound source and microphone arrays using 3D geometry. (c) Estimation of microphone arrays initial positions and orientations using ICP. (d) {Estimation of inter-array initial time offset and sampling clock difference using LLS.}}
\label{fig} 
\end{figure}

\textbf{(i) Estimation of the sound source position by triangulation:} 
Without loss of generality, the initial trajectory of the moving sound source in the global frame $\left\{ \mathrm{\mathbf{x}}_{arr\_1}\right\}$ is illustrated in Fig. 2(a). Then, from geometry, the initial position of the sound source can be obtained by triangulation and 
using the first two consecutive DOA measurements as follows:
\begin{equation}
\hat{d}_{1}^{1}=\frac{L\sin(\left\langle \mathbf{d}_{1}^{2},\mathbf{s}_{\Delta}^{1}\right\rangle)}{\sin(\left\langle \mathbf{d}_{1}^{1},\mathbf{d}_{1}^{2} \right\rangle)}, \text{ } \hat{\mathbf{s}}^{1}=\mathbf{d}_{1}^{1}\cdot \hat{d}_{1}^{1} \label{eq:start l}
\end{equation}
where $L=\left\Vert \mathbf{s}_{\Delta}^{1}\right\Vert _2  $, i.e., $L$ is the measured distance that the source moves between the first two consecutive
moments, $\left\langle \cdot ,\cdot \right\rangle $ is the angle of two vectors, and $\hat{d}_{1}^{1}$ is the distance between the first sound source position and
the origin. Note that $\mathbf{s}_{\Delta}^{1}, \mathbf{d}_{1}^{1},\mathbf{d}_{1}^{2}$ can be obtained from the relative position and DOA measurements, respectively. Once the initial position of the sound source is obtained as above, the sound source positions at different
time steps can be estimated:
\begin{equation}
\hat{\mathbf{s}}^{k+1}=\mathbf{\hat{s}}^{k}+\mathbf{s}_{\Delta}^{k}.\label{eq:relative position}
\end{equation}

\textbf{(ii) Estimation of the distance between the sound source and microphone
arrays using 3D geometry:} We calculate the distance between
each source node and microphone arrays to provide constraints for
estimating microphone array poses. One can construct an over-constrained
NLS for estimating the distance $\hat{d}_{i}^{k}$ between each source node
and microphone arrays by using the law of cosines constraints. To
illustrate, as shown in Fig. \ref{fig}(b), each microphone array
and any four source positions $A$, $B$, $C$, and $D$ at the corresponding time instances form a polyhedron (when
the four nodes are coplanar, it is tetrahedral, and when the four
nodes are on different planes, it forms a five-vertex hexahedral structure).
We construct an NLS problem by enforcing the law of cosines
for each face of the polyhedron (including the two inner faces). For the scenario shown in Fig. \ref{fig}(b), denote the estimated squared
distance between any two sound source nodes $a$, $b$ among the four source positions $A$, $B$, $C$, and $D$ as:
\[
\hat{L}_{ab}^{2}=(\hat{d}_{i}^{a})^{2}+(\hat{d}_{i}^{b})^{2}-2\hat{d}_{i}^{a}\hat{d}_{i}^{b}\cos 
\left\langle \mathbf{d}_{i}^{a},\mathbf{d}_{i}^{b} \right\rangle,
\]
where $\mathbf{d}_{i}^{a}$ and $\mathbf{d}_{i}^{b}$
are the unit direction vectors of the corresponding sides with length
$\hat{d}_{i}^{a}$ and $\hat{d}_{i}^{b}$, respectively. Denote the
difference between $\hat{L}_{ab}^{2}$ and $L_{ab}^{2}$ as:
\[
\begin{array}{c}
F_{m}(a,b)=\hat{L}_{ab}^{2}-L_{ab}^{2}
\end{array}
\]
where $m=1,2,\cdots,6$ and $L_{ab}=\left\Vert \hat{\mathbf{s}}^{a}-\hat{\mathbf{s}}^{b}\right\Vert _{2}$. Consider a system of six nonlinear
equations, given by $F(d_{i}^{A,B,C,D})=\left[F_{1};F_{2};\cdots;F_{6}\right]$.
We use ${d}_{i}^{A,B,C,D}$ to collectively denote the distances between the four sound source positions and the
$i\raisebox{0mm}{-}th$ microphone array, which can be estimated by solving
\begin{equation}\label{eq: EDGE}
\begin{array}{c}
\underset{d_{i}^{A,B,C,D}}{\min}\left\Vert F(d_{i}^{A,B,C,D})\right\Vert _{2}^{2}\\
\mathrm{subject\ to}:\ d_{i}^{A,B,C,D}>0
\end{array}
\end{equation}

{Note that the nonlinear optimization problem in (\ref{eq: EDGE}) features a polynomial cost function with a fixed number of unknown parameter dimensions, namely, four edge lengths. However, the batch optimization problem in (\ref{eq_ls}) has a more intricate objective function, incorporating polynomials, exponentials, and trigonometric functions, with $8(N-1)+3K$ optimization variables, where $N$ and $K$ represents the numbers of microphone arrays and time steps, respectively. Hence, in general, the optimization problem in (\ref{eq: EDGE}) will be much easier to solve (it can be conveniently solved, for instance, using the trust region reflective method \cite{Branch1999}) than the entire batch optimization problem in (\ref{eq_ls}). To improve the estimation accuracy of $\hat{d}_{i}^{k}$ at all time instances $K$, we form combinations by selecting any four sound source positions from all time instances, where the  $i\raisebox{0mm}{-}th$ microphone array-to-$k\raisebox{0mm}{-}th$ source line exists within multiple polyhedra. This implies that we can leverage multiple estimation results to achieve greater accuracy. By solving for the edge lengths of each polyhedron and employing the well-known interquartile range (IQR) method \cite[pp. 236]{Probability}, we calculate the average value of these same edges in different polyhedra. This average serves as the estimated distance $\hat{d}_{i}^{k}$ between the $i\raisebox{0mm}{-}th$ microphone array and the sound source position at the $k\raisebox{0mm}{-}th$ time step.
}

\textbf{(iii) Estimation of microphone arrays positions and orientations using ICP:} Note that the positions of the sound source in the frame $\left\{ \mathrm{\mathbf{x}}_{arr\_i}\right\} $
can be estimated as:
\begin{equation}
\mathbf{\hat{s}}_{i}^{k}=\mathbf{d}_{i}^{k}\cdot \hat{d}_{i}^{k}.\label{eq:s_i_k}
\end{equation}
{We treat the sound source positions as features in each coordinate frame. To find the transformation that optimally aligns the sound source positions with the reference frame is akin to representing the same features in the reference frame. To tackle this challenge, we formulate an NLS problem to minimize the mapping error of sound source positions between $\left\{ \mathrm{\mathbf{x}}_{arr\_i}\right\} $
and $\left\{ \mathrm{\mathbf{x}}_{arr\_1}\right\} $:
\begin{equation}
\underset{\mathbf{{R}_{\mathrm{i}}},\mathbf{{x}}_{arr\_i}^p}{\min}\sum_{k=1}^{K}\left\Vert \mathbf{\hat{s}}^{k}-(\mathbf{{R}}_{i}\mathbf{\hat{s}}_{i}^{k}+\mathbf{{x}}_{arr\_i}^p)\right\Vert _{2}^{2},\label{eq:ICP PRO}
\end{equation}
which is conceptually a point-to-point registration problem that can be tackled effectively using ICP \cite{Pomerleau F}.} Hence, as in \cite{Pomerleau F}, let $\mathbf{p}$ and $\mathbf{p'_i}$ be the geometric mean
of the source position in $\left\{ \mathrm{\mathbf{x}}_{arr\_1}\right\} $
and $\left\{ \mathrm{\mathbf{x}}_{arr\_i}\right\} $, and they can be computed based on the estimated sound source positions $\mathbf{\hat{s}}_{1}^{k}$  and $\mathbf{\hat{s}}_{i}^{k}$, respectively. 
The covariance of the sound source trajectory expressed in the two different frames becomes:
\begin{equation}
\Omega=\sum_{k=1}^{K}\left(\mathbf{\hat{s}}^{k}-\mathbf{p}\right)\left(\hat{\mathbf{s}}_{i}^{k}-\mathbf{p'_i}\right)^{\mathrm{T}}.
\end{equation}
We perform singular value decomposition on this covariance matrix:
\begin{equation}
\Omega=\mathbf{U}\Sigma \mathbf{V^{\mathrm{T}}}.
\end{equation}
The optimal rotation
matrix \cite{Pomerleau F,K. Arun} can be obtained as:
\begin{equation}
\hat{\mathbf{R}}_{i}=\mathbf{UV^{\mathrm{T}}}.
\end{equation}
Then, we can transform the rotation matrix $\hat{\mathbf{R}}_{i}$ into the corresponding ZYX Euler angles \cite{rotation}. Thus, the initial guess 
of microphone array positions can be expressed as:
\begin{equation}
\mathbf{\hat{x}}_{arr\_i}^{p}=\mathbf{p}-\mathbf{\hat{R}}_{i}\mathbf{p'_i}.
\end{equation}

\textbf{(iv) Estimation of microphone arrays asynchronous parameters using LLS:} In part (ii), the distances between the sound source and microphone arrays at different time steps have been estimated. By using the inter-array TDOA measurements, the initial guess of the microphone array asynchronous factors can be obtained by solving the following LLS problem:
\begin{equation}
\underset{{x}_{arr\_i}^{\tau},{x}_{arr\_i}^{\sigma}}{\min}\sum_{k=1}^{K}\left\Vert T_{i}^{k}-\left(\frac{\hat{d}_{i}^{k}}{c}-\frac{\hat{d}_{1}^{k}}{c}\right)-{x}_{arr\_i}^{\tau}-{\Delta_{k}}{x}_{arr\_i}^{\sigma}\right\Vert _{2}^{2}.\label{eq: fitting}
\end{equation}
{To identify outliers and improve the estimation accuracy of the inter-array asynchronous factors, we first solve the optimization problem (\ref{eq: fitting}). Then, we calculate the residuals by determining the differences between the value $T_{i}^{k}-\left(\hat{d}_{i}^{k}-\hat{d}_{1}^{k}\right)/{c}$ and the corresponding fitted value at each time step, and their average and standard deviation. Subsequently, we perform normalization to the residuals, i.e., dividing each residual by the standard deviation to identify and exclude the outliers. Using the data with the outliers removed as described above, we solve the optimization problem (\ref{eq: fitting}) again, and the final estimates of the asynchronous factors are obtained.
}
\begin{algorithm}[t]
\caption{Joint Calibration of Multi-asynchronous Microphone Arrays and Sound
Source Localization}
\label{alg:3-1} \begin{algorithmic} \REQUIRE Sensors measurements
$\mathbf{z}$ \ENSURE Estimation of all unknown parameters $\hat{\mathbf{x}}$

\STATE// Initialize $\hat{\mathbf{x}}$

\STATE Compute the sound source positions $\mathbf{\hat{s}}^{k}$ with Eq. (\ref{eq:start l})-(\ref{eq:relative position});

\FOR{$i\in[2,N]$}

\FOR{$k\in[1,K]$}

\STATE Solve for the distance $\hat{d}_{i}^{k}$ and {the sound source position $\mathbf{\hat{s}}_{i}^{k}$ in frame $\left\{ \mathrm{\mathbf{x}}_{arr\_i}\right\} $ via optimization problems (\ref{eq: EDGE})-(\ref{eq:s_i_k}), respectively;}

\ENDFOR

\STATE 
{$\mathbf{\hat{R}_{\mathrm{i}}},\mathbf{\hat{x}}_{arr\_i}^{p}\leftarrow\arg\min\sum_{k=1}^{K}\left\Vert \mathbf{\hat{s}}^{k}-(\mathbf{{R}}_{i}\mathbf{\hat{s}}_{i}^{k}+\mathbf{{x}}_{arr\_i}^{p})\right\Vert _{2}^{2}$;}

\STATE Transform $\mathbf{\hat{R}_{\mathrm{i}}}$ into ZYX Euler
angles;

\STATE Linear fitting $\hat{x}_{arr\_i}^{\tau},\hat{x}_{arr\_i}^{\sigma}$
with (\ref{eq: fitting});

\ENDFOR

\STATE //{ }Error Minimization

\FOR{$iter$}

\STATE $\mathbf{H}\leftarrow\mathbf{0};\mathbf{b}\leftarrow\mathbf{0};$

\FORALL{$\mathbf{z}_{ij}\in$$\mathbf{z}$}

\STATE Compute $\mathbf{H}_{ij},\mathbf{b}_{ij}$ with Eq. (\ref{eq:H AND b})-(\ref{eq:b_ij});

\STATE$\mathbf{H}\leftarrow\mathbf{H}+\mathbf{H}_{ij};\mathbf{b}\leftarrow\mathbf{b}+\mathbf{b}_{ij}$;

\ENDFOR

\STATE $\mathbf{H}[1:8,1:8]=\mathbf{I}_{8}$; //Fixed the global frame $\left\{ \mathrm{\mathbf{x}}_{arr\_1}\right\} $

\STATE $\triangle\mathbf{x}=\mathbf{H}^{-1}\cdot(-\mathbf{b})$;

\IF{$\left\Vert \triangle\mathbf{x}\right\Vert_2 <\xi$}

\STATE break;

\ELSE

\STATE $\hat{\mathbf{x}}\leftarrow\hat{\mathbf{x}}+\triangle\mathbf{x}$;

\ENDIF

\ENDFOR

\end{algorithmic} 
\end{algorithm}

\subsection{The Batch Optimization Procedure}

As described in (\ref{eq_ls}), we construct a standard NLS problem
by considering the microphone arrays as landmarks and the sound source
locations as robot positions. For the Gauss-Newton iterations,
the increment of each iteration can be obtained by solving:
\[\mathbf{H}\mathbf{\triangle x}=\mathbf{-b},\]
where $\mathbf{H}$ is the approximation matrix of the Hessian matrix
and $\mathbf{b}$ is the coefficient vector \cite{Grisetti2010}:
\begin{equation}
\begin{array}{c}
\mathbf{H}=\sum_{i,j\in\mathcal{C}}\mathbf{H}_{ij}=\sum_{i,j\in\mathcal{C}}\mathbf{J}_{ij}^{\mathrm{T}}\mathbf{W}^{-1}\mathbf{J}_{ij}\\
\mathbf{b}=\sum_{i,j\in\mathcal{C}}\mathbf{b}_{ij}=\sum_{i,j\in\mathcal{C}}\mathbf{J}_{ij}^{\mathrm{T}}\mathbf{W}^{-1}\mathbf{e}_{ij}
\end{array}\label{eq:H AND b}
\end{equation}
where $i$ and $j$ are the two nodes in the graph (formed by the sound source at different positions and microphone arrays), $\mathcal{C}$
is the full set of measurements, and $\mathbf{J}_{ij}$ is the Jacobian matrix
of the error function of the corresponding nodes. For the position-position constraint, denote the error between the expected measurement and real measurement $\mathbf{z}_{p,p}^{k,k+1}$ collected by the robot as:
\begin{equation}
\mathbf{e}_{p,p}^{k,k+1}=\mathbf{{s}}^{k+1}-\mathbf{{s}}^{k}-\mathbf{z}_{p,p}^{k,k+1}.
\end{equation}
The Jacobian matrix w.r.t. position $\mathbf{{s}}^{k}$ and position $\mathbf{{s}}^{k+1}$ are:
\begin{equation}
\begin{array}{cc}
\mathbf{A}_{p,p}^{k,k+1}=\dfrac{\partial\mathbf{e}_{p,p}^{k,k+1}}{\partial\mathbf{\mathbf{{s}}}^{k}}=-\mathbf{I}_{3}, & \mathbf{B}_{p,p}^{k,k+1}=\dfrac{\partial\mathbf{e}_{p,p}^{k,k+1}}{\partial\mathbf{\mathbf{{s}}}^{k+1}}=\mathbf{I}_{3}.\end{array}
\end{equation}
 For the position-landmark constraint, denote the error between the expected measurement and the real measurement $\mathbf{z}_{p,l}^{k}$ collected by microphone arrays as:
\begin{equation}
\mathbf{e}_{p,l}^{k}=\left[\begin{array}{cc}
{T}_{i}^{k}; & {\mathbf{d}}_{i}^{k}\end{array}\right]-\mathbf{z}_{p,l}^{k}.
\end{equation}

\noindent The Jacobian matrices corresponding to landmark $l$ and position $p$ are:
\begin{equation}
\begin{array}{cc}
\mathbf{A}_{p,l}^{k}=\dfrac{\partial\mathbf{e}_{p,l}^{k}}{\partial\mathbf{{x}}_{arr}}, & \mathbf{B}_{p,l}^{k}=\dfrac{\partial\mathbf{e}_{p,l}^{k}}{\partial\mathbf{{s}}^{k}}.\end{array}
\end{equation}
The structure of the Jacobian matrix is elaborated in Eq. (\ref{eq:L})-Eq. (\ref{eq:part TK}). 
For corresponding nodes $i$ and $j$, the Jacobian matrix $\mathbf{J}_{i,j}$ can be succinctly represented as:
\begin{equation}
\mathbf{J}_{i,j}=\left[\mathbf{0};\mathbf{0},\underset{node\ i}{\underbrace{\mathbf{A}_{i,j}}},\mathbf{0},\underset{node\ j}{\underbrace{\mathbf{B}_{i,j}}},\mathbf{0};\mathbf{0}\right].
\end{equation}
By omitting the zero blocks, the corresponding sparse block matrix
$\mathbf{H}_{ij}$ and the vector $\mathbf{b}_{ij}$ (see Eq. (\ref{eq:H AND b})) can be expressed
as:
\begin{small}
\begin{equation}
\mathbf{H}_{ij}=\left[\begin{array}{ccccc}
\ddots\\
 & \mathbf{A}_{i,j}^{\mathrm{T}}\mathbf{W}_{ij}^{-1}\mathbf{A}_{i,j} & \cdots & \mathbf{A}_{i,j}^{\mathrm{T}}\mathbf{W}_{ij}^{-1}\mathbf{B}_{i,j}\\
 & \vdots & \ddots & \vdots\\
 & \mathbf{B}_{i,j}^{\mathrm{T}}\mathbf{W}_{ij}^{-1}\mathbf{A}_{i,j} & \cdots & \mathbf{B}_{i,j}^{\mathrm{T}}\mathbf{W}_{ij}^{-1}\mathbf{B}_{i,j}\\
 &  &  &  & \ddots
\end{array}\right],\label{eq: H_ij}
\end{equation}
\end{small}
\begin{small}
\begin{equation}
\mathbf{b}_{ij}=\left[\begin{array}{c}
\vdots\\
\mathbf{A}_{i,j}^{\mathrm{T}}\mathbf{W}_{ij}^{-1}\mathbf{e}_{i,j}\\
\vdots\\
\mathbf{B}_{i,j}^{\mathrm{T}}\mathbf{W}_{ij}^{-1}\mathbf{e}_{i,j}\\
\vdots
\end{array}\right].\label{eq:b_ij}
\end{equation}
\end{small}respectively. Combining the initial guess selection pipeline and the Gauss-Newton iteration procedure, we then have the entire calibration algorithm as shown in Algorithm \ref{alg:3-1}.

\begin{figure*}[htbp]
    \centering
    \begin{minipage}{0.32\linewidth}
        \centering
        \subfigure[]{
        \includegraphics[width=1\linewidth]{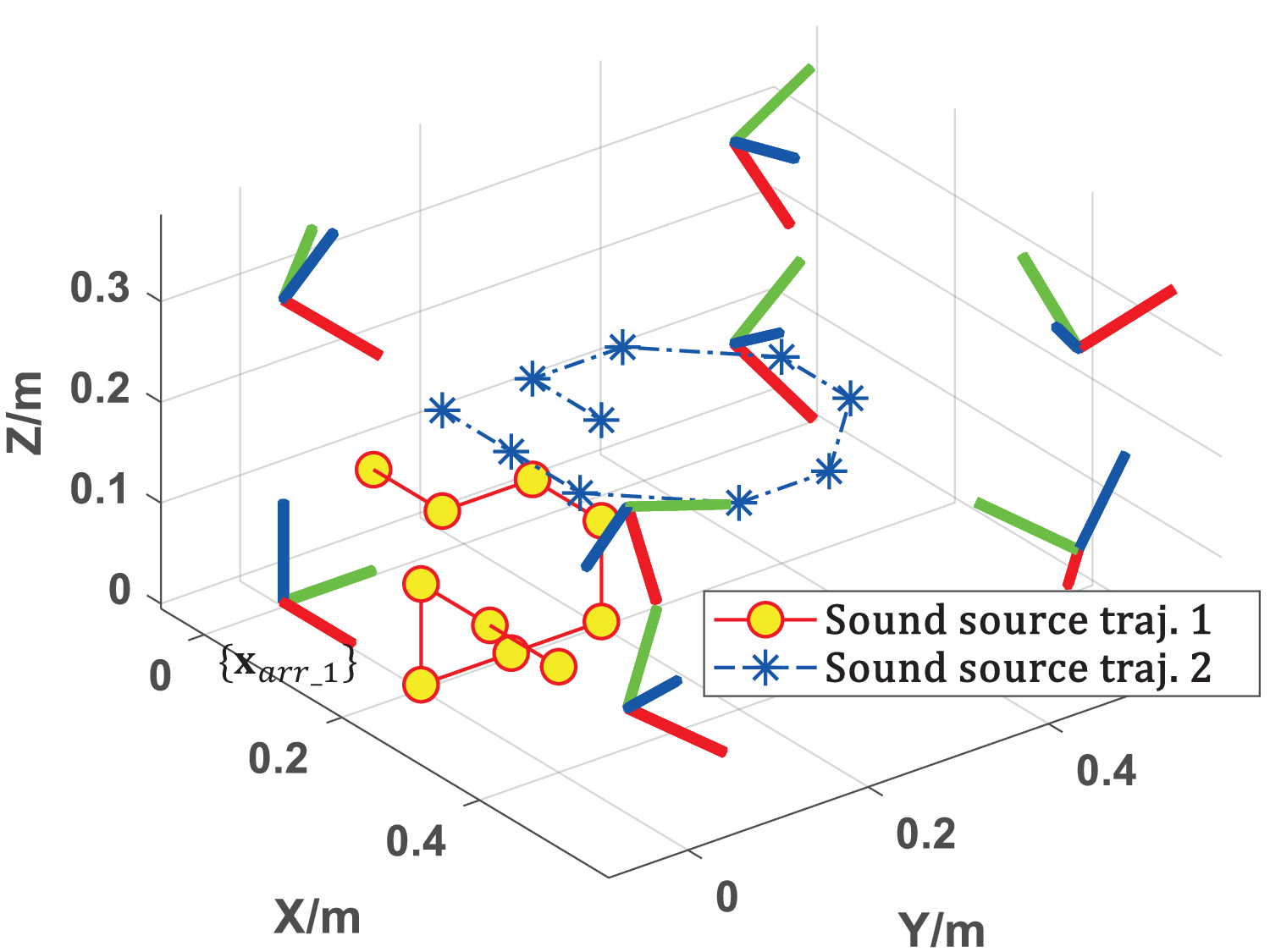}}
        \subfigure[]{
        \includegraphics[width=1\linewidth]{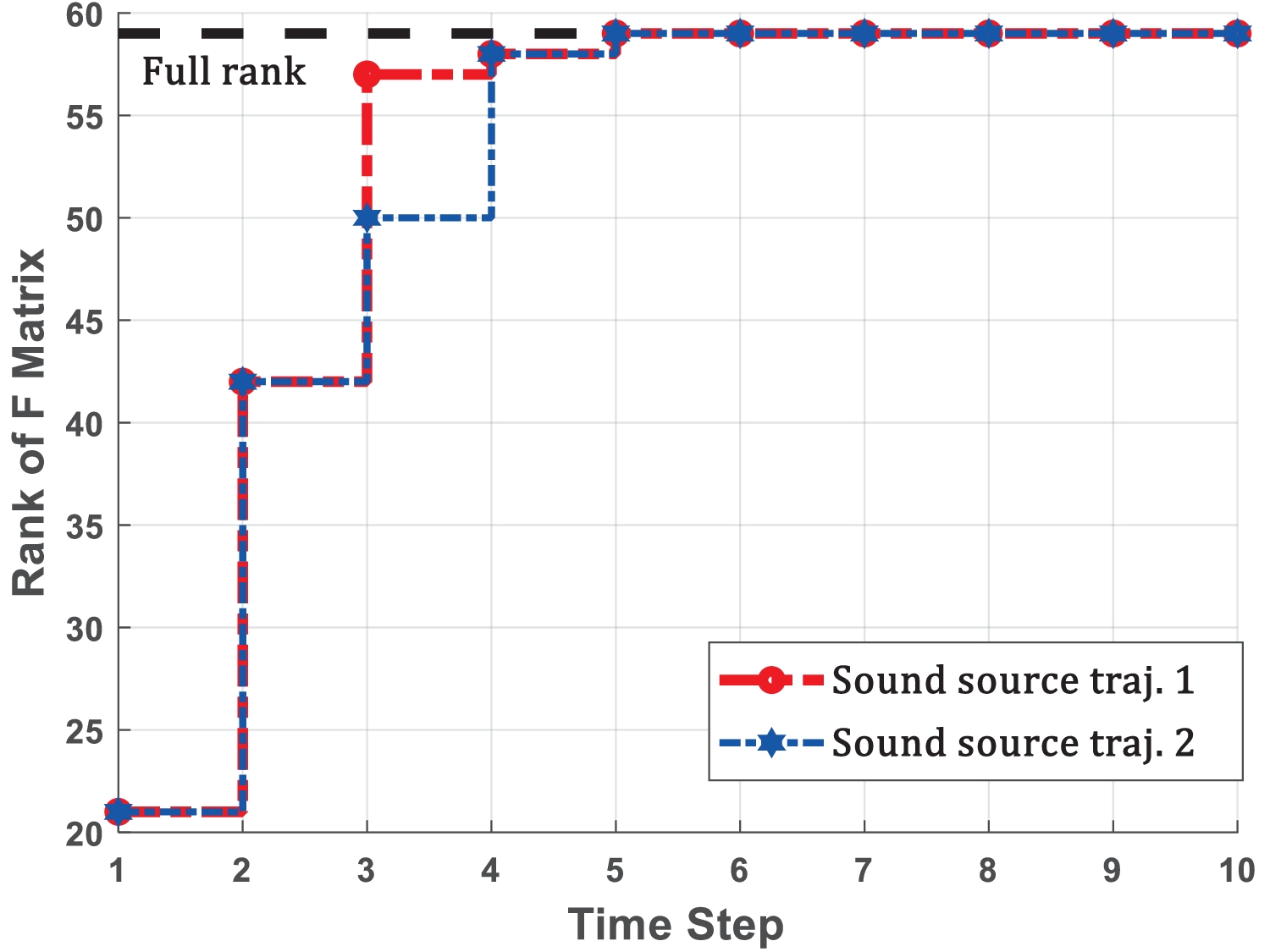}}
    \end{minipage}
    \begin{minipage}{0.32\linewidth}
        \centering
        \subfigure[]{
        \includegraphics[width=1\linewidth]{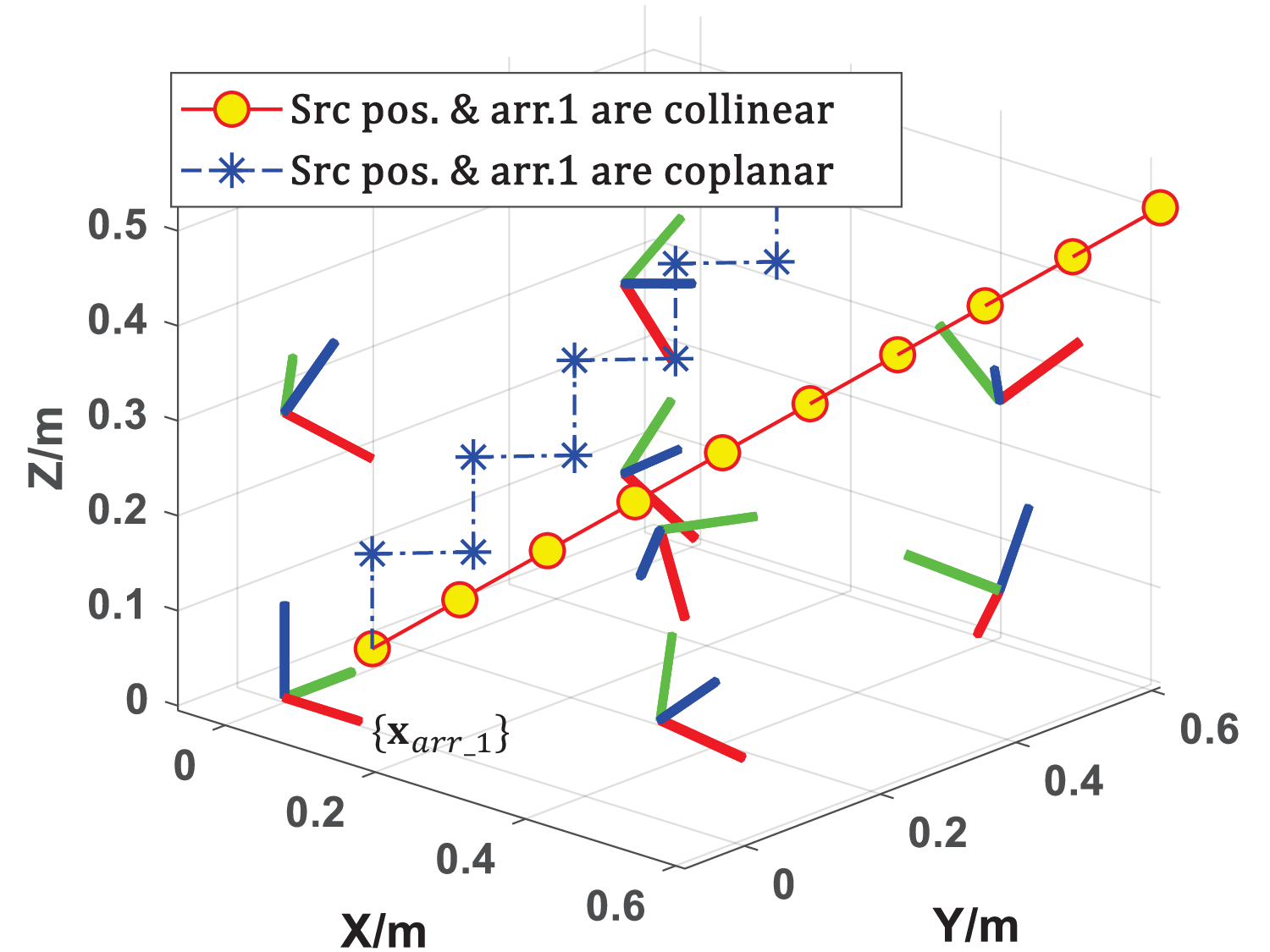}}
        \subfigure[]{
            \includegraphics[width=1\linewidth]{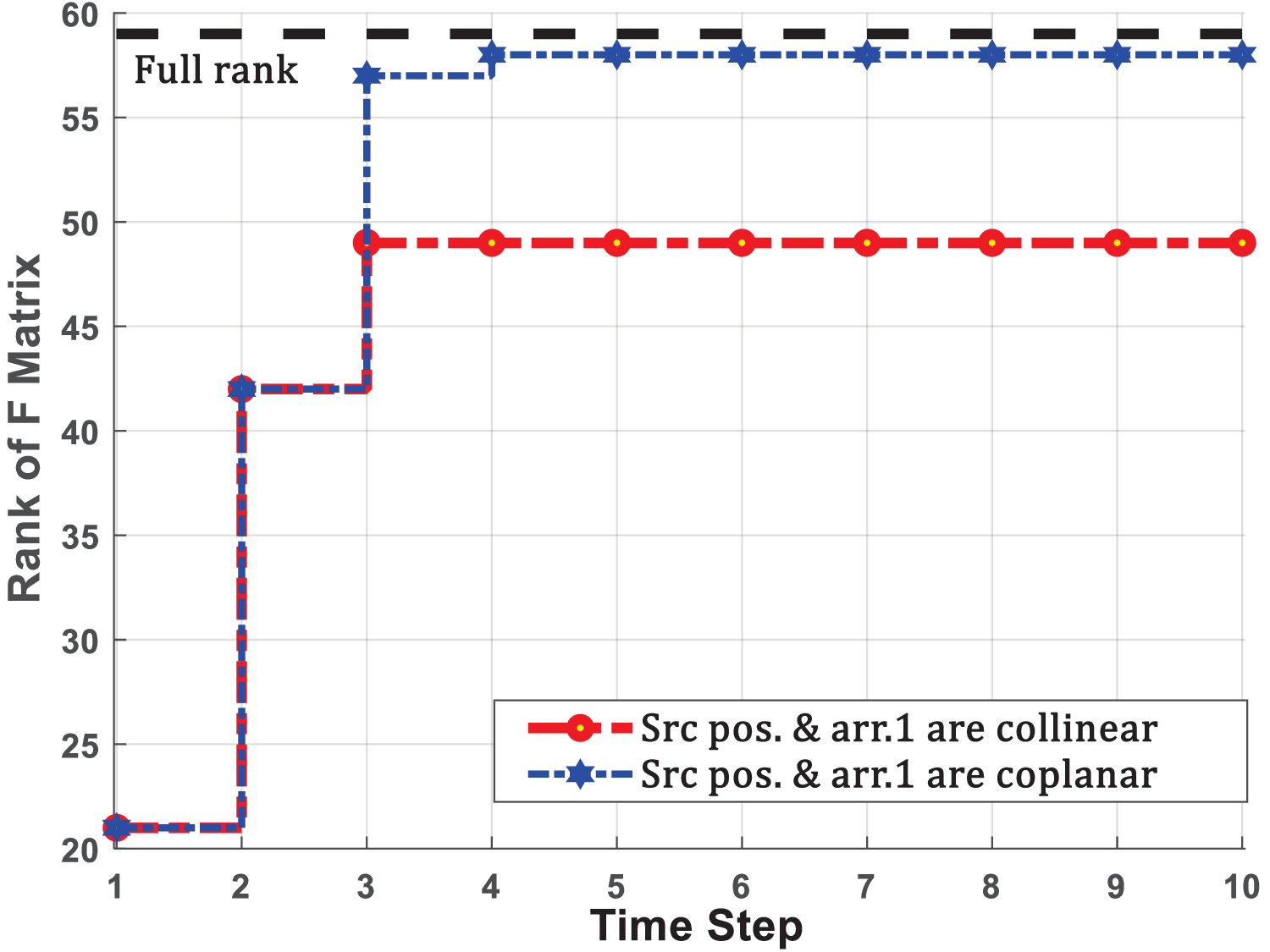}}
    \end{minipage}
    \begin{minipage}{0.32\linewidth}
        \centering
        \subfigure[]{
        \includegraphics[width=1\linewidth]{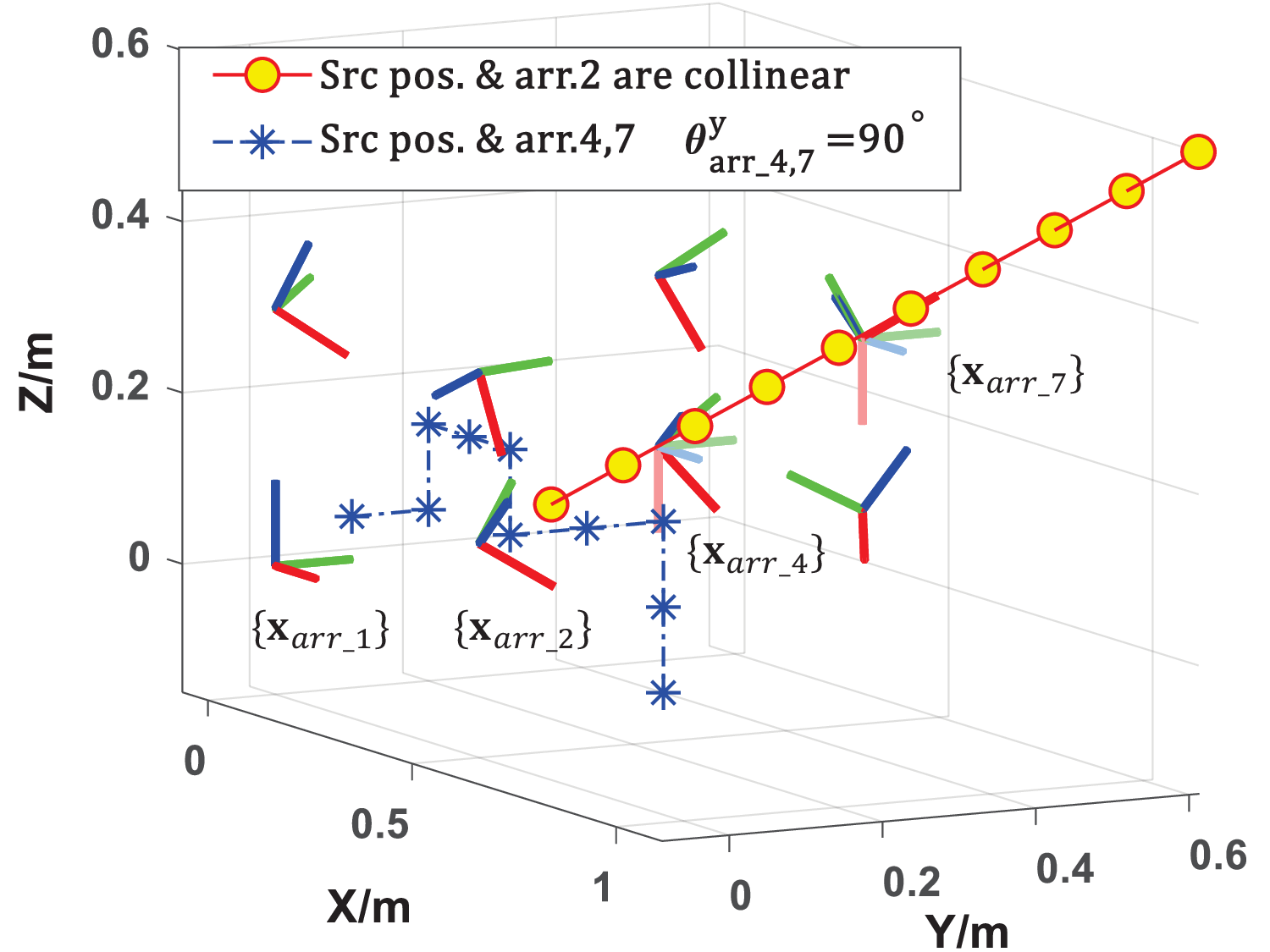}}
        \subfigure[]{
            \includegraphics[width=1\linewidth]{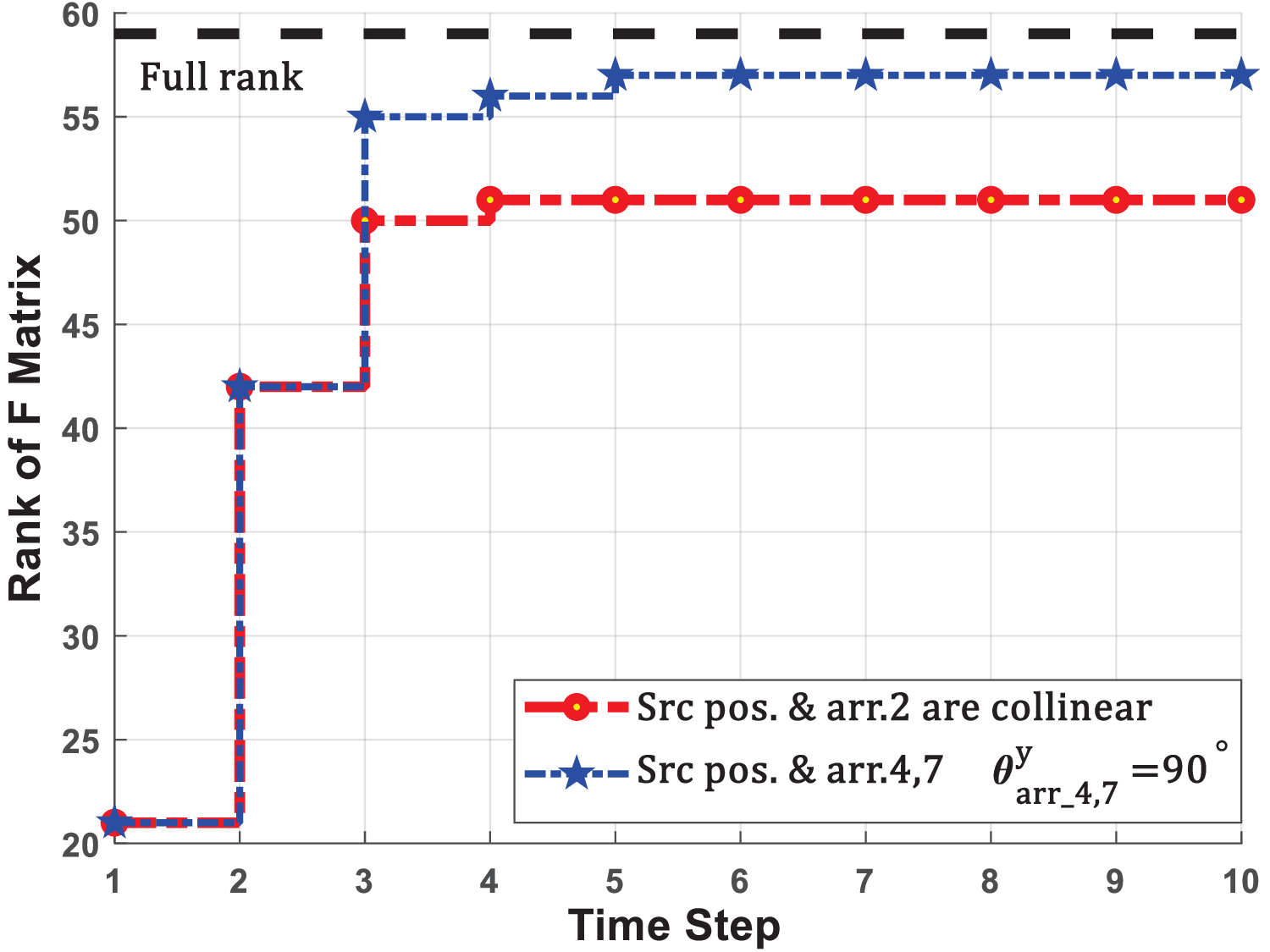}}
    \end{minipage}

    \caption{The scenarios for microphone array calibration and the corresponding variations in the rank of the $\mathbf{F}$ matrices. (a) The geometric relationships between the moving sound source and multiple microphone arrays in two observable cases. (b) Variation of the $\mathbf{F}$ matrix rank with the movement of the source in two observable cases. (c) The geometric relationships when the moving sound source remains co-linear or co-planar with $\left\{ \mathrm{\mathbf{x}}_{arr\_1}\right\}$. (d) Variation of the $\mathbf{F}$ matrix rank in the corresponding unobservable scenarios (e) The geometric relationships when the moving sound source remains co-linear with $\left\{ \mathrm{\mathbf{x}}_{arr\_2}\right\} $
    or $\theta_{arr\_4,7}^{y}=\pi/2$. (f) Variation of the $\mathbf{F}$ matrix rank in the corresponding unobservable scenarios.}
    \label{fig: scenario_rank}
\end{figure*}

\section{\label{simu}NUMERICAL SIMULATIONS AND RESULTS}
We next present extensive numerical simulations to validate the results in Sections \ref{observability} and \ref{3D}. Firstly, we verify the observability analysis results, along with intuitive physical
interpretations of unobservable scenarios. Secondly, we compare
our proposed initialization method (which does not require ground
truth) with initialization schemes of adding {different levels of noise to the ground truth (GT) and random initialization}. Thirdly, we verify
the robustness of the calibration algorithm by varying the sound source
trajectories.

\subsection{Observable Cases}
We firstly present two observable scenarios {as shown in Fig. \ref{fig: scenario_rank}(a). Each scenario comprises eight stationary microphone arrays and a moving sound source. In case 1, the source follows a randomly generated 3D trajectory, while in case 2, it moves along a path on a plane that does not coincide with the global
reference frame. In both scenarios, the moving sound source emits signals at ten consecutive locations, which are recorded by the microphone arrays.} 

The rank of the $\mathbf{F}$ matrix in (\ref{eq:L,T}) changes over time steps, as illustrated in Fig. \ref{fig: scenario_rank}(b). Based on Theorem 3, since $rank(\mathbf{M}_{2\_T})=11$ (note that $\mathbf{M}_{2\_T}$ is defined in (\ref{eq:M2_t})) and
$rank(diag(\mathbf{\bar{L}}_{i}))=48$, $i=3,4,\cdots,8$, it is evident that with an increasing time step and the source's movement along these two trajectories, the $\mathbf{F}$ matrix (with dimensions $336\times59$) gradually become full column rank, i.e. its Jacobian matrix $\mathbf{J}$ (with dimensions $497\times86$) in (\ref{eq:Jacobi}) is full column rank. This implies that the calibration scenarios are observable.
At the time step when the Jacobian matrix becomes full column rank, it also can be verified that $rank(diag(\mathbf{\bar{L}}_{i}))=56$, $i=2,3,\cdots,8$, and $rank(\mathbf{\bar{T}})=3$ so that Theorem 2 holds. 
Hence, the simulations presented so far based on the theoretical analysis worked as expected. It is worth noting that the sound source positions are not always in the {same line with any array frame or on the same plane with the reference array frame}. Hence, a sound source trajectory with more motion varieties often can help to ensure that the necessary conditions stated in Theorem 2 are met, thereby potentially avoiding the unobservable scenarios.

\subsection{Unobservable Cases}
Several unobservable scenarios are presented in the following to verify the conclusions in Theorems 4-5.

(i) For the Jacobian matrix to have full column rank, it is necessary that the time steps are greater than or equal to 3 so that the number
of rows of the Jacobian matrix is greater than the number of
columns, according to (\ref{eq:neccessary_condion}). {In addition}, as can be seen from Fig. \ref{fig: scenario_rank}(b),
when the number of time steps is greater than or equal to 3 but less
than 5, the Jacobian matrix is not of full column rank.
This reflects that the system is unobservable when the number of time steps is less than 5.

(ii) For the sound source trajectories shown in Fig. \ref{fig: scenario_rank}(c), 
the first case is that the sound source stays co-linear with the origin of the global frame $\left\{ \mathrm{\mathbf{x}}_{arr\_1}\right\} $
during the entire process, and ${\lambda}_{1:9}$ in Theorem 4 (ii) take on the values of $2,\dfrac{3}{2},\dfrac{4}{3},\ldots$, and $\dfrac{10}{9}$ respectively. The second case is that the sound source remains co-planar with global frame $\left\{ \mathrm{\mathbf{x}}_{arr\_1}\right\} $. For this scenario, the sound source positions all lie on the Euclidean plane defined by the equation $x-y=0$ within the three-dimensional $x-y-z$ Cartesian coordinate frame $\left\{ \mathrm{\mathbf{x}}_{arr\_1}\right\} $. From Fig. \ref{fig: scenario_rank}(d), we can see that both cases are unobservable due to the rank deficiency of matrix $\mathbf{F}$.

(iii) For the sound source trajectories shown in Fig. \ref{fig: scenario_rank}(e),
the first case is that the sound source keeps co-linear with the origin of $\left\{ \mathrm{\mathbf{x}}_{arr\_2}\right\} $
during the movement, and ${\epsilon}_{1:9}$ in Theorem 4 (iii) take on the values of $2,\dfrac{3}{2},\dfrac{4}{3},\ldots$, and $\dfrac{10}{9}$ respectively. In the second case, the Euler angles $\theta_{arr\_4}^y$ and $\theta_{arr\_7}^y$ of $\left\{ \mathrm{\mathbf{x}}_{arr\_4}\right\} $ and $\left\{ \mathrm{\mathbf{x}}_{arr\_7}\right\} $ are $\frac{\pi}{2}$,
and the sound source travels along the route of the observable
scenario mentioned in case 1 of Fig. \ref{fig: scenario_rank}(f). The rotation angle is at the singular point of observation, rendering the system unobservable. Hence, the simulations presented above validate the conclusions in Theorems 4-5.

\begin{table}
    \begin{center}
    \renewcommand\arraystretch{1.2}
    \caption{\label{tab:1}NUMERICAL SIMULATIONS EXPERIMENT PARAMETERS}
    \centering 
    \begin{tabular}{p{4.5cm} p{3cm}}
    \toprule
    Parameters &	Values \\
    \midrule
    
    Inter-array TDOA noise STD & 0.067ms \\
    
    Elevation angle (DOA) noise STD & 5 degrees \\
    
    Azimuth angle (DOA) noise STD & 5 degrees \\
    
    Relative position noise STD  & $diag_3(0.03m)$ \\
    
    Max. time offset & 0.1s \\
    
    Max. clock difference & 0.1ms \\
    
    Sound speed in air & 346m/s \\
    
    Max. iterations & 50 \\
    
    Threshold $\xi$ & 1e-5 \\

\toprule
\end{tabular}
\end{center}
\end{table}

\subsection{Calibration under Different Initialization Schemes}
\begin{figure*}[htbp]
    \centering
    \begin{minipage}{0.36\linewidth}
        \centering
        \subfigure[]{
        \includegraphics[width=1.05\linewidth]{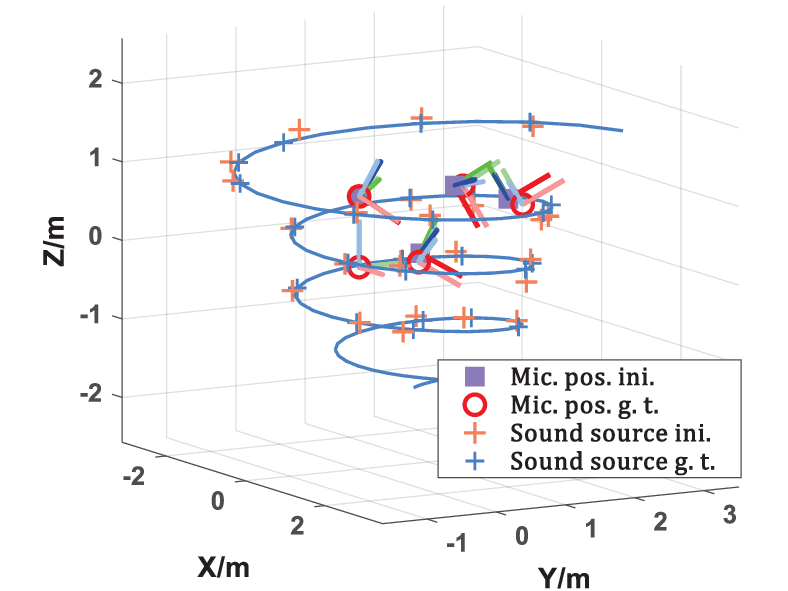}}
    \end{minipage}
    \begin{minipage}{0.36\linewidth}
        \centering
        \subfigure[]{
        \includegraphics[width=1.05\linewidth]{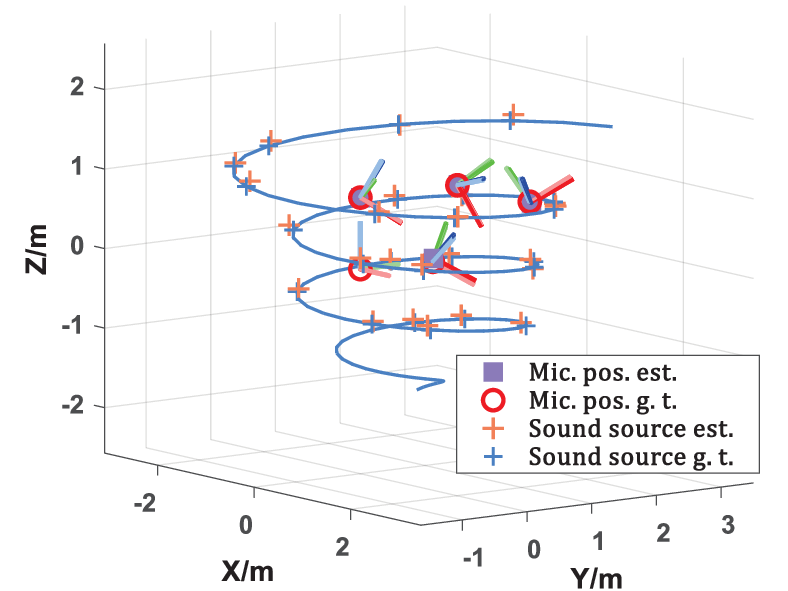}}
    \end{minipage}
        \begin{minipage}{0.265\linewidth}
        \centering
        \subfigure[]{
        \includegraphics[width=0.9\linewidth]{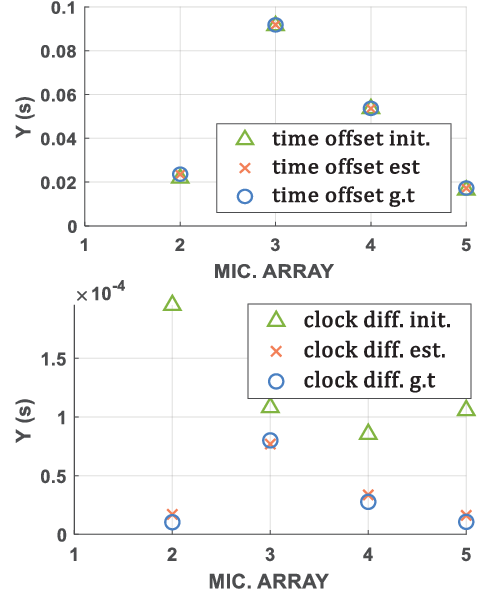}}
    \end{minipage}
    \caption{Estimation results of the preset trajectory with 5 microphone arrays and 24 sound signals. (a) The initial and the true values of microphone array positions, orientations, and sound source positions. (b) The fine-tuned and true values of microphone array positions, orientations, and sound source positions. (c) The initial, fine-tuned, and true values of microphone array time offsets and sampling clock differences between microphone arrays.}
    \label{fig:traj9}
\end{figure*}

To validate the initialization pipeline, we employed a predefined trajectory for the sound source, as illustrated in Fig. \ref{fig:traj9}. We utilized our proposed pipeline to initialize the unknown parameters. For comparison, we added varying levels of noises to the true values of the unknown parameters for the same trajectory. These noisy values were then used as initial guesses for the Gauss-Newton iterations. 

In detail, we set the base noise standard deviation for the microphone array positions, orientations, asynchronous parameters, and source positions to be $diag_3(0.2m)$, $diag_3$(10 degrees), $10^{-2}s$, $10^{-5}s$, and $diag_3(0.2m)$, respectively. 
We selected six sets of initial values, i.e., the ground truth (GT), Random, and Lv1, Lv2, Lv3, Lv4 where Gaussian noises with a standard deviation of 1, 3, 6, 9 times of the base noise are added to the GT. For these different initialization schemes, we conducted 200 Monte Carlo simulations with randomly selected initial values and used the root mean square error (RMSE) to measure the accuracy of the estimated
values (the specific formulas are provided in Appendix B).

Furthermore, we also investigated the impact of different initialization schemes on the convergence ratio of the Gauss-Newton algorithm. For each initialization scheme, we define the convergence ratio as the proportion of successful convergence instances to the total number of experiments. During the optimization process, we assessed the convergence of the Gauss-Newton algorithm based on the square norm of the optimization step size, i.e., $\left\Vert \Delta\mathbf{x}\right\Vert _{2}$. We classify any of the following three scenarios as divergent: (1) $\left\Vert \Delta\mathbf{x}\right\Vert _{2}$ exceeds 1e8 in any iteration; (2) $\left\Vert \Delta\mathbf{x}\right\Vert _{2}$ exhibits oscillations above 1e3 and does not come down below 1e3; (3) $\left\Vert \Delta\mathbf{x}\right\Vert _{2}$ keeps growing as the iteration step increases. Otherwise, the Gauss-Newton algorithm is deemed convergent.

Note that, during the numerical experiments, we keep the multiple microphone
arrays stationary while the sound source is in motion. The parameters
used in the numerical experiments are summarized in Table \ref{tab:1} (note that in practice, the DOA information can be conveniently indicated by elevation and azimuth angles in 3D. Hence, we will use the latter two angles to represent DOA in the remainder of the paper). Specifically, in our Monte Carlo simulations, the true values of all unknown parameters remain fixed, and the initial values for the Gauss-Newton iterations of each simulation are obtained as described above. Additionally, each simulation utilizes measurements with the same noise level. In other words, noises are added to the theoretical measurement values with standard deviations (STD), as shown in Table \ref{tab:1}, resulting in the final measurement values used in simulations for inter-array TDOA, DOA, and sound source relative positions.
\begin{table}
    \caption{\label{tab:2}THE RMSE OF CALIBRATION RESULTS UNDER VARYING INITIALIZATION NOISE LEVELS: ANALYSIS OF 200 MONTE CARLO SIMULATIONS (BOLD MEANS BETTER)}
        \scalebox{0.9}{\begin{tabular}{ccccccc}  
            \toprule [1pt]
            \multicolumn{1}{l}{\multirow{3}{0.5cm}{Noise Levels}} &\multicolumn{4}{c}{Microphone Array} & \multicolumn{1}{c}{SRC} & \multicolumn{1}{l}{\multirow{3}{0.5cm}{Convg. Ratio}}\\ 
            \cmidrule(lr){2-5}
            &  \makecell[c]{Pos. \\ (m)} & \makecell[c]{Orie. \\ (deg.)} & \makecell[c]{ Offset\\ (ms)}& \makecell[c]{ Clock \\ (us)} & \makecell[c]{Pos. \\ (m)} \\
            \midrule [1pt]
            GT  & \textbf{2.796e-02}	&\textbf{1.173}	&1.078e-01	&\textbf{7.579}	&\textbf{4.228e-02}
            & \textbf{100\%} \\
            \rule{0pt}{10pt}
            Ours  & 2.797e-02	&2.348	&1.078e-01	&7.584	&4.229e-02 &\textbf{100\%}\\
            \rule{0pt}{10pt}
            Lv1  & 2.973e-02	&6.299  &\textbf{0.992e-01}	&7.865	&4.475e-02 & \textbf{100\%}\\
            \rule{0pt}{10pt}
            Lv2 & 3.143e-02	&19.790 &1.348e-01	&8.730 	&4.611e-02 & 99.0\% \\
            \rule{0pt}{10pt}
            Lv3 & 6.026e-02 &42.860 &3.010e-01	&33.196 &1.011e-01 & 78.5\%
            \\
            \rule{0pt}{10pt}
            Lv4 & 7.861e-01	&64.250 &3.239	&68.573 &2.754e-01 & 44.0\%
            \\
            \rule{0pt}{10pt}
            Random & 6.928e-01 & 67.636 &2.416	&82.810  &2.635e-01 & 43.0\%
            \\
            \bottomrule [1pt]
        \end{tabular}}
\end{table}

The results are presented in Table \ref{tab:2}. It is evident that with an increase in the noise level of initialized values for the unknown parameters, the final estimation errors gradually increase (except for the time offset, where Lv1 has a negligible advantage over GT), and the convergence ratio decreases. Furthermore, it can be observed that without relying on the GT for initial guess selection, the performance of our calibration algorithm is comparable to the case using the GT as the initial value. In terms of estimating the microphone array orientation, our method is slightly less accurate compared to using the GT as the initial guess. 
This demonstrates the effectiveness of our proposed framework. In contrast, the random initialization method, frequently used in many optimization problems, exhibits inferior performance. Although it outperforms Lv4 in terms of the accuracy of some parameters, it has the lowest convergence ratio, indicating the unreliability of a random strategy. The above comparisons highlight the necessity of an appropriate initialization algorithm in the calibration process and the effectiveness of our proposed pipeline.

\subsection{Calibration Using Random Trajectories}
\begin{figure*}
    \begin{minipage}{0.36\linewidth}
        \centering
        \subfigure[]{
        \includegraphics[width=1.05\linewidth]{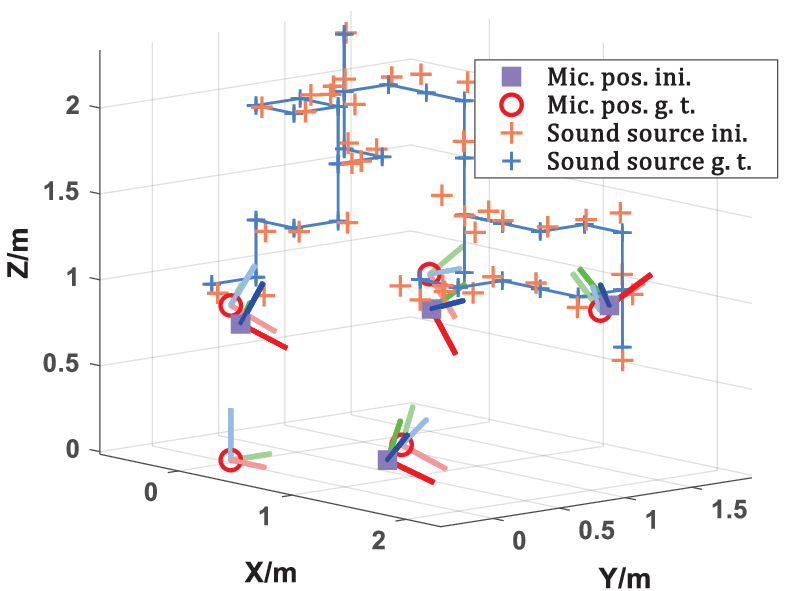}}
    \end{minipage}
    \begin{minipage}{0.36\linewidth}
        \centering
        \subfigure[]{
        \includegraphics[width=1.05\linewidth]{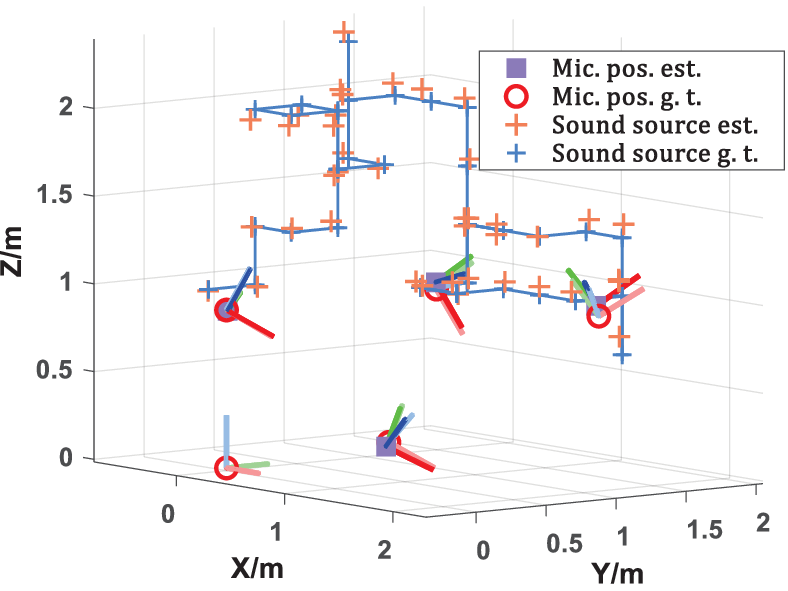}}
    \end{minipage}
    \begin{minipage}{0.26\linewidth}
        \centering
        \subfigure[]{
        \includegraphics[width=0.9\linewidth]{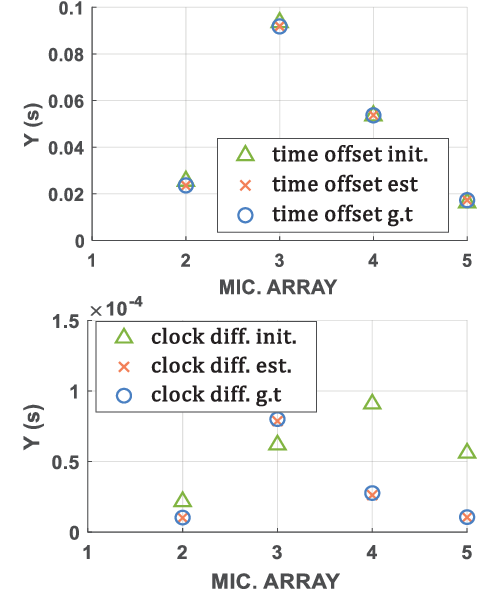}}
    \end{minipage}
    \caption{Estimation results of the random trajectory with 5 microphone arrays and 80 sound signals. (a) The initial and the true values of microphone array positions, orientations, and sound source positions. (b) The fine-tuned and true values of microphone array positions, orientations, and sound source positions. (c) The initial, fine-tuned, and true values of microphone array time offset and sampling clock differences between microphone arrays.}
    \label{fig:traj3}
\end{figure*}

To verify the robustness of the proposed calibration framework, we generate ten random trajectories, each involving five microphone arrays and 80 sound-emitting events. Take a trajectory shown in Fig. \ref{fig:traj3} as an example {(only the first 40 sound-emitting events for illustration purposes)}. Even with measurement noise interference,
the parameter initialization procedure can obtain initial values that are close to the ground truth. The initialized values are used
in Gauss-Newton iterations to improve calibration accuracy. 

\begin{figure*}[htbp]
	\centering
	\begin{minipage}{0.19\linewidth}
		\centering
		\subfigure[]{\includegraphics[width=0.95\linewidth]{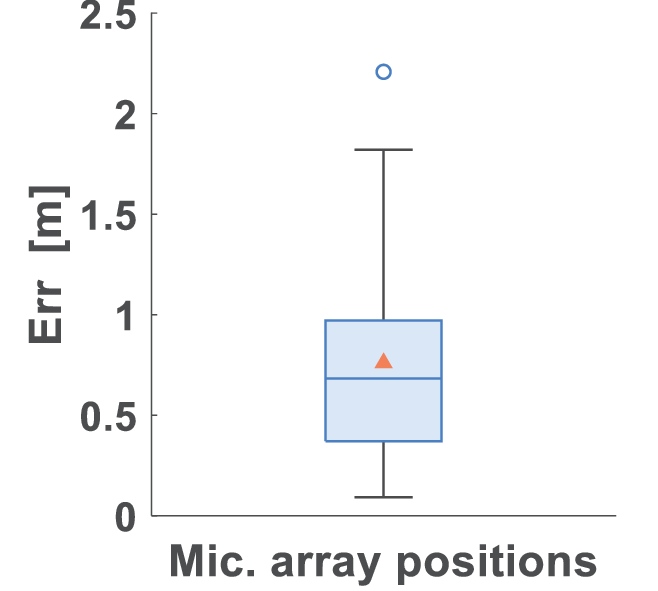}}
	\end{minipage}
	\begin{minipage}{0.19\linewidth}
		\centering
		\subfigure[]{\includegraphics[width=0.95\linewidth]{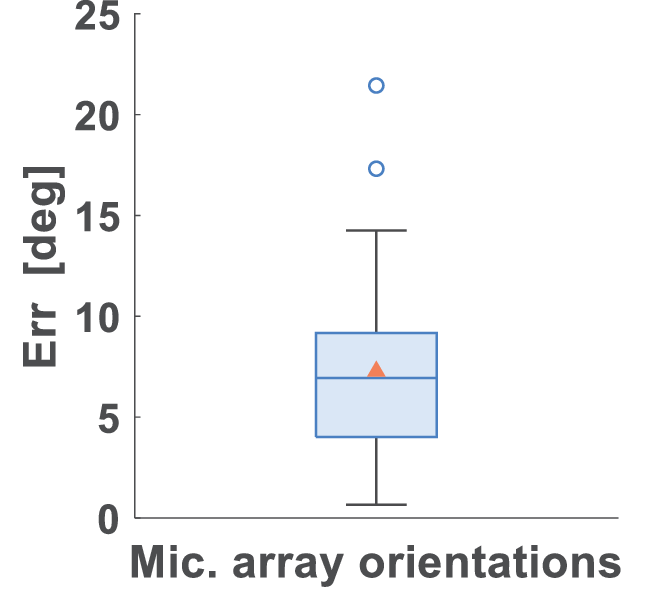}}
	\end{minipage}
	\begin{minipage}{0.19\linewidth}
		\centering
		\subfigure[]{\includegraphics[width=0.95\linewidth]{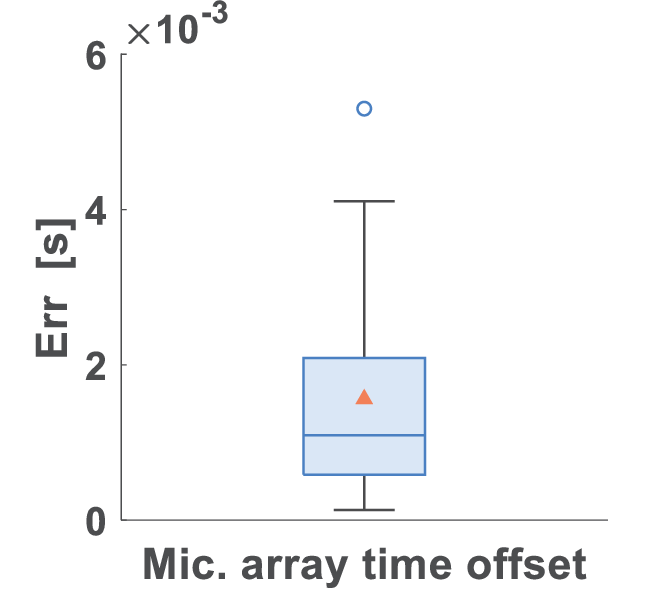}}
	\end{minipage}
	\begin{minipage}{0.19\linewidth}
		\centering
		\subfigure[]{\includegraphics[width=0.95\linewidth]{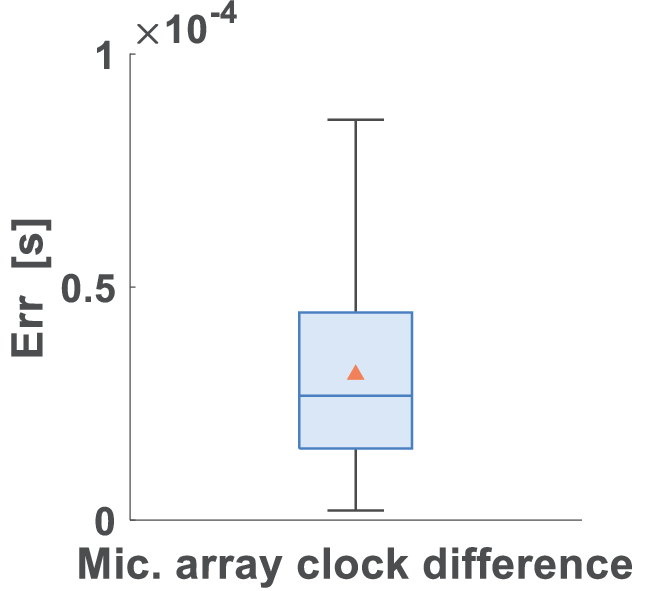}}
	\end{minipage}
	\begin{minipage}{0.19\linewidth}
		\centering
		\subfigure[]{\includegraphics[width=0.95\linewidth]{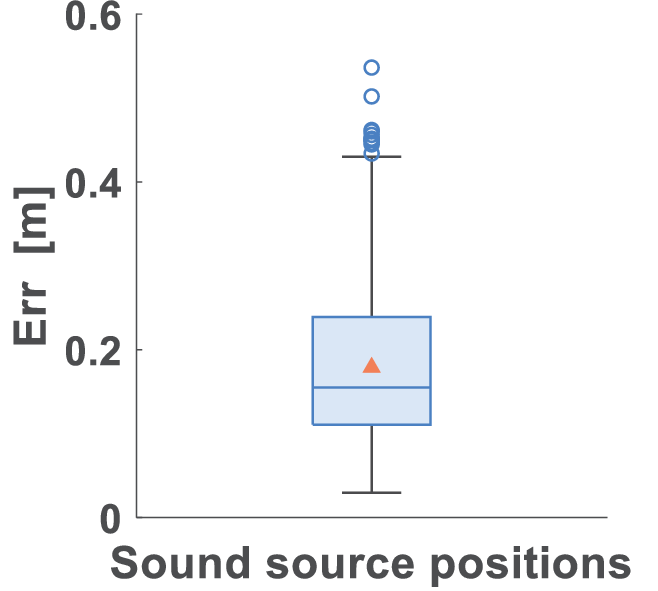}}
	\end{minipage}
	\caption{Error distributions between the initial values and true values for 10 different trajectories. (a) Microphone array positions. (b)  Microphone array orientations. (c) Time offsets. (d) Sampling clock differences. (e) Sound source positions.} 
    \label{fig:6}
\end{figure*}

Fig. \ref{fig:6}
shows the error distribution between the initialized values obtained
by our proposed initialization method and the ground truth for
ten different trajectories. In the box plot, the blue circle represents
the outliers obtained from the interquartile range, while the upper
and lower black horizontal lines represent the maximum and minimum
values of the non-outlier errors. The upper and lower edges of each
box represent the upper and lower quartiles, respectively, and the
middle blue line corresponds to the median of the errors. The orange
triangle represents the mean of the errors. The errors between our
initial values and the ground truth are small, which promotes the convergence
of the calibration algorithm. 

\begin{figure*}[htbp]
	\centering
	\begin{minipage}{0.19\linewidth}
		\centering
		\subfigure[]{\includegraphics[width=0.95\linewidth]{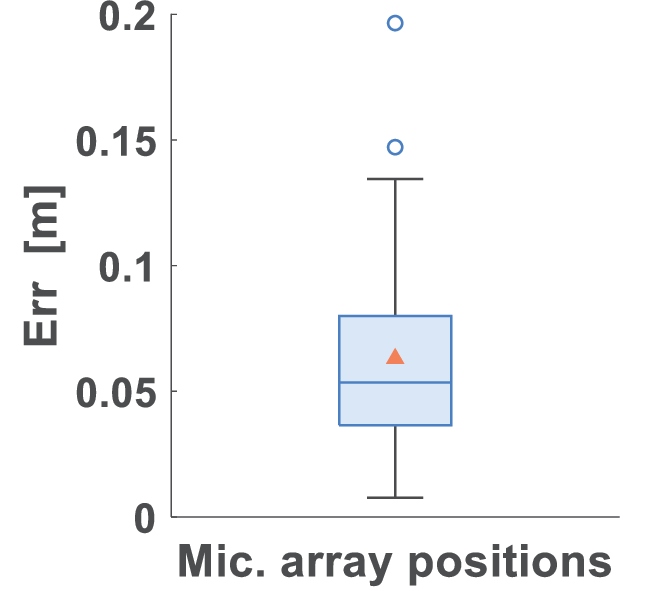}}
	\end{minipage}
	\begin{minipage}{0.19\linewidth}
		\centering
		\subfigure[]{\includegraphics[width=0.95\linewidth]{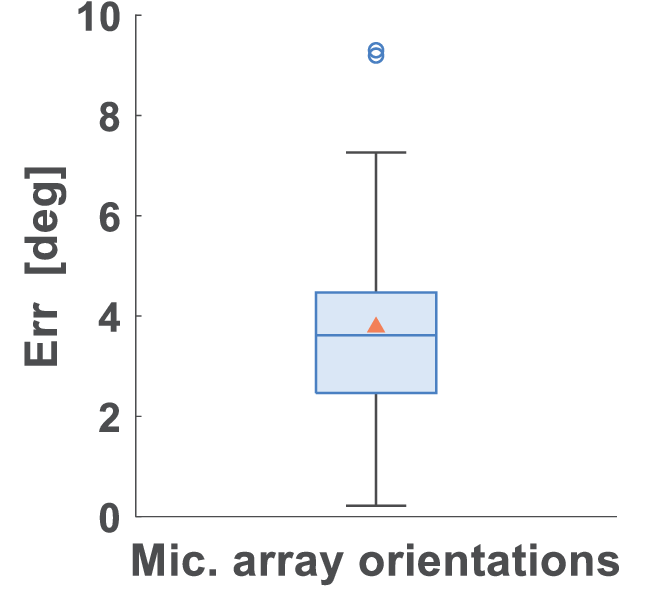}}
	\end{minipage}
	\begin{minipage}{0.19\linewidth}
		\centering
		\subfigure[]{\includegraphics[width=0.95\linewidth]{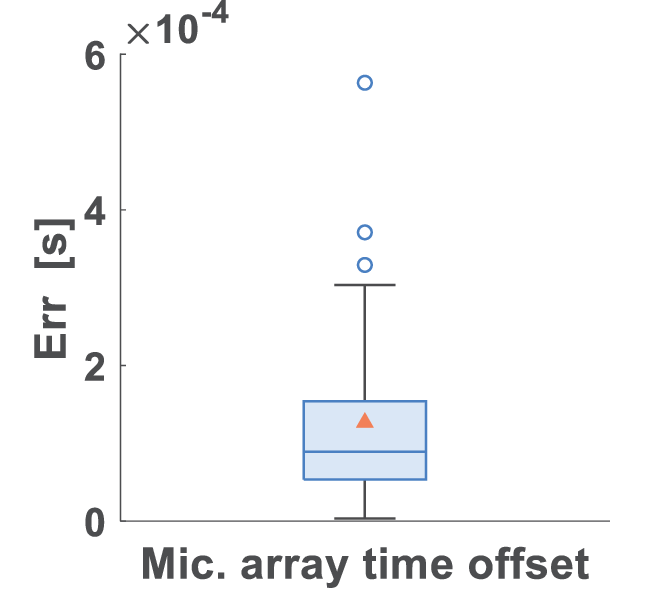}}
	\end{minipage}
	\begin{minipage}{0.19\linewidth}
		\centering
		\subfigure[]{\includegraphics[width=0.95\linewidth]{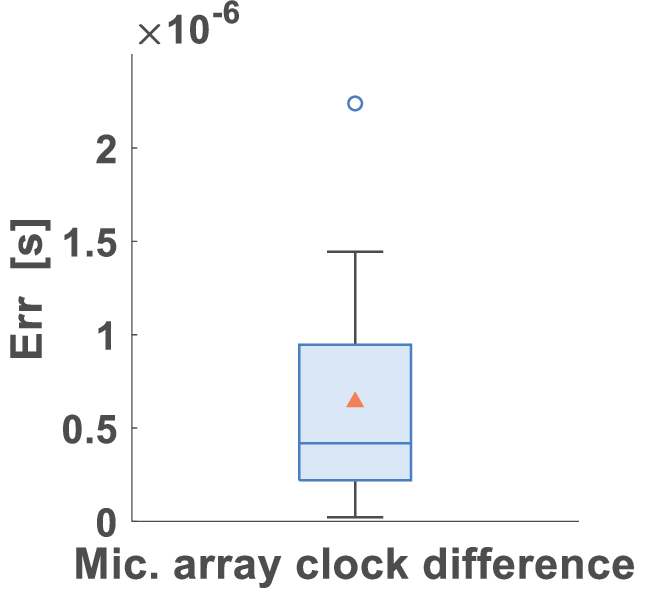}}
	\end{minipage}
	\begin{minipage}{0.19\linewidth}
		\centering
		\subfigure[]{\includegraphics[width=0.95\linewidth]{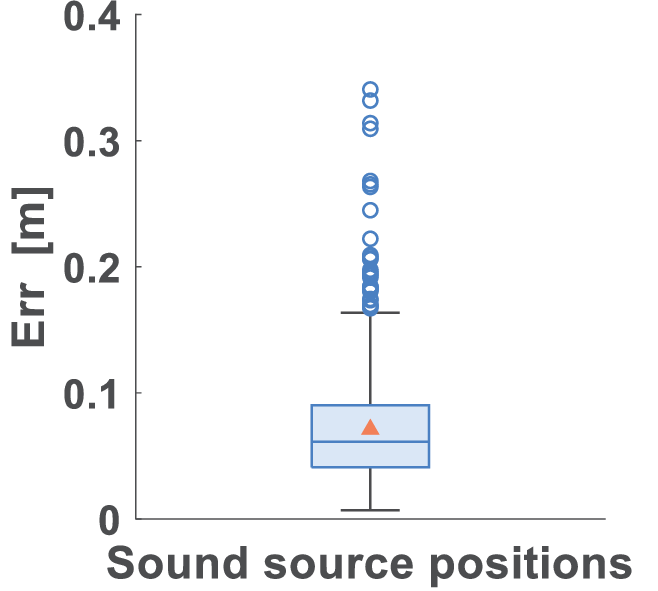}}
	\end{minipage}
	\caption{Error distributions between the estimated values and true values for 10 different trajectories. (a) Microphone array positions. (b)  Microphone array orientations. (c) Time offsets. (d) Sampling clock differences. (e) Sound source positions.}
    \label{fig:7}
\end{figure*}

Fig. \ref{fig:7} shows the error distribution between the final estimated
values and the ground truth for ten different trajectories. Similar to Table \ref{tab:2}, the results indicate that while the accuracy of the microphone array orientation estimation is slightly lower than that of other parameters due to the larger DOA measurement noise, the calibration of all parameters is accurate.

Finally, we remark that the relatively poorer accuracy for microphone array orientation shown in Table \ref{tab:2} and Fig. \ref{fig:7} is mainly attributed to the large magnitude of DOA
measurement noise used in the simulation. As indicated in Table \ref{tab:1}, for our simulations, the elevation and azimuth angle noise STD are both 5°. If we reduce the elevation and azimuth angle noise STD, the accuracy of microphone array orientation will be improved. However, due to limited space, we skip these comparisons and results here.


\begin{figure}[t]
	\centering
	\begin{minipage}{0.415\linewidth}
		\centering
		\subfigure[]{\includegraphics[width=1\linewidth]{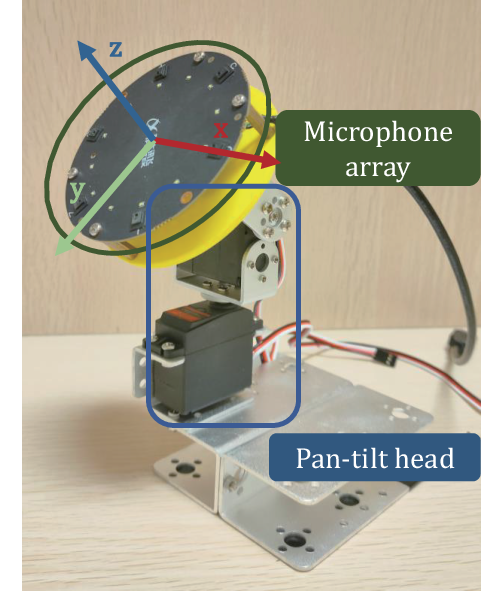}}
	\end{minipage}
	\begin{minipage}{0.53\linewidth}
		\centering
		\subfigure[]{\includegraphics[width=1.04\linewidth]{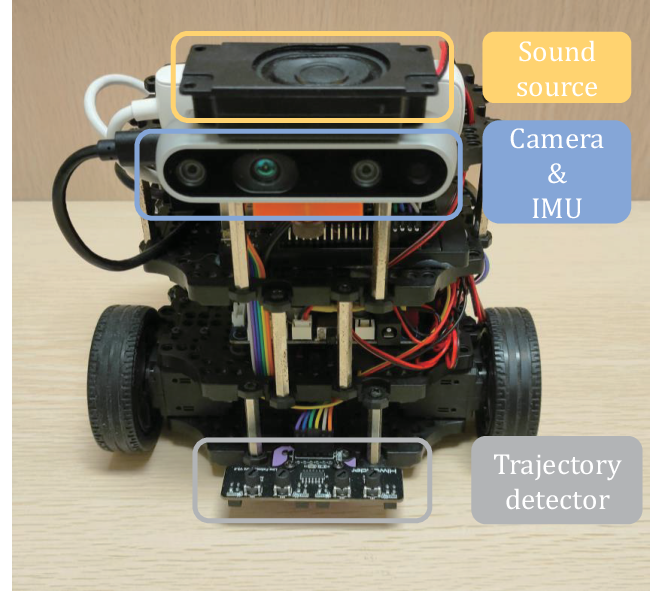}}
	\end{minipage}

    \begin{minipage}{0.93\linewidth}
    \centering
    \subfigure[]{\includegraphics[width=1.05\linewidth]{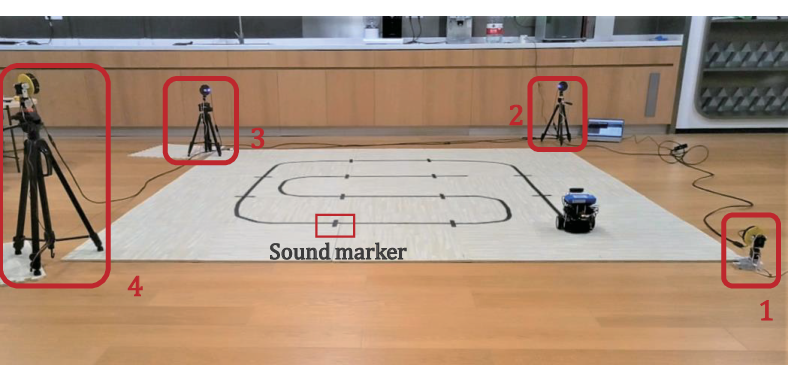}}
    \end{minipage}
    	\caption{Real-world 3D asynchronous microphone arrays calibration environment setup. (a) Microphone array with pan-tilt head. (b) Turtlebot3 robot with Multi-sensors. (c) Typical physical scenario. }
    \label{fig:exp_scene}
\end{figure}

\begin{figure}[t]
	\centering
	\begin{minipage}{0.75\linewidth}
		\centering
		\subfigure[]{\includegraphics[width=1.1\linewidth]{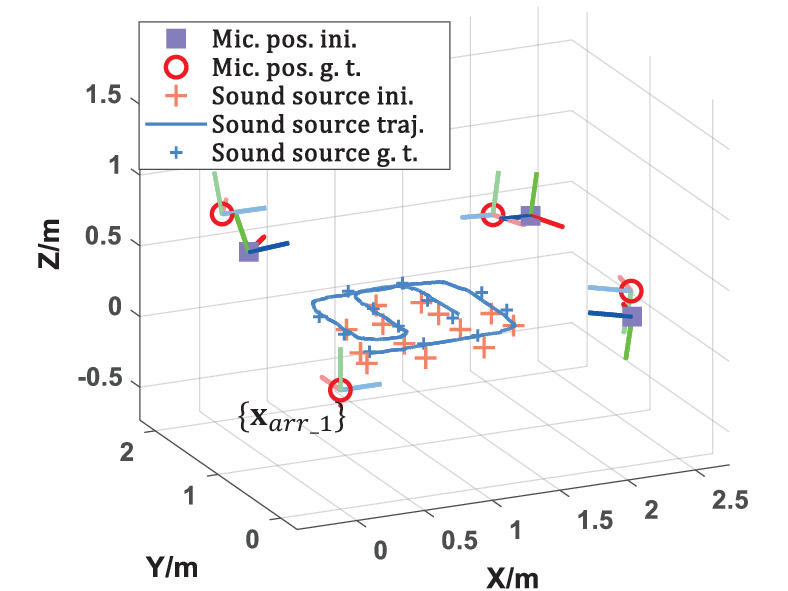}}
	\end{minipage}

	\begin{minipage}{0.75\linewidth}
		\centering
		\subfigure[]{\includegraphics[width=1.1\linewidth]{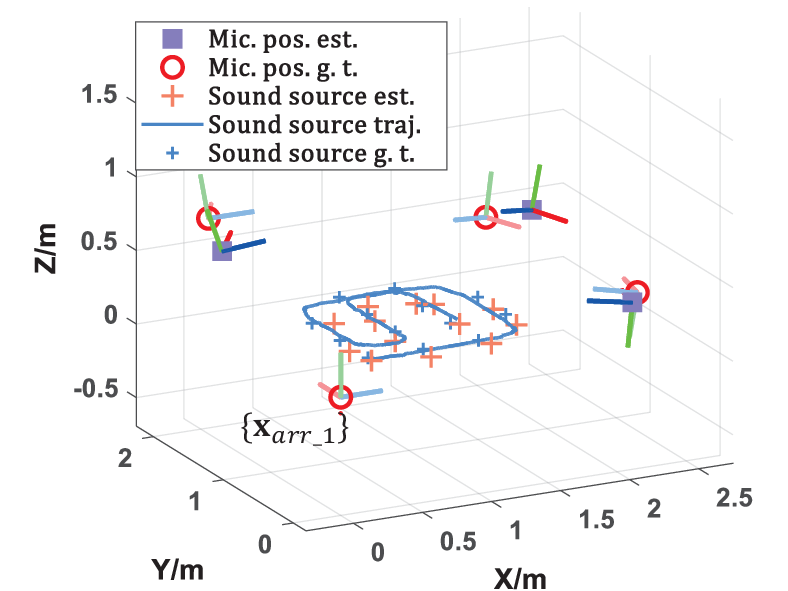}}
	\end{minipage}

    \begin{minipage}{0.49\linewidth}
		\centering
		\subfigure[]{\includegraphics[width=1\linewidth]{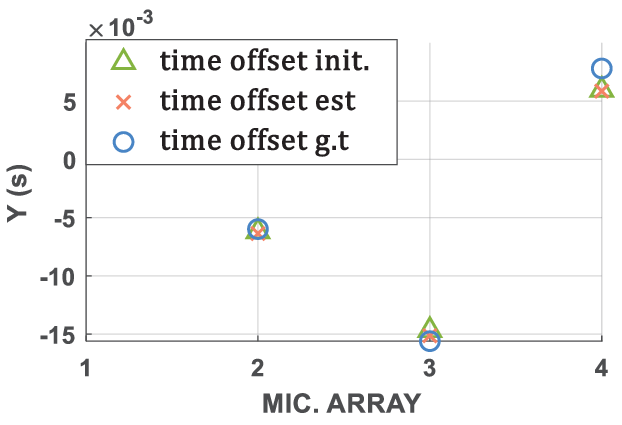}}
	\end{minipage}
	\begin{minipage}{0.49\linewidth}
		\centering
		\subfigure[]{\includegraphics[width=1\linewidth]{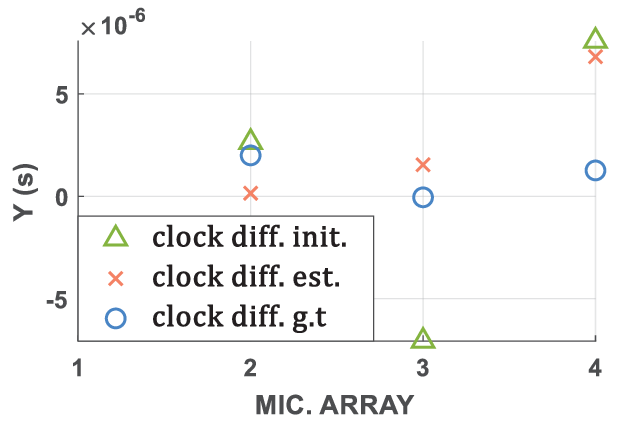}}
	\end{minipage}
    	\caption{Real environment microphone arrays calibration results. (a) The initial and the true values of microphone arrays positions, orientations, and sound source positions. (b) The fine-tuned and true values of microphone arrays positions, orientations, and sound source positions. (c) The calibration result of microphone arrays time offsets (e) The calibration result of microphone arrays sampling clock differences.}
    \label{fig:8}
\end{figure}

\section{\label{real}REAL-WORLD EXPERIMENTS}
{In this section, we validate our calibration method using real data. In our experiment, a Turtlebot3 mobile robot moves in an indoor environment, and multiple microphone arrays capture the sound signal emitted by the robot. More specifically, we use the iFLYTEK M160C microphone array consisting of six independent microphones arranged in a circular and evenly distributed configuration, with a diameter of 70.85mm, a sampling depth of 32 bits, a sampling rate of 16 KHz, and an effective pickup range of 3.5 meters}, as shown in Fig. \ref{fig:exp_scene}(a).  The mobile robot is equipped with an Intel D435i camera with an integrated inertial measurement unit (IMU), as shown in Fig. \ref{fig:exp_scene}(b). It also includes a four-channel trajectory detector for tracking predefined paths and a 3W 8$\Omega$ speaker for sound emission.

In the experimental setup, four microphone arrays are placed in an open area of an academic building. The experimental area is 15.5 meters long, 10 meters wide, and 3.3 meters high, as shown in Fig. \ref{fig:exp_scene}(c). The microphone arrays remain stationary and receive audio signals while the mobile robot travels along the black trajectory on the ground. When the robot detects the cross-shaped sound markers on the ground, it immediately emits a chirp signal with a frequency of 1000 Hz to 2000 Hz through a speaker driven by a Class-D amplifier, lasting for 300 ms, and then moves on the trajectory. We carry out the following activities to validate the effectiveness and performance of the proposed calibration pipeline across diverse scenarios:

{1) Firstly, we compare the calibration results achieved through various initialization strategies (see Section VI.B).} 

2) Next, we explore the influence of the absence of sound source relative position measurements on the estimation accuracy in the optimization process (see Section VI.C).

{3) Moreover, we vary the spacing between the microphone arrays to cover a range of scenarios and scene scales to assess the calibration performance of our method (see Section VI.D).}

{4) Last but not least, in Section VI.E, we compare the performance of the proposed initialization method (IM) and its fine-tuning (FT) version (i.e., the results are obtained by feeding the initialized values to batch optimization with Gauss-Newton iterations) with those of other existing methods, including the open-source passive geometry calibration method for microphone arrays based on the differential evolution algorithm (PGM) \cite{Plinge2017} and the two-step calibration method (TSM) based on the L-BFGS algorithm \cite{Wozniak 2019}.}

\subsection{{Data Collection and Ground Truth}}
{The trajectory of the robot is pre-defined to obtain the ground truth of sound source positions in the global frame. The position of the speaker during audio playback, corresponding to the sound marker's coordinates and the mobile robot's height, is regarded as the ground truth for the sound source positions. The microphone arrays were placed w.r.t. each other according
to known preset values (i.e., these are taken to be the true values of microphone array positions) before the experiment started. The frame $\left\{ \mathrm{\mathbf{x}_{\mathit{arr\_1}}}\right\} $
attached to the first
microphone array is taken to be the global coordinate system. As shown in Fig. \ref{fig:exp_scene}(a), we affix each microphone array to a pan-tilt head. Subsequently, the pan-tilt head attached to each microphone array (except the first one) is rotated by certain known pre-set angles which are used to calculate the ground truth value of the Euler angles of the microphone arrays.}

{We determine the GT values of time offset and sampling clock difference as follows. To compute the ground truth for time offset, we subtract the theoretical time difference (excluding time offset and sampling clock difference) between the first sound marker received by each microphone array and the reference microphone array from the actual time difference. Note that due to the robot's quick arrival at the first sound marker, the clock difference is so small at this moment that it can be considered negligible. For clock difference, we have recorded 8 hours of audio using multiple microphone arrays placed at the same distance relative to the sound source. This recording includes start and end signals. We calculate the ground truth for sampling clock difference by comparing the number of samples recorded by each microphone array with that of the reference microphone array during this period.}

{The following three kinds of measurements are obtained during the experiment:}

{1) For inter-array TDOA measurements between any microphone array and the reference microphone array at the \textit{k-th} sound marker, we employ a sliding window technique to break down the sound signal into short frames. Subsequently, we compute the power spectrum of each frame to determine the valid sound region. Each frame has a duration of 25 ms, and a Hamming window is applied to prevent spectral leakage. For the valid sound region, we apply the GCC-PHAT algorithm \cite{GCC1976}, widely used in robotic sound localization, to compute the inter-channel time differences for all combinations of 6 channels $\times$ 6 channels. The average time difference is calculated as the inter-array TDOA.}

{2) For DOA measurements of the microphone array, we employ the Steered Response Power-Phase Transform (SRP-PHAT) algorithm \cite{srp} on the obtained signal region, with a discrete search angle resolution of 3 degrees.} This technique leverages the spatial filtering capability of the microphone array to estimate the received power from a set of candidate directions. The source is then identified by selecting the location associated with the highest energy. The estimated azimuth and elevation angles are subsequently transformed into three-dimensional unit direction vectors.

{3) For the sound source relative position measurements, we utilize a visual-inertial odometry (VIO) method \cite{Shen2018} that integrates camera and IMU data. This approach fuses visual information and inertial data, providing more accurate and robust displacement measurements. This allows us to integrate more measurements related to robot motions, thereby enhancing the accuracy and reliability of the sound source relative position measurements.}

\begin{table}
    \caption{\label{tab:real init}THE RMSE OF CALIBRATION RESULTS UNDER VARYING INITIALIZATION NOISE LEVELS USING REAL DATA: ANALYSIS OF 200 MONTE CARLO EXPERIMENTS (BOLD MEANS BETTER)}
    \begin{center}
        \scalebox{1.0}{\begin{tabular}{ccccccc}  
            \toprule [1pt]
            \multicolumn{1}{l}{\multirow{3}{0.5cm}{Noise Levels}} &\multicolumn{4}{c}{Microphone Array} & \multicolumn{1}{c}{SRC} & \multicolumn{1}{l}{\multirow{3}{0.5cm}{Convg. Ratio}}\\ 
            \cmidrule(lr){2-5}
            &  \makecell[c]{Pos. \\ (m)} & \makecell[c]{Orie. \\ (deg.)} & \makecell[c]{ Offset\\ (ms)}& \makecell[c]{ Clock \\ (us)} & \makecell[c]{Pos. \\ (m)} \\
            \midrule [1pt]
            GT  & \textbf{0.233} 	&\textbf{7.936} 	&\textbf{1.514} 	&\textbf{12.712} 	&\textbf{0.156}
            & \textbf{100.0\%} \\
            \rule{0pt}{10pt}
            Ours  & \textbf{0.233} 	& 9.650	& 1.515 	& 12.749 	& \textbf{0.156}  &\textbf{100.0\%}\\
            \rule{0pt}{10pt}
            Lv1  & \textbf{0.233} 	&8.291 	&1.521 	&12.713 	&\textbf{0.156}             & 99.87\%\\
            \rule{0pt}{10pt}
            Lv2 & 0.561 	&10.511 	&2.915 	&12.898 	&0.179  & 39.30\% \\
            \rule{0pt}{10pt}
            Lv3 & 1.068 	&34.799 	&4.886 	&13.819 	&0.419
            & 3.53\%
            \\
            \rule{0pt}{10pt}
            Random & 0.839 & 78.303 &20.730	& 56.709 &0.775 & 0.10\%
            \\
            \bottomrule [1pt]
        \end{tabular}}
    \end{center}
\end{table}
\subsection{{Comparisons between Different Initialization Methods}}
{For the case when four microphone arrays are placed on the corners of a square (2m $\times$ 2m), we collect data for five different trajectories, each repeated three times, resulting in a total of 15 datasets. These collected datasets have been used to explore the impact of different initial values in real-world experiments.}

{Based on the GT and measurement models in Section II (see (\ref{measurements})-(\ref{random_walk})), we then calculate the following measurement errors: the inter-array TDOA measurement error has mean value 3.15e-4 seconds with STD of 1.25e-3; the azimuth angle error has mean value of 6.02 degrees with STD of 4.69 degrees; the elevation angle error has mean value of 5.45 degrees with STD of 5.97 degrees, and the VIO measurement error has mean value [2.06e-2, 2.49e-2, 6.13e-3] meters with STD of [9.64e-3, 3.65e-2, 8.44e-3]  meters. These errors were obtained by comparing the measured values from the sound signal with the theoretical values.}

{We set the base noise standard deviation for the microphone array positions, orientations, asynchronous parameters, and source positions to be $diag_3(0.2m)$, $diag_3$(10 degrees), $10^{-2}s$, $10^{-5}s$, and $diag_3(0.2m)$, respectively. Subsequently, for comparison, we obtain initial values with different levels of errors (GT, Lv1, Lv2, Lv3, and Random), similar to those described in Section V.C. Using each of the 15 datasets, for these different initialization schemes, we conducted 200 Monte Carlo experiments with randomly selected initial values (note that for our proposed initialization method and the initialization using GT, there is only one experiment) and the corresponding real measurements.} The results using the above different initialized values are shown in Table \ref{tab:real init} and the calibration results of our method for one of the datasets are depicted in Fig. \ref{fig:8}.

{From Table \ref{tab:real init}, it can be observed that for real data, the overall calibration accuracy is slightly lower compared to that of simulation studies, primarily due to noise sources such as motion noise from the mobile robot, sensor measurement noise, and manual interference. However, the effectiveness of our proposed method is evident. In real-world settings, our initialization method produces calibration results almost identical to those obtained using ground truth values directly or Lv1 as initial values, with only slightly reduced orientation accuracy. This is because, unlike simulations, the measurement
noises in real-world settings are, in general, not Gaussian and the accuracy of DOA
and TDOA measurements is lower. Notably, the noises in DOA measurements (w.r.t. the
ground truth values) almost overshadow the performance differences between the initialization
methods GT, Lv1, and our initial values. Despite that, the results indicate the effectiveness of our method in scenarios with non-Gaussian and large measurement noises.

Regarding convergence ratio, both our initialization method and direct use of ground truth values achieve 100\% convergence. As the initial noise level increases, the convergence ratio of initialization using noise-corrupted GT values significantly decreases, especially in cases with higher levels of noise, such as the Lv3 and random initialization methods, where all Monte Carlo experiments across all 15 datasets almost always diverge. This underscores the effectiveness and robustness of our proposed method in real-world scenarios.}
\begin{table}
    \caption{\label{tab:without vio}THE RMSE OF CALIBRATION RESULTS UNDER VARYING
INITIALIZATION NOISE LEVELS WITH ONLY ACOUSTIC MEASUREMENTS: ANALYSIS OF 200 MONTE CARLO EXPERIMENTS (BOLD MEANS BETTER)}
    \begin{center}
        \scalebox{1.0}{\begin{tabular}{ccccccc}  
            \toprule [1pt]
            \multicolumn{1}{l}{\multirow{3}{0.5cm}{Noise Levels}} &\multicolumn{4}{c}{Microphone Array} & \multicolumn{1}{c}{SRC} & \multicolumn{1}{l}{\multirow{3}{0.5cm}{Convg. Ratio}}\\ 
            \cmidrule(lr){2-5}
            &  \makecell[c]{Pos. \\ (m)} & \makecell[c]{Orie. \\ (deg.)} & \makecell[c]{ Offset\\ (ms)}& \makecell[c]{ Clock \\ (us)} & \makecell[c]{Pos. \\ (m)} \\
            \midrule [1pt]
            GT  & \textbf{0.425} 	&\textbf{11.580} 	&2.015	&12.064 	&\textbf{0.226} 
            & \textbf{100.0\%} \\
            \rule{0pt}{10pt}
            Lv1  & 0.426 	&12.819 	&\textbf{2.012} 	&\textbf{12.062} 	&\textbf{0.226 }             & 99.90\%\\
            \rule{0pt}{10pt}
            Lv2 & 0.658 	&14.270 	&3.100 	&12.185 	&0.231  & 32.07\% \\
            \rule{0pt}{10pt}
            Lv3 & -- 	&-- 	&-- 	&-- 	&--
            & 0.0\%
            \\
            \rule{0pt}{10pt}
            Random & -- & -- &--	&--  &-- & 0.0\%
            \\
            \bottomrule [1pt]
        \end{tabular}}
    \end{center}
\end{table}
\subsection{Calibration with Only Acoustic Measurements}
To validate the influence of the sound source relative position measurements obtained from the VIO method on the calibration results, this section focuses on conducting calibration experiments using only acoustic measurements obtained from the microphone arrays (inter-array TDOA and DOA measurements). Given that our initialization method relies on relative position measurements of the sound source, we cannot use it for comparison purposes. Hence, we perform Gauss-Newton optimizations initialized by ground truth values corrupted by Gaussian noise across varying levels. Following Section VI.B, Monte Carlo experiments are carried out under varying initialization noise levels using the real measurements from the 15 datasets (excluding the sound source relative position measurements from VIO).

{Comparing the results in Table \ref{tab:real init} and Table \ref{tab:without vio} (including and excluding the sound source relative position measurements, respectively), it is evident that without the sound source relative position measurements, the overall parameter estimation results are poorer. In particular, the estimation accuracy of the relative transforms (i.e., orientation, translation) between microphone
arrays and sound source positions, and the convergence ratio are lower than the case with sound source relative position measurements. However, as it can also be seen from Table \ref{tab:real init} and Table \ref{tab:without vio}, the absence of sound source relative position measurements from VIO has less impact on the estimation accuracy of asynchronous parameters between the microphone arrays.

\begin{figure}[t]
	\centering
    \begin{minipage}{0.8\linewidth}
		\centering
		\subfigure[]{\includegraphics[width=1\linewidth]{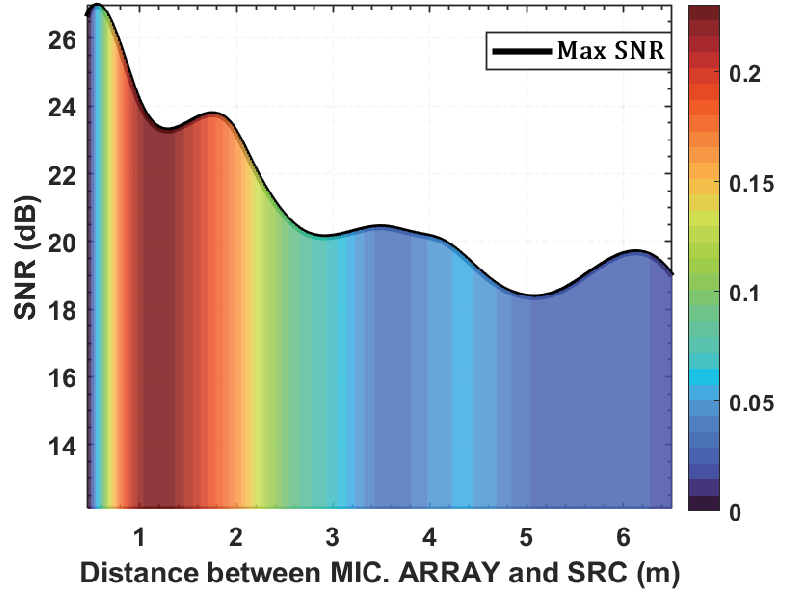}}
	\end{minipage}

	\begin{minipage}{0.8\linewidth}
		\centering
		\subfigure[]{\includegraphics[width=1\linewidth]{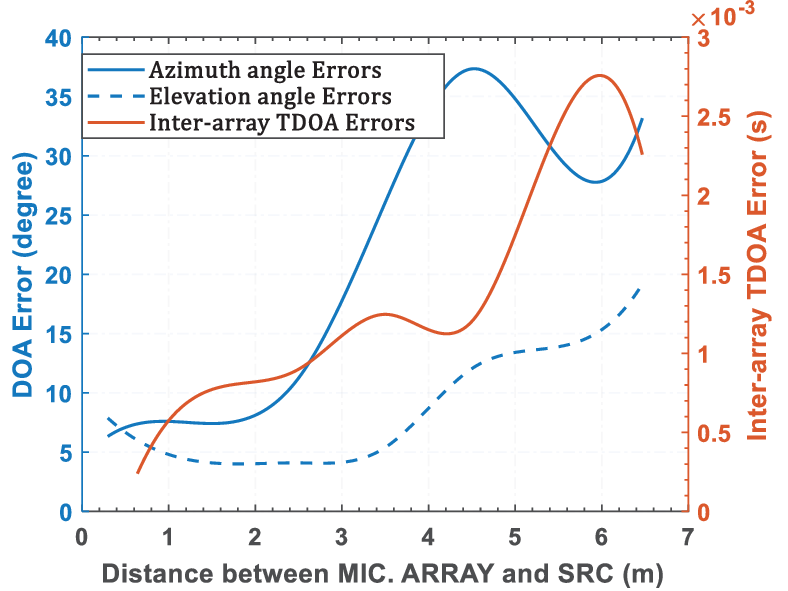}}
	\end{minipage}
    	\caption{Distance's impact on sound perception. (a) The relationship between the distance of the sound source relative to the microphone array and maximum SNR; 
     the color map illustrates the ratio of having a certain distance (the x-axis) between the sound source and the microphone arrays during the whole experiment process, across the 12 real datasets. (b) The relationship between sound source distance relative to the microphone array and DOA estimation errors, as well as inter-array TDOA estimation errors.}
     
    \label{fig:Varied scales}
\end{figure}
\begin{table}
    \caption{\label{tab:scale compare}
{THE RMSE OF CALIBRATION RESULTS UNDER VARYING SCENE SCALES IN REAL-WORLD EXPERIMENTS (BOLD MEANS BETTER)}}
    \begin{center}
            \begin{tabular}{cccccc}  
            \toprule [1pt]
            \multicolumn{1}{l}{\multirow{3}{0.8cm}{Distance}} &\multicolumn{4}{c}{Microphone Array} & \multicolumn{1}{c}{SRC}\\
            \cmidrule(lr){2-5}
            &  \makecell[c]{Positions \\ (m)} & \makecell[c]{Orientations \\ (deg.)} & \makecell[c]{ Offset\\ (ms)}& \makecell[c]{ Clock diff.\\ (us)} & \makecell[c]{Position \\ (m)} \\
            \midrule [1pt]
            1 m  & 0.197 	&7.381 	&\textbf{0.847} 	&\textbf{2.956} 	&\textbf{0.071} 
            \\
            \rule{0pt}{10pt}
            2 m  & \textbf{0.173} 	&\textbf{5.825} 	&1.272 	&4.327 	&0.116
            \\
            \rule{0pt}{10pt}
            3 m & 0.618 & 55.691 & 1.251 & 13.817 & 0.240
            \\
            \rule{0pt}{10pt}
            5 m & 2.557 	&81.966 	&3.811 	&20.712 	&0.253 
            \\
            \bottomrule [1pt]
        \end{tabular}
    \end{center}
\end{table}
\subsection{Calibration Across Varied Scene Scales}
{In this section, we conducted experiments to investigate the influence of the distances between the microphone arrays and the sound source on the calibration results. This factor plays a pivotal role in the calibration process of microphone arrays, as it impacts the propagation of sound signals and the level of measurement noises. For instance, in scenarios involving long-distance sound propagation, sound signals undergo propagation loss and are subject to noise interference, resulting in signal attenuation and a decrease in the signal-to-noise ratio (SNR) \cite{Rascon C2017}. We consider four scenarios with different microphone array spacings: 1 meter, 2 meters, 3 meters, and 5 meters. Under these varying distances, the Turtlebot3 robot moves in proximity to the microphone arrays, emitting chirp signals with consistent sound intensity. We record data for each setup, with each experiment repeated three times, resulting in a
total of 12 datasets.}

{Using the datasets collected with different microphone array spacings (1 meter, 2 meters, 3 meters, and
5 meters, respectively), we calculate the SNR that the microphone arrays can capture, and the ground truth values of the microphone arrays positions
and the sound markers positions in the global frame are directly measured using a rangefinder. Subsequently, the distances from the microphone arrays to the sound source at different sound marker positions in the 12 collected datasets could be easily calculated.
In Fig. \ref{fig:Varied scales}(a), it can be observed that there is a significant decrease in the maximum SNR that the microphone array can capture as the distance between the mobile robot and the microphone arrays gradually increases. The color map illustrates the ratio of having a certain distance (the
x-axis of Fig. 10(a)) between the sound source and the microphone arrays,
during the whole experiment process, across the 12 real datasets\footnote{For example, in our experiments, there are 4 microphone arrays; for every dataset, there were 13 sound events; so in total, there are $4*12*13 = 624$ scenarios; if there are 138 scenarios where the sound source is 1m--1.5m away from any microphone array, then its corresponding ratio is $138/624 \sim 0.22$.}. Meanwhile, Fig. \ref{fig:Varied scales}(b) clearly shows a significant decrease in the accuracy of both DOA and inter-array TDOA estimations with the increasing distance between the sensor array and the signal source. Table \ref{tab:scale compare} summarizes the calibration results for different spacing cases. It can be seen from Table \ref{tab:scale compare} that, compared to the greater spacings of 3 meters and 5 meters, our proposed calibration pipeline
achieves better performance
for the spacing cases of 1 meter and 2 meters. Moreover, one can also notice from Table \ref{tab:scale compare} that the calibration results for microphone array positions and orientations
at 2m were better than those at 1m, because the SNR increases in the 1-2m range (see Fig. \ref{fig:Varied scales}),
while the elevation angle measurement error (w.r.t. the ground truth values) gradually decreases
and the azimuth angle measurement error (w.r.t. the ground truth values) almost stays the same.
The above results further illustrate the impact of distance on calibration performance.}

\begin{table}
    \caption{\label{tab:method compare}THE RMSE OF CALIBRATION RESULTS FROM DIFFERENT METHODS IN
    REAL-WORLD EXPERIMENTS (BOLD MEANS BETTER)}
    \begin{center}
        \scalebox{1.0}{\begin{tabular}{ccccccc}  
            \toprule [1pt]
            \multicolumn{1}{c}{\multirow{3}{0.5cm}{Method}} &\multicolumn{4}{c}{Microphone Array} & \multicolumn{1}{c}{SRC} & \multicolumn{1}{l}{\multirow{3}{0.5cm}{Average Time
            (s/dataset)}}\\ 
            \cmidrule(lr){2-5}
            &  \makecell[c]{Pos. \\ (m)} & \makecell[c]{Orie. \\ (deg.)} & \makecell[c]{ Offset\\ (ms)}& \makecell[c]{ Clock \\ (us)} & \makecell[c]{Pos. \\ (m)} \\
            \midrule [1pt]
            PGM \cite{Plinge2017}  & 1.589 & 45.083 & -- & -- & -- & 3661.152 \\
            \rule{0pt}{10pt}
            TSM \cite{Wozniak 2019}  & 1.227 	& 47.461 	& 1.671 	& --	& 1.027  &48.064\\
            \rule{0pt}{10pt}
            IM (Our)  & 0.378 	&11.730 	&1.896 	&18.334 	&0.219  & 6.770
            \\
            \rule{0pt}{10pt}
            FT (Our) & \textbf{0.233} 	& \textbf{9.650 }	& \textbf{1.515} 	& \textbf{12.749} 	& \textbf{0.156}  &2.892\\
            \bottomrule [1pt]
        \end{tabular}}
    \end{center}
\end{table}
\subsection{{Comparisons with Existing Methods}}
{We next compare our proposed calibration pipeline with the existing algorithms using the datasets collected in Section VI.B. These algorithms include the passive geometry calibration method\footnote{The original algorithms in \cite{Plinge2017} is for the 2D case. For comparison purposes, we have revised it accordingly for the 3D case.} for microphone arrays based on the differential evolution algorithm (PGM) \cite{Plinge2017} and the two-step calibration method based on the L-BFGS algorithm (TSM) \cite{Wozniak 2019}.} {It is worth noting that these two calibration methods do not incorporate relative position measurements, i.e., they overlook the constraints among the positions of the sound source. Additionally, PGM does not include the calibration of time offsets and sampling clock differences among microphone arrays, while TSM disregards sampling clock differences, and the above methods lack an effective initialization process.}

{To showcase the efficiency of each calibration algorithm, we measure the average time required for each of the 15 calibration datasets for different methods on a PC with 32 GB RAM and an Intel Core 3.1 GHz i5-10505 processor. Table \ref{tab:method compare} provides a summary of quantitative comparisons, where the RMSE is calculated based on the metrics listed in Appendix B. The experimental results indicate that our proposed methods (both IM and FT) outperform both PGM and TSM. Besides, the proposed method takes approximately 9 seconds (the total time that both initialization and Gauss-Newton iteration take) to automatically generate a highly accurate calibration of the multiple microphone arrays in 3D, which is faster than TSM and PGM, demonstrating its desirable efficiency.}

\subsection{{Discussions}}
{It is evident from the previous simulation and experimental results that the proposed method demonstrates strong robustness, outperforming existing calibration methods in terms of both accuracy and speed. Moreover, one should note that calibration accuracy is influenced by the measurement noises of the sensors, which is a critical factor. It is also worth noting that \noindent the SNR decreases as distances increase between the sound source and the microphone arrays, as pointed out in the existing works \cite{AnTRO}, \cite{Evers2018}, and \cite{Valin2007}. Consequently, calibration accuracy gradually diminishes with increasing distance. This phenomenon is also observed for our proposed calibration framework, as shown in Section VI.C.}

{Finally, we remark that while the proposed method can tolerate certain noises such as the robot motion noise and air conditioner noise, it might face challenges in more complex scenarios with diffraction, reflection, and multiple sound sources. In these scenarios, to achieve satisfactory calibration accuracy, one has to incorporate other advanced techniques reported in the literature \cite{AnTRO}, \cite{Valin2007}, \cite[pp. 217-241]{BSS}.}

\section{\label{CONCLUSION}CONCLUSION}
This paper is concerned with the joint calibration of multiple asynchronous microphone arrays and sound source localization via batch SLAM. First of all, using the FIM approach, we have conducted a systematic observability analysis of the batch SLAM framework for the above-mentioned calibration problem. More specifically, we have established necessary/sufficient conditions guaranteeing that the FIM and the Jacobian matrix have full column rank, which further implies the identifiability of the unknown parameters. Several scenarios where the unknown parameters are not uniquely identifiable have also been discovered and discussed. Subsequently, for solving the corresponding NLS problem, an effective framework has been proposed to obtain initialized values for the unknown parameters, which are used as the initial guesses in Gauss–Newton types of iterations in batch SLAM and further improve optimization accuracy and convergence. 
{Extensive Monte Carlo simulations and real experiments confirm that the proposed method exhibits high efficiency, accuracy, and robustness in parameter calibration in 3D cases, outperforming the state-of-the-art frameworks for multiple microphone arrays calibration.}

The main focus of our current and future work is to consider the active calibration problem of single or multiple microphone arrays where the sound source can optimize its trajectory in real-time to actively collect measurements that contain richer information for improved accuracy and performance, in contrast to the scenarios where the sound source is operated by a human. 
The calibration problem of moving microphone arrays is also of interest in our future work.

\section{ACKNOWLEDGMENT}
The authors would like to thank the reviewers and Editors for their constructive suggestions which have helped
to improve the quality and presentation of this paper significantly. This work was supported by the Science, Technology, and Innovation Commission of Shenzhen Municipality, China, under Grant No. ZDSYS20220330161800001, the Shenzhen Science and
Technology Program under Grant No. KQTD20221101093557010, the National Natural Science Foundation of China (NSFC) under Grant No. 62350055.

\begin{appendices}
\section*{Appendix A}\label{secA1}
\begin{proof of Proposition 1} Firstly, we note that the relative position of the sound source satisfies $\mathbf{s}_{\Delta}^{k-1}=\mathbf{s}^{k}-\mathbf{s}^{k-1}+\mathbf{w}^{k-1}$ whose corresponding Jacobian matrices are 
\[
\dfrac{\partial\mathbf{s}_{\Delta}^{k-1}}{{\partial}\mathbf{s}^{k-1}}=-\mathbf{I}_{3},\text{ }\dfrac{\partial\mathbf{s}_{\Delta}^{k-1}}{{\partial}\mathbf{s}^{k}}=\mathbf{I}_{3}.
\]

Secondly, for $i=2,...,N$, the distance between the $i\raisebox{0mm}{-}th$
microphone array and the sound source at time instance $t^{k}$ can
be computed as 
\begin{equation}
d_{i}^{k}=\sqrt{{({\Delta x}_{i}^{k})}^{2}+{({\Delta y}_{i}^{k})}^{2}+{({\Delta z}_{i}^{k})}^{2}}\label{dist}
\end{equation}
where 
\begin{equation}
\begin{array}{c}
{\Delta x}_{i}^{k}=s_{x}^{k}-x_{arr\_i}^{x} \text{, }
{\Delta}y_{i}^{k}=s_{y}^{k}-x_{arr\_i}^{y} \text{, }
{\Delta}z_{i}^{k}=s_{z}^{k}-x_{arr\_i}^{z}.
\end{array}\label{delta}
\end{equation}
When $i=1,$ i.e., for the first microphone array, we have 
\begin{equation}
d_{1}^{k}=\sqrt{{(s_{x}^{k})}^{2}+{(s_{y}^{k})}^{2}+{(s_{z}^{k})}^{2}}.\label{eq:s}
\end{equation}
Based on the DOA and TDOA models in (\ref{expression_DOA}) and (\ref{eq:TDOA}), then
\begin{equation}
\mathbf{L}^{k}=\dfrac{\partial\mathbf{y}^{k}(\mathbf{x}_{arr},\mathbf{s}^{k})}{{\partial}\mathbf{x}_{arr}}=\left[\begin{array}{ccc}
\mathbf{J}_{arr\_2}^{k},\cdots,\mathbf{J}_{arr\_N}^{k}\end{array}\right]\in\mathbb{R}^{4(N-1)\times8(N-1)}
\end{equation}
where for $i=2,...,N,$ and $k=1,\ldots,K$, and only entries of $\mathbf{J}_{arr\_i}^{k}$
on its $(4i-7:4i-4)$ rows are nonzero.  Then, $\mathbf{L}^{k}$ can be re-expressed as: 
\begin{equation}
\mathbf{L}^{k}=diag(\mathbf{H}_{arr\_2}^{k},\mathbf{H}_{arr\_3}^{k},\cdots,\mathbf{H}_{arr\_N}^{k}).\label{eq:L}
\end{equation}
Denote{\small{}{} }$\mathbf{h}_{i}^{k},\mathbf{U}_{i}^{k}$
as the partial derivative of TDOA and DOA w.r.t. microphone array positions, respectively; denote $\mathbf{V}_{i}^{k}$ as the partial
derivative of DOA w.r.t. XYZ Euler angles. We then have: 
\begin{equation}
\begin{array}{c}
\mathbf{H}_{arr\_i}^{k}\triangleq\mathbf{J}_{arr\_i}^{k}(4i-7:4i-4,:)\\
=\left[\begin{array}{cccc}
\mathbf{h}_{i}^{k} & \mathbf{0} & 1 & {{\Delta}_{k}}\\
\mathbf{U}_{i}^{k} & \mathbf{V}_{i}^{k} & \mathbf{0} & \mathbf{0}
\end{array}\right]\in\mathbf{\mathbb{R}}^{4\times8}
\end{array}\label{eq:Block of neccessary matrix}
\end{equation}
where 
\[
\mathbf{h}_{i}^{k}=\text{\ensuremath{\left[\dfrac{{\scriptstyle {\displaystyle -{\Delta x}_{i}^{k}}}}{cd_{i}^{k}},\dfrac{{\scriptstyle {\displaystyle -{\Delta y}_{i}^{k}}}}{cd_{i}^{k}},\dfrac{{\scriptstyle {\displaystyle -{\Delta z}_{i}^{k}}}}{cd_{i}^{k}}\right]}},
\]
\begin{equation}
\begin{array}{c}
\mathbf{U}_{i}^{k}=-\mathbf{R}_{i}^{\mathrm{T}}\mathbf{A}\\
=-\mathbf{R}_{i}^{\mathrm{T}}\left[\begin{array}{ccc}
\dfrac{{\scriptstyle (\Delta y_{i}^{k})^{2}+(\Delta z_{i}^{k})^{2}}}{{\scriptstyle (d_{i}^{k})^{3}}} & \dfrac{{\scriptstyle -\Delta x_{i}^{k}\Delta y_{i}^{k}}}{{\scriptstyle (d_{i}^{k})^{3}}} & \dfrac{{\scriptstyle -\Delta x_{i}^{k}\Delta z_{i}^{k}}}{{\scriptstyle (d_{i}^{k})^{3}}}\\
\dfrac{{\scriptstyle -\Delta x_{i}^{k}\Delta y_{i}^{k}}}{{\scriptstyle (d_{i}^{k})^{3}}} & \dfrac{{\scriptstyle (\Delta x_{i}^{k})^{2}+(\Delta z_{i}^{k})^{2}}}{{\scriptstyle (d_{i}^{k})^{3}}} & \dfrac{{\scriptstyle -\Delta y_{i}^{k}\Delta z_{i}^{k}}}{{\scriptstyle (d_{i}^{k})^{3}}}\\
\dfrac{{\scriptstyle -\Delta x_{i}^{k}\Delta z_{i}^{k}}}{{\scriptstyle (d_{i}^{k})^{3}}} & \dfrac{{\scriptstyle -\Delta y_{i}^{k}\Delta z_{i}^{k}}}{{\scriptstyle (d_{i}^{k})^{3}}} & \dfrac{{\scriptstyle (\Delta x_{i}^{k})^{2}+(\Delta y_{i}^{k})^{2}}}{{\scriptstyle (d_{i}^{k})^{3}}}
\end{array}\right],
\end{array}\label{eq:U formation}
\end{equation}
and 
\begin{equation}
\mathbf{V}_{i}^{k}={\scriptstyle \dfrac{1}{{\scriptstyle {\displaystyle d_{i}^{k}}}}}\left[\begin{array}{c}
\left[{\scriptstyle \left(\dfrac{{\scriptstyle \partial\mathbf{R}_{i\_x}^{\mathrm{T}}}}{{\scriptstyle \partial\theta_{x}}}\right)\mathbf{R}_{i\_y}^{\mathrm{T}}\mathbf{R}_{i\_z}^{\mathrm{T}}\left(\begin{array}{c}
{\Delta x}_{i}^{k}\\
{\Delta y}_{i}^{k}\\
{\Delta z}_{i}^{k}
\end{array}\right)}\right]^{\mathrm{T}}\\
\left[{\scriptstyle {\scriptstyle \mathbf{R}_{i\_x}^{\mathrm{T}}}\left(\dfrac{{\scriptstyle \partial\mathbf{R}_{i\_y}^{\mathrm{T}}}}{{\scriptstyle {\scriptstyle \partial\theta_{y}}}}\right){\scriptstyle \mathbf{R}_{i\_z}^{\mathrm{T}}}{\scriptstyle \left(\begin{array}{c}
{\Delta x}_{i}^{k}\\
{\Delta y}_{i}^{k}\\
{\Delta z}_{i}^{k}
\end{array}\right)}}\right]^{\mathrm{T}}\\
\left[{\scriptstyle \mathbf{R}_{i\_x}^{\mathrm{T}}{\scriptstyle \mathbf{R}_{i\_y}^{\mathrm{T}}\left(\dfrac{{\scriptstyle \partial\mathbf{R}_{i\_z}^{\mathrm{T}}}}{{\scriptstyle \partial\theta_{z}}}\right)}}{\scriptstyle \left(\begin{array}{c}
{\Delta x}_{i}^{k}\\
{\Delta y}_{i}^{k}\\
{\Delta z}_{i}^{k}
\end{array}\right)}\right]^{\mathrm{T}}
\end{array}\right]^{\mathrm{T}}\label{eq:V formation}
\end{equation}
where $\mathbf{R}_{i\_x},\mathbf{R}_{i\_y}$ and $\mathbf{R}_{i\_z}$
are the rotation matrices about coordinate frame axes $x,\text{ }y$, and
$z$, respectively. $\mathbf{R}_{i}^{\mathrm{T}}$
can be expressed as $\mathbf{R}_{i}^{\mathrm{T}}=\mathbf{R}_{i\_x}^{\mathrm{T}}\mathbf{R}_{i\_y}^{\mathrm{T}}\mathbf{R}_{i\_z}^{\mathrm{T}},$
with
\[
\begin{array}{c}
\mathbf{R}_{i\_x}=\left[\begin{array}{ccc}
1 & 0 & 0\\
0 & \cos\theta_{x} & -\sin\theta_{x}\\
0 & \sin\theta_{x} & \cos\theta_{x}
\end{array}\right]\\
\mathbf{R}_{i\_y}=\left[\begin{array}{ccc}
\cos\theta_{y} & 0 & \sin\theta_{y}\\
0 & 1 & 0\\
-\sin\theta_{y} & 0 & \cos\theta_{y}
\end{array}\right]\\
\mathbf{R}_{i\_z}=\left[\begin{array}{ccc}
\cos\theta_{z} & -\sin\theta_{z} & 0\\
\sin\theta_{z} & \cos\theta_{z} & 0\\
0 & 0 & 1
\end{array}\right]
\end{array}.
\]
Denote $\mathbf{T}^{k}=\dfrac{\partial\mathbf{y}^{k}(\mathbf{x}_{arr},\mathbf{s}^{k})}{\partial\mathbf{s}^{k}}\in\mathbf{\mathbb{R}}^{4(N-1)\times3}$
as the partial derivative of TDOA and DOA observations w.r.t. sound
source position at time instance $t^{k}$, for $k=1,\ldots,K$. We
then have the expression of $\mathbf{T}^{k}$ as follows: 
\begin{equation}
\begin{array}{c}
\mathbf{T}^{k}=\dfrac{\partial\mathbf{y}^{k}(\mathbf{x}_{arr},\mathbf{s}^{k})}{{\partial}\mathbf{s}^{k}}=\left[\begin{array}{ccc}
\mathbf{J}_{x}^{k} & \mathbf{J}_{y}^{k} & \mathbf{J}_{z}^{k}\end{array}\right]\\
=\left[\begin{array}{c}
-\mathbf{h}_{2}^{k}\\
\mathbf{-U}_{2}^{k}\\
\vdots\\
\mathbf{-h}_{N}^{k}\\
\mathbf{-U}_{N}^{k}
\end{array}\right]-\left[\begin{array}{c}
\left(\dfrac{\mathbf{s}^{k}}{cd_{1}^{k}}\right)^{\mathrm{T}}\\
\mathbf{0}_{3\times3}\\
\vdots\\
\left(\dfrac{\mathbf{s}^{k}}{cd_{1}^{k}}\right)^{\mathrm{T}}\\
\mathbf{0}_{3\times3}
\end{array}\right]
\end{array}.\label{eq:part TK}
\end{equation}
The results then follow the definition of the Jacobian matrix \cite[pp. 569]{Siciliano2009}.
This completes the proof. \end{proof of Proposition 1}

\begin{proof of Theorem 2} 
\noindent By performing elementary row transformation of $\mathbf{F}$,
we can obtain:

\begin{equation}
\begin{array}{c}
\overline{\mathbf{F}}=\left[\begin{array}{ccccc}
\mathbf{H}_{arr\_2}^{1} &  &  &  & \mathbf{T}_{arr\_2}^{1}\\
\vdots &  &  &  & \vdots\\
\mathbf{H}_{arr\_2}^{K} &  &  &  & \mathbf{T}_{arr\_2}^{K}\\
 & \mathbf{H}_{arr\_3}^{1} &  &  & \mathbf{T}_{arr\_3}^{1}\\
 & \vdots &  &  & \vdots\\
 & \mathbf{H}_{arr\_3}^{K} &  &  & \mathbf{T}_{arr\_3}^{K}\\
 &  & \ddots &  & \vdots\\
 &  &  & \mathbf{H}_{arr\_N}^{1} & \mathbf{T}_{arr\_N}^{1}\\
 &  &  & \vdots & \vdots\\
 &  &  & \mathbf{H}_{arr\_N}^{K} & \mathbf{T}_{arr\_N}^{K}
\end{array}\right]\\
=\underset{\overline{\mathbf{L}}}{\underbrace{\left[\begin{array}{cccc}
    \mathbf{H}_{arr\_2}\\
     & \mathbf{H}_{arr\_3}\\
     &  & \ddots\\
     &  &  & \mathbf{H}_{arr\_N}
    \end{array}\right.}}\underset{\overline{\mathbf{T}}}{\underbrace{\left.\begin{array}{c}
    \mathbf{T}_{arr\_2}\\
    \mathbf{T}_{arr\_3}\\
    \vdots\\
    \mathbf{T}_{arr\_N}
    \end{array}\right]}}
\end{array}\label{eq:L T}
\end{equation}
where 
\[
\begin{array}{c}
\mathbf{H}_{arr\_i}=\left[\begin{array}{ccc}
\mathbf{H}_{arr\_i}^{1};\cdots;\mathbf{H}_{arr\_i}^{K}\end{array}\right]\in\mathbf{\mathbb{R}}^{4K\times8}\\
\mathbf{T}_{arr\_i}=\left[\begin{array}{ccc}
\mathbf{T}_{arr\_i}^{1};\cdots;\mathbf{T}_{arr\_i}^{K}\end{array}\right]\in\mathbf{\mathbb{R}}^{4K\times3}
\end{array}
\]
for $i=2,...,N$. Apparently, it holds that $rank(\mathbf{F})=rank(\overline{\mathbf{F}})$.
Also, due to the structure of $\mathbf{H}_{arr\_i}$, their columns
are independent of each other. For each microphone array, denote $\mathbf{F}_{arr\_i}=\left[\begin{array}{cc}
\mathbf{H}_{arr\_i} & \mathbf{T}_{arr\_i}\end{array}\right]$. We then perform the following elementary transformation on the matrix
$\mathbf{F}_{arr\_i}$:

(i) adding the first column block $\left[\mathbf{h}_{i}^{1};\mathbf{U}_{i}^{1};\cdots;\mathbf{h}_{i}^{K};\mathbf{U}_{i}^{K}\right]$
of $\mathbf{H}_{arr\_i}$ to $\mathbf{T}_{arr\_i}$;

(ii) exchanging row blocks to collect all $\mathbf{h}_{i}^{k}$ and
$\mathbf{U}_{i}^{k}$ together, respectively, thereby obtaining 
\begin{equation}
\overline{\mathbf{F}}_{arr\_i}={\left[\begin{array}{cccc}
{\scriptstyle \mathbf{M}_{h\_i}} & {\scriptstyle \mathbf{0}} & {\scriptstyle \mathbf{1}_{K\times1}} & {\scriptstyle \varphi_{\mathbf{k}}}\\
{\scriptstyle \mathbf{M}_{U\_i}} & {\scriptstyle \mathbf{M}_{V\_i}} & {\scriptstyle \mathbf{0}} & {\scriptstyle \mathbf{0}}
\end{array}\right.}{\left.\begin{array}{c}
{\scriptstyle -\mathbf{t}_{\mathbf{k}}}\\
{\scriptstyle \mathbf{0}}
\end{array}\right]}\in\mathbf{\mathbb{R}}^{4K\times11}\label{eq:L';T'}
\end{equation}
where 
\[
\begin{array}{c}
\mathbf{M}_{h\_i}=[\mathbf{h}_{i}^{1};\mathbf{h}_{i}^{2};\ldots;\mathbf{h}_{i}^{K}], \text{ } 
\mathbf{M}_{U\_i}=\left[\mathbf{U}_{i}^{1};\mathbf{U}_{i}^{2};\ldots;\mathbf{U}_{i}^{K}\right],\\
\mathbf{M}_{V\_i}=\left[\mathbf{V}_{i}^{1};\mathbf{V}_{i}^{2};\ldots;\mathbf{V}_{i}^{K}\right]\text{, } 
{{\varphi}_{\mathbf{k}}=\left[\begin{array}{c}
    \Delta_{1};\Delta_{2};\ldots;\Delta_{K}\end{array}\right],}\\
\mathbf{t}_{\mathbf{k}}=\left[\begin{array}{c}
\left(\frac{{\scriptstyle \mathbf{s}^{1}}}{{\scriptstyle cd_{1}^{1}}}\right)^{\mathrm{T}};\left(\frac{{\scriptstyle \mathbf{s}^{2}}}{{\scriptstyle cd_{1}^{2}}}\right)^{\mathrm{T}};\ldots;\left(\frac{{\scriptstyle \mathbf{s}^{K}}}{{\scriptstyle cd_{1}^{K}}}\right)^{\mathrm{T}}\end{array}\right].
\end{array}
\]
We further perform the following elementary operations on $\overline{\mathbf{F}}_{arr\_i}$,
$i=2,3,\cdots,N$:

(i) dividing the fourth column block by ${\Delta}_{1}$;

(ii) for $k=2,3,\cdots,K$, deducing the $k\raisebox{0mm}{-}th$ row
by the first row;

(iii) transforming the elements in the first row (except the third
one) to zero by the third column block (the first element therein
equals 1 while the other elements equal zero after the elementary
operations listed above);

(iv) for $k=3,4,\cdots,K$, deducing the $k\raisebox{0mm}{-}th$ row
by the second row multiplied by $\frac{{\Delta}_{k}-{\Delta}_{1}}{{\Delta}_{2}-{\Delta}_{1}}$;

(v) transforming the elements in the second row (except the fourth
one) to zero by the fourth column block (the second element therein
equals 1 while the other elements equal zero after the elementary
operations listed above);

(vi) moving column blocks 3 and 4 to columns blocks 1 and 2, respectively.

After the above operations, we obtain 
\begin{equation}
\begin{array}{l}
\overline{\mathbf{F}}_{arr\_i}^{\prime}=\left[\begin{array}{cc}
\mathbf{\bar{L}}_{i} & \bar{\mathbf{T}}\end{array}\right]\end{array}\label{eq:F'_arr_i}
\end{equation}
where
\begin{equation}
    {
\bar{\mathbf{T}}=\left[\begin{array}{c}
\mathbf{0}\\
\Psi\\
\mathbf{0}
\end{array}\right]=\left[\begin{array}{c}
\mathbf{0_{\mathrm{2\times3}}}\\
{\scriptstyle \Theta_{1,3}\left(\dfrac{{\scriptstyle \left({\scriptstyle \mathbf{s}^{k}}\right)^{\mathrm{T}}}}{{\scriptstyle cd_{1}^{k}}}\right)-\dfrac{{\scriptstyle \Theta_{3,1}(\Delta_{k})}}{{\scriptstyle \Theta_{2,1}(\Delta_{k})}}\Theta_{1,2}\left(\dfrac{{\scriptstyle \left({\scriptstyle \mathbf{s}^{k}}\right)^{\mathrm{T}}}}{{\scriptstyle cd_{1}^{k}}}\right)}\\
{\scriptstyle {\scriptstyle \Theta_{1,4}\left(\dfrac{{\scriptstyle \left({\scriptstyle \mathbf{s}^{k}}\right)^{\mathrm{T}}}}{{\scriptstyle cd_{1}^{k}}}\right)-\dfrac{{\scriptstyle \Theta_{4,1}(\Delta_{k})}}{{\scriptstyle \Theta_{2,1}(\Delta_{k})}}\Theta_{1,2}\left(\dfrac{{\scriptstyle \left({\scriptstyle \mathbf{s}^{k}}\right)^{\mathrm{T}}}}{{\scriptstyle cd_{1}^{k}}}\right)}}\\
\vdots\\
{\scriptstyle {\scriptstyle \Theta_{1,K}\left(\dfrac{{\scriptstyle \left({\scriptstyle \mathbf{s}^{k}}\right)^{\mathrm{T}}}}{{\scriptstyle cd_{1}^{k}}}\right)-\dfrac{{\scriptstyle \Theta_{K,1}(\Delta_{k})}}{{\scriptstyle \Theta_{2,1}(\Delta_{k})}}}\Theta_{1,2}\left(\dfrac{{\scriptstyle \left({\scriptstyle \mathbf{s}^{k}}\right)^{\mathrm{T}}}}{{\scriptstyle cd_{1}^{k}}}\right)}\\
\mathbf{0_{\mathrm{\mathit{3K}\times3}}}
\end{array}\right]\label{eq:T-BAR}}
    \end{equation}{and}    
    \begin{equation}
\begin{aligned}\mathbf{\bar{L}}_{i}= & diag(\mathbf{I}_{2},\Phi_i)\\
= & \left[\begin{array}{cccc}
1 & 0 & \mathbf{0} & \mathbf{0}\\
0 & 1 & \mathbf{0} & \mathbf{0}\\
0 & 0 & {\scriptstyle \Theta_{3,1}(\mathbf{h}_{arr\_i}^{k})-\dfrac{{\scriptstyle \Theta_{3,1}(\Delta_{k})}}{{\scriptstyle \Theta_{2,1}(\Delta_{k})}}\Theta_{2,1}(\mathbf{h}_{arr\_i}^{k})} & \mathbf{0}\\
0 & 0 & {\scriptstyle \Theta_{4,1}(\mathbf{h}_{arr\_i}^{k})-\dfrac{{\scriptstyle \Theta_{4,1}(\Delta_{k})}}{{\scriptstyle \Theta_{2,1}(\Delta_{k})}}\Theta_{2,1}(\mathbf{h}_{arr\_i}^{k})} & \mathbf{0}\\
\vdots & \vdots & \vdots & \vdots\\
0 & 0 & {\scriptstyle \Theta_{K,1}(\mathbf{h}_{arr\_i}^{k})-\dfrac{{\scriptstyle \Theta_{K,1}(\Delta_{k})}}{{\scriptstyle \Theta_{2,1}(\Delta_{k})}}\Theta_{2,1}(\mathbf{h}_{arr\_i}^{k})} & \mathbf{0}\\
\mathbf{0} & \mathbf{0} & \mathbf{U}_{arr\_i}^{1} & \mathbf{V}_{arr\_i}^{1}\\
\mathbf{0} & \mathbf{0} & \mathbf{U}_{arr\_i}^{2} & \mathbf{V}_{arr\_i}^{2}\\
\vdots & \vdots & \vdots & \vdots\\
\mathbf{0} & \mathbf{0} & \mathbf{U}_{arr\_i}^{K} & \mathbf{V}_{arr\_i}^{K}
\end{array}\right]
\end{aligned}
\label{eq: L-BAR}
    \end{equation}
    {with $\mathbf{h}$, $\mathbf{U}$, and $\mathbf{V}$ being defined
    in (\ref{eq:Block of neccessary matrix}), $\Theta_{m,n}(\boldsymbol{f}(k))$
    represents $\boldsymbol{f}(m)-\boldsymbol{f}(n).$} With the above elementary row and column transformations, we have
\begin{equation}
\overline{\mathbf{F}}\sim\overline{\mathbf{F}}^{\prime}=\underset{\mathbf{\overline{L}^{\prime}}}{\underbrace{\left[\begin{array}{cccc}
    \mathbf{\bar{L}}_{2}\\
     & \mathbf{\bar{L}}_{3}\\
     &  & \ddots\\
     &  &  & \mathbf{\bar{L}}_{N}
    \end{array}\right.}}\underset{\mathbf{\overline{T}^{\prime}}}{\underbrace{\left.\begin{array}{c}
    \mathbf{\bar{T}}\\
    \mathbf{\bar{T}}\\
    \vdots\\
    \mathbf{\bar{T}}
    \end{array}\right]}}.\label{eq:F bar prime}
\end{equation}
It holds that $rank(\mathbf{F})=rank(\overline{\mathbf{F}})=rank(\overline{\mathbf{F}}^{\prime})$.
From the structure of $\overline{\mathbf{F}}^{\prime}$, we can see
that the block columns containing $\mathbf{\bar{L}}_{i}$,
$i=2,...,N$, are independent of each other. A necessary condition
for $\overline{\mathbf{F}}^{\prime}$ to be of full column rank is
that $\mathbf{\bar{L}}_{i}$ and $\mathbf{\bar{T}}$ are
of full column rank, respectively, $i=2,...,N$. This completes the
proof.\end{proof of Theorem 2}

\begin{proof of Theorem 3} Here we take $j=2$ as an example. For
$\overline{\mathbf{F}}^{\prime}$, we could perform elementary row
block changes: for $i=3,\ldots,N$, deduce $\mathbf{\bar{L}}_{i}$
row block by the first-row block and obtain: 
\begin{equation}
\left[\begin{array}{cccccc}
\mathbf{\bar{L}}_{2} &  &  &  &  & \mathbf{\bar{T}}\\
-\mathbf{\bar{L}}_{2} & \mathbf{\bar{L}}_{3} &  &  &  & \mathbf{0}\\
\vdots &  & \ddots &  &  & \mathbf{\vdots}\\
-\mathbf{\bar{L}}_{2} &  &  &  & \mathbf{\bar{L}}_{N} & \mathbf{0}
\end{array}\right].\label{eq:L_sufficient}
\end{equation}

\noindent Denote the submatrix of this matrix as: 
\begin{equation}
\mathbf{M}_{2\_T}=\left[\begin{array}{cc}
\mathbf{\bar{L}}_{2} & \mathbf{\bar{T}}\\
\mathbf{\vdots} & \mathbf{\vdots}\\
-\mathbf{\bar{L}}_{2} & \mathbf{0}
\end{array}\right].\label{eq:M2_t}
\end{equation}

\noindent From the structure in (\ref{eq:L_sufficient}), we can see
clearly that if:

(i) $\mathbf{M}_{2\_T}$ is of full column rank, and

(ii) $diag(\mathbf{\bar{L}}_{3},\ldots,\mathbf{\bar{L}}_{N})$\ is
of full column rank,\\
 then $\overline{\mathbf{F}}^{\prime}$ will be of full column rank.
Due to the fact that $rank(\mathbf{F})=rank(\overline{\mathbf{F}})=rank(\overline{\mathbf{F}}^{\prime})$,
the Jacobian matrix $\mathbf{J}$ is of full column rank. Similarly,
the same conditions hold when $j$ equals to $3,\ldots,N$. So the
Jacobian matrix $\mathbf{J}$ is of full column rank if any matrix
consisting of the $(j-1)\raisebox{0mm}{-}th$ column block and the
last column block in $\overline{\mathbf{F}}^{\prime}$ is of full
column rank, $2\leq j\leq N$, and $\mathbf{\bar{L}}_{i}$
are of full column rank, $i=2,\ldots,N$ and $i\neq j$. This completes
the proof. \end{proof of Theorem 3}

\begin{proof of Theorem 4} (i) $\bar{\mathbf{T}}$ in (\ref{eq:T-BAR})
is of full column rank only if a 3 \texttimes{} 3 matrix formed by
at least one of the three-permutation of its rows is full rank. For
$\left(\mathbf{s}^{k}\right)^{\mathrm{T}}\in \mathbb{R}^{1\times3},1\leq k\leq K$,
the necessary condition for $\bar{\mathbf{T}}$ to be of full column
rank is $K\geq5$. If $K<5$, $\bar{\mathbf{T}}$ can not be of the
full column rank.

(ii) Based on (\ref{eq:s}), when $\mathbf{\mathbf{s}}^{k}={\lambda}_{k-1}\mathbf{s}^{k-1}$,
we could derive $\frac{\mathbf{s}^{k}}{d_{1}^{k}}=\frac{\mathbf{s}^{k-1}}{d_{1}^{k-1}}.$ From the expression of $\bar{\mathbf{T}}$, we can see that $\bar{\mathbf{T}}$
cannot be of full rank if $\mathbf{s}^{k}$ is proportional to each
other, $k=1,\cdots,K$. In this case, the sound source is collinear with the origin of the reference microphone array frame $\left\{ \mathrm{\mathbf{x}}_{arr\_1}\right\} $ at all time steps.

(iii) If the sound source lies on any Euclidean plane of $x+\alpha y=0$, $x+\beta z=0$, and $y+\gamma z=0$  within the three-dimensional $x-y-z$ Cartesian coordinate frame $\left\{ \mathrm{\mathbf{x}}_{arr\_1}\right\} $ at all moments, where $\alpha$, $\beta$, and $\gamma$ are arbitrary
scalars, the sound source position $\mathbf{s}^{k},$ $1\leq k\leq K$,
could be expressed as $\left[-\alpha s_{y}^{k};s_{y}^{k};s_{z}^{k}\right]$,
$\left[-\beta s_{z}^{k};s_{y}^{k};s_{z}^{k}\right]$, and $\left[s_{x}^{k};-\gamma s_{z}^{k};s_{z}^{k}\right]$,
respectively. $\bar{\mathbf{T}}$ will not be of full column rank. Specifically, if $\alpha=0$ or $\beta=0$ or $\gamma=0$, the sound
source position $\mathbf{s}^{k}$ will have $s_{x}^{k}=0$, $s_{y}^{k}=0$,
and $s_{z}^{k}=0$, respectively, i.e., YOZ, XOZ, and XOY planes in global frame. This completes the proof. \end{proof of Theorem 4}

\begin{proof of Theorem 5} (i) {If the sound source, at all of $K\,(K\geq5)$
time steps, is collinear w.r.t. the origin of the microphone array frame $\left\{ \mathrm{\mathbf{x}}_{arr\_i}\right\} $, i.e., $(\mathbf{\mathbf{s}}^{k}-\mathbf{x}_{arr\_i}^{p})={\epsilon}_{k-1}(\mathbf{\mathbf{s}}^{k-1}-\mathbf{x}_{arr\_i}^{p})$ always 
 holds true,} then for $i\geq2,$ $k=2,3,\ldots,K$, we can get the following expression:
\[
\begin{cases}
{\scriptstyle \left[\begin{array}{ccc}
\Delta x_{i}^{k};\Delta y_{i}^{k};\Delta z_{i}^{k}\end{array}\right]={\epsilon}_{k-1}\left[\begin{array}{ccc}
\Delta x^{k-1}_i;\Delta y^{k-1}_i;\Delta z^{k-1}_i\end{array}\right]}\\
\mathbf{h}_{i}^{k}=\mathbf{h}_{i}^{k-1}, \text{ }\mathbf{U}_{i}^{k}=\frac{1}{{\epsilon}_{k-1}}\mathbf{U}_{i}^{k-1}, \text{ } \mathbf{V}_{i}^{k}=\mathbf{V}_{i}^{k-1}
\end{cases}
\]
where $\mathbf{h},\mathbf{U}$, and $\mathbf{V}$ are defined in (\ref{eq:Block of neccessary matrix}). For an arbitrary single time step, we have $rank(\mathbf{U}_{i}^{k})=rank(\mathbf{R}_{i}^{\mathrm{T}}\mathbf{A})$
as shown in (\ref{eq:U formation}). It can also be seen that $det(\mathbf{A})=0$
and the second-order sub-determinant of $\mathbf{A}$ is not equal
to 0, we know that $rank(\mathbf{A})=2$. $\mathbf{R}_{i}^{\mathrm{T}}$
is a rotation matrix, $rank(\mathbf{R}_{i}^{\mathrm{T}})=3$, thus
$rank(\mathbf{U}_{i}^{k})=2$. Therefore, $\mathbf{\bar{L}}_{i}$
will not be of full column rank.

(ii) When $\theta_{arr\_i}^{y}=\pm \frac{\pi}{2}$, for the corresponding
microphone array at any different time steps, $\mathbf{V}_{i}^{k}$
defined in (\ref{eq:V formation}) has the same structure, i.e., 
\[
\begin{array}{c}
\mathbf{V}_{i}^{k}\text{(\ensuremath{{\scriptstyle \theta_{arr\_i}^{y}=\frac{\pi}{2}}})}=\left[\begin{array}{cc}
{\scriptstyle 0} & {\scriptstyle \Delta x_{i}^{k}c_{z}+\Delta y_{i}^{k}s_{z}}\\
{\scriptstyle \Delta y_{i}^{k}s_{x-z}-\Delta x_{i}^{k}c_{x-z}} & {\scriptstyle \Delta z_{i}^{k}s_{x}}\\
{\scriptstyle \Delta y_{i}^{k}c_{x-z}+\Delta x_{i}^{k}s_{x-z}} & {\scriptstyle \Delta z_{i}^{k}c_{x}}
\end{array}\right.\left.\begin{array}{c}
{\scriptstyle 0}\\
{\scriptstyle -\Delta y_{i}^{k}s_{x-z}+\Delta x_{i}^{k}c_{x-z}}\\
{\scriptstyle -\Delta y_{i}^{k}c_{x-z}-\Delta x_{i}^{k}s_{x-z}}
\end{array}\right]\\
\mathbf{V}_{i}^{k}\text{(\ensuremath{{\scriptstyle \theta_{arr\_i}^{y}=-\frac{\pi}{2}}})}=\left[\begin{array}{cc}
{\scriptstyle 0} & {\scriptstyle -\Delta x_{i}^{k}c_{z}-\Delta y_{i}^{k}s_{z}}\\
{\scriptstyle \Delta y_{i}^{k}s_{x+z}+\Delta x_{i}^{k}c_{x+z}} & {\scriptstyle -\Delta z_{i}^{k}s_{x}}\\
{\scriptstyle \Delta y_{i}^{k}c_{x+z}-\Delta x_{i}^{k}s_{x+z}} & {\scriptstyle -\Delta z_{i}^{k}c_{x}}
\end{array}\right.\left.\begin{array}{c}
{\scriptstyle 0}\\
{\scriptstyle \Delta y_{i}^{k}s_{x+z}+\Delta x_{i}^{k}c_{x+z}}\\
{\scriptstyle \Delta y_{i}^{k}c_{x+z}-\Delta x_{i}^{k}s_{x+z}}
\end{array}\right],
\end{array}
\]
where $s,c$ represent $sin,cos$, respectively and $rank(\mathbf{V}_{i}^{k})\equiv2$.
Therefore, the matrix of $\mathbf{\bar{L}}_{i}$ in (\ref{eq:F bar prime})
will not be of full column rank. This completes the proof. \end{proof of Theorem 5}
\section*{Appendix B}
\label{secB}
\begin{evaluation metrics}
The errors of microphone arrays positions, orientations, time offsets,
clock differences and sound source positions can be expressed as follows:
\[
E(\mathbf{x}_{arr\_i}^{p})=\left\Vert \mathbf{\hat{x}}_{arr\_i}^{p}-\mathbf{x}_{0}\right\Vert _{2},\text{ }E(\mathbf{x}_{arr\_i}^{\theta})=\arccos\left(\frac{\small{\mathbf{\hat{R}}_{i}\mathbf{v}\cdotp\mathbf{X}_{0}\mathbf{v}}}{\left\Vert \mathbf{v}\right\Vert _{2}^{2}}\right),
\]
\[
E(x_{arr\_i}^{\tau})=\hat{x}_{arr\_i}^{\tau}-x_{0},\text{ }E(x_{arr\_i}^{\delta})=\hat{x}_{arr\_i}^{\delta}-x_{0}, \text{ }E(\mathbf{s}_{k})=\left\Vert \hat{\mathbf{s}}_{k}-\mathbf{x}_{0}\right\Vert _{2},
\]
where $\hat{\cdot}$ represents the estimate of the unknown scalars/vectors/matrix
parameters, $x_{0}/\mathbf{x}_{0}/\mathbf{X}_{0}$ represents the
true value of the corresponding parameter and $\mathbf{\mathbf{v}}=\left[1;1;1\right]$.

In the experiments in Sections \ref{simu} and \ref{real}, we utilized
the root mean square error (RMSE) to evaluate the accuracy of the
calibration algorithm for parameters estimations. The RMSE of the parameter
$\mathbf{x}$ was calculated as $RMSE(\mathbf{x})=\sqrt{\frac{1}{M}\sum_{i=1}^{M}E_{i}^{2}(\mathbf{x})}$, where $M$ is equal to the total number of the corresponding parameter $\mathbf{x}$.

\end{evaluation metrics}
\end{appendices}
\newpage
\begin{IEEEbiography}
    [{\includegraphics[width=0.76in,height=0.95in,clip,keepaspectratio]
    {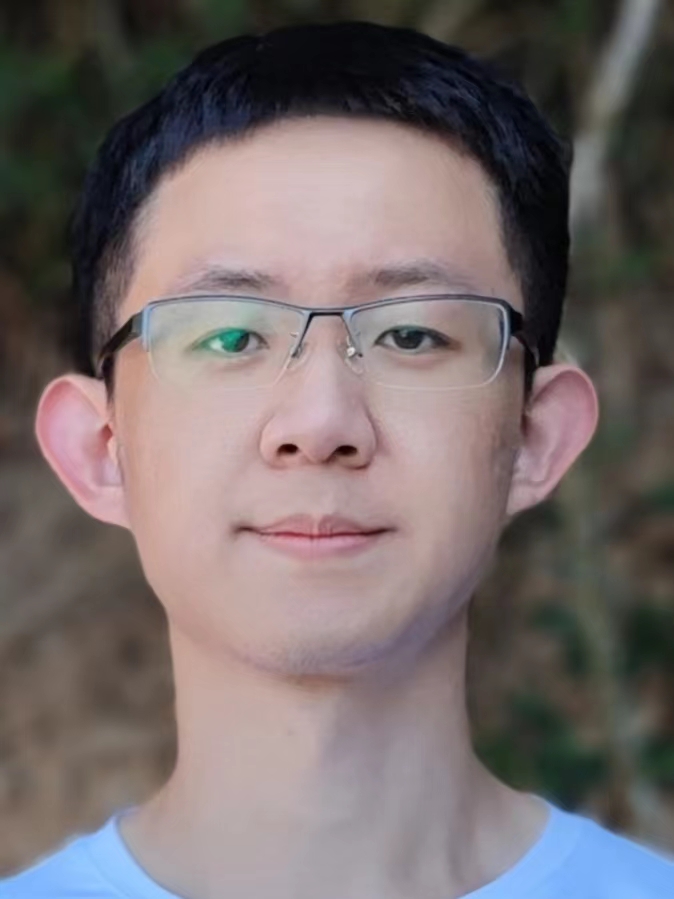}}]{Jiang Wang}
    \footnotesize{received the B.Eng. in Electrical Engineering and Automation from the Shenyang Agricultural University, Shenyang, China, in 2020. Since September 2021, he has been working
towards the M.S. degree in Electronic Science and Technology, Southern University of Science and Technology, Shenzhen, China. His major research interests include sensor calibration, robot audition, SLAM, sensor fusion.}
\end{IEEEbiography}
\begin{IEEEbiography}
    [{\includegraphics[width=0.76in,height=0.95in,clip,keepaspectratio]
    {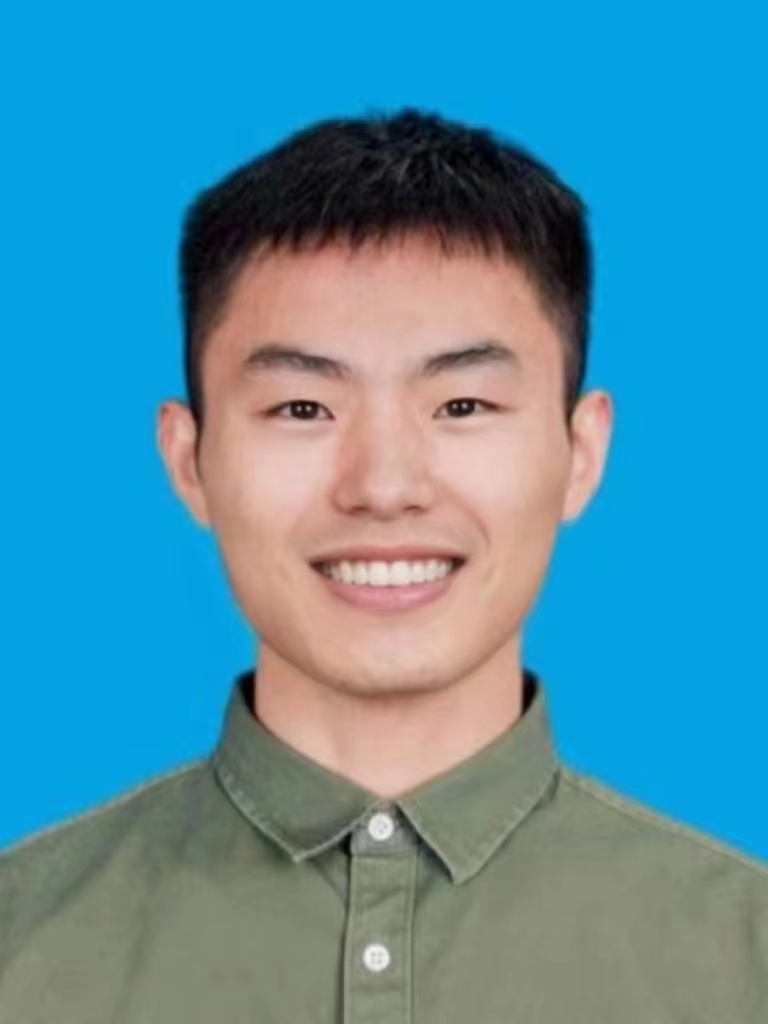}}]{Yuanzheng He}
     \footnotesize{received the B.Eng. in Electronic and Information Engineering from the Henan University of Technology, Zhengzhou, China, in 2021. Since September 2021, he has been working towards the M.S. degree in Electronic Science and Technology, Southern University of Science and Technology, Shenzhen, China. His major research interests include robot audition, robot perception, and multi-sensor fusion.}
\end{IEEEbiography}
\begin{IEEEbiography}
[{\includegraphics[width=0.76in,height=0.95in,clip,keepaspectratio]
{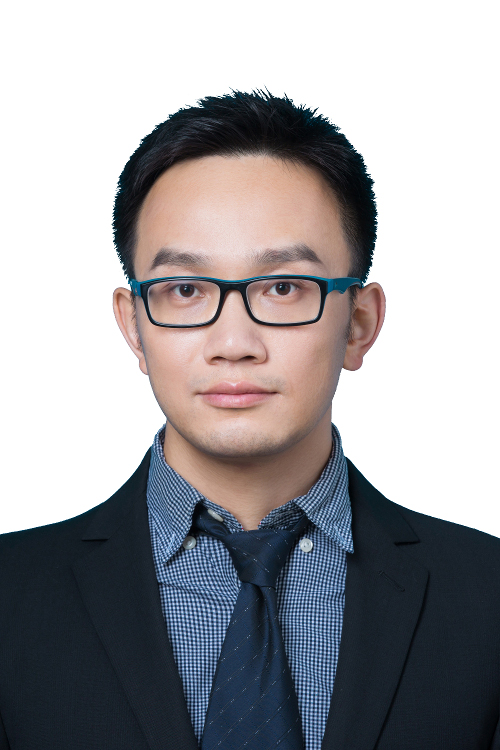}}]{Daobilige Su}
\footnotesize{received his B. Eng. in Mechatronic Engineering from Zhejiang University, China in 2010, M. Eng. in Automation and Robotics from Warsaw University of Technology, Poland and M. Eng. in Automation from University of Genova, Italy through European Master on Advanced Robotics (EMARO) program in 2012, and Ph. D. in robotics at Centre for Autonomous System (CAS), University of Technology Sydney (UTS), Australia in 2017. He was a post-doctoral research associate at Australian Centre for Filed Robotics (ACFR), The University of Sydney from 2017 to 2020. He is currently an Associate Professor at College of Engineering, China Agricultural University, China. His current research areas include field robotics, SLAM, robot audition, computer vision, and machine learning. }
\end{IEEEbiography}
\begin{IEEEbiography}
[{\includegraphics[width=0.76in,height=0.95in,clip,keepaspectratio]
{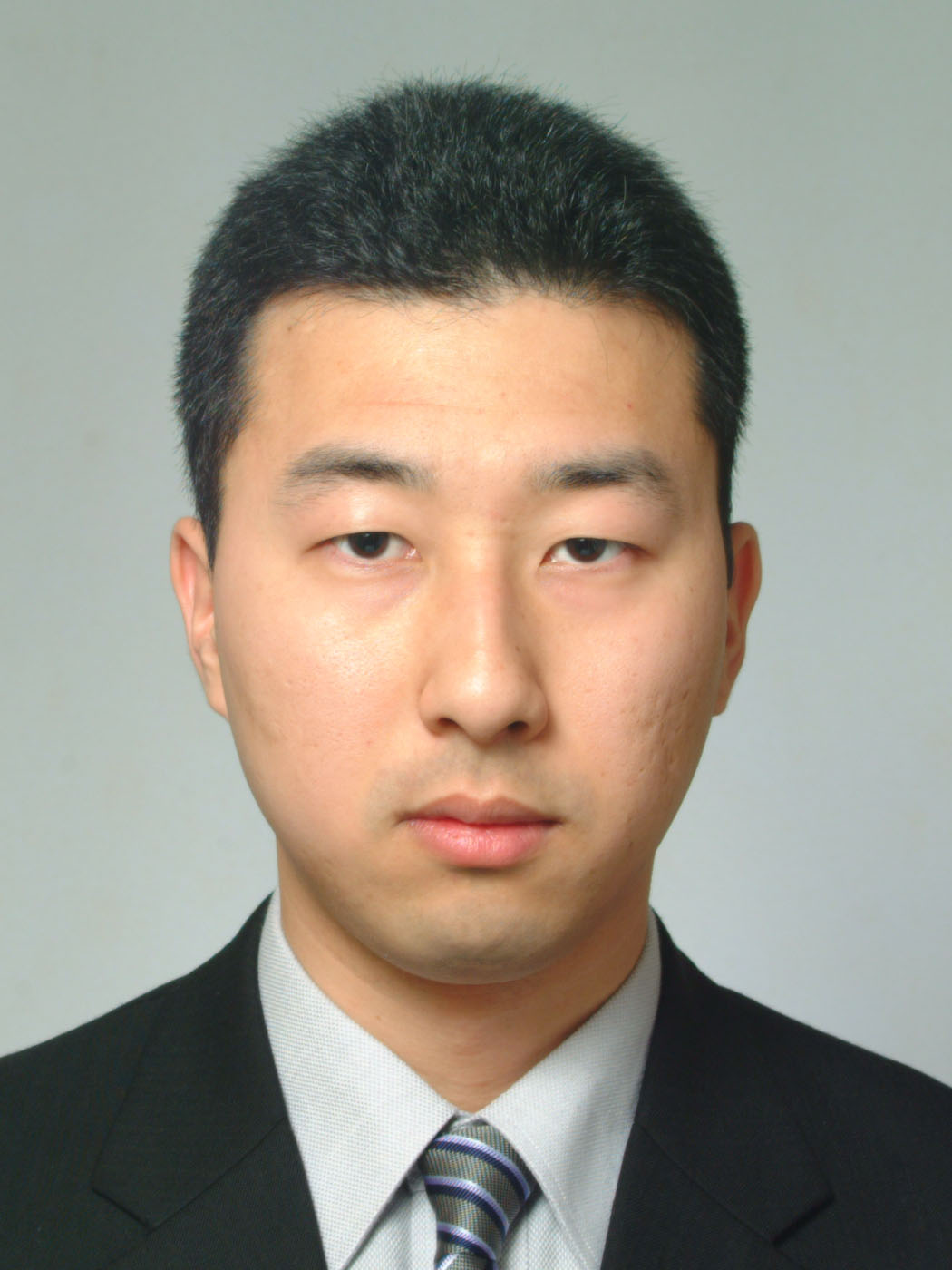}}]{Katsutoshi Itoyama}
\footnotesize{received the M.S. and Ph.D. degrees in informatics from Kyoto University, Kyoto, Japan, in 2008 and 2011, respectively. He had been an Assistant Professor with the Graduate School of Informatics, Kyoto University, until 2018 and is currently a Associate Professor with the Tokyo Institute of Technology, Tokyo, Japan. His research interests include sound source separation, music listening interfaces, and music information retrieval.}
\end{IEEEbiography}
\begin{IEEEbiography}
[{\includegraphics[width=0.76in,height=0.95in,clip,keepaspectratio]{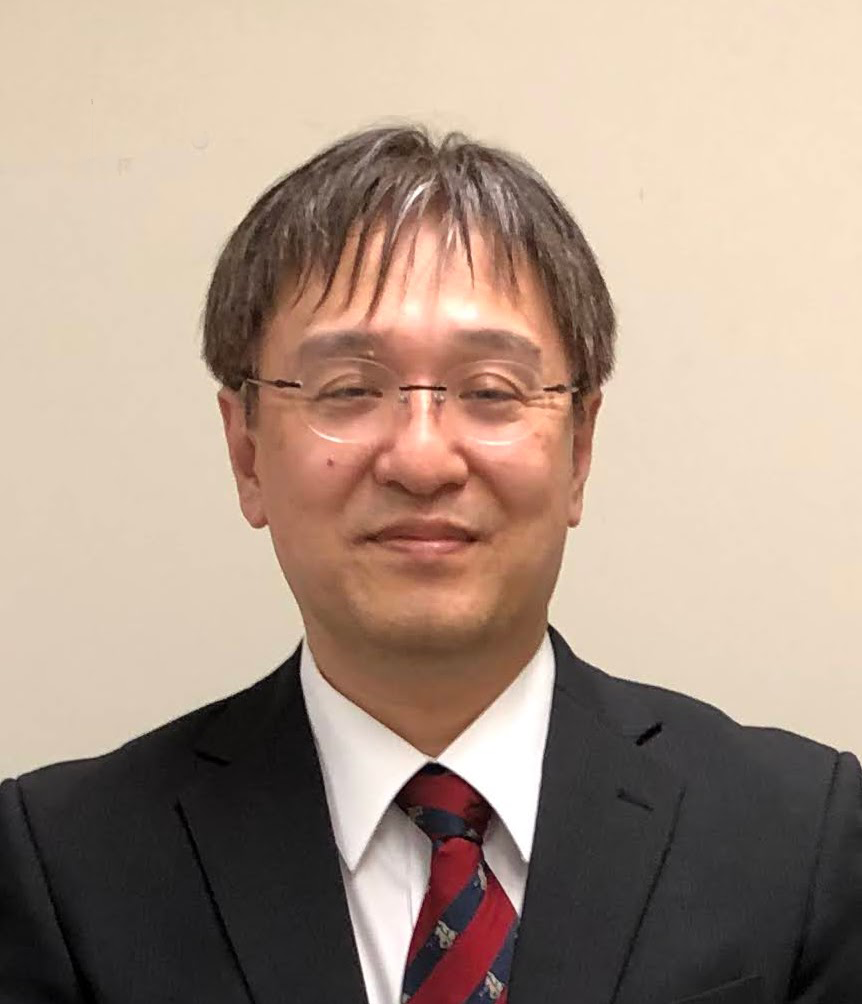}}]{Kazuhiro Nakadai} \footnotesize{received a B.E. in electrical engineering in 1993, an M.E. in information engineering in 1995, and a Ph.D. in electrical engineering in 2003 from the University of Tokyo. He worked 
with Nippon Telegraph and Telephone for four years as a system engineer from 1995 to 1999, with the Kitano Symbiotic Systems Project, ERATO, JST as a researcher from 1999 to 2003, and 
with Honda Research Institute Japan, Co., Ltd. as a principal scientist from 2003 to 2022. Currently he is a professor at the Department of Systems and Control Engineering, School of Engineering, Tokyo Institute of Technology. He has had a concurrent position at Tokyo Institute of Technology, as a visiting associate professor from 2006 to 2010, a visiting professor from 2011 to 2017, and a specially-appointed professor from 2017 to 2022. He also had a concurrent position as a guest professor at Waseda University from 2011 to 2018. His research interests include AI, robotics, signal processing, computational auditory scene analysis, multi-modal integration, and robot audition. He has been an executive board member for JSAI from 2015 to 2016, and for RSJ from 2017 to 2018. He is a Fellow of the IEEE and also a member of JSAI, RSJ, IPSJ, ASJ, HIS, ISCA, ACM.} 
\end{IEEEbiography}
\newpage
\begin{IEEEbiography}
[{\includegraphics[width=0.76in,height=0.95in,clip,keepaspectratio]{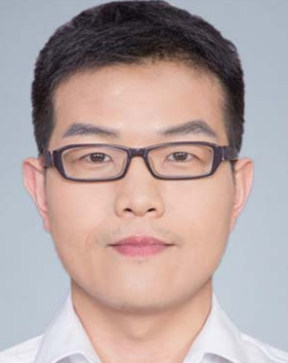}}]{Junfeng Wu}
\footnotesize{received the B.Eng. degree from the Department of Automatic Control, Zhejiang University,
Hangzhou, China, and the Ph.D. degree in electrical and computer engineering from the Hong Kong
University of Science and Technology, Hong Kong,
in 2009, and 2013, respectively.
From 2014 to 2017, he was a Postdoctoral Researcher with the ACCESS (Autonomic
Complex Communication nEtworks, Signals and Systems) Linnaeus Center,
School of Electrical Engineering, KTH Royal Institute of Technology, Stockholm, Sweden. From 2017 to 2021, he was with the College of Control Science
and Engineering, Zhejiang University, Hangzhou, China. He is currently an
Associate Professor with the School of Data Science, The Chinese University of
Hong Kong, Shenzhen, China. His research interests include networked control
systems, state estimation, and wireless sensor networks, multi-agent systems, robot perception and localization. He currently serves as an Associate Editor for \textit{IEEE Transactions on Control of Network Systems}.}
\end{IEEEbiography}
\begin{IEEEbiography}
[{\includegraphics[width=0.76in,height=0.95in,clip,keepaspectratio]{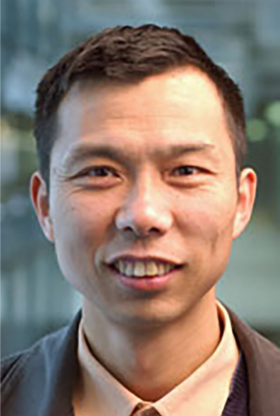}}]{Shoudong Huang}
\footnotesize{received the bachelor’s and master’s degrees in mathematics and the Ph.D. in automatic control from Northeastern University, Shenyang, China, in 1987, 1990, and 1998, respectively. He is currently a Professor with the Centre for Autonomous Systems, Faculty of Engineering and Information Technology, University of Technology, Sydney, NSW, Australia. His research interests include nonlinear control systems and mobile robots 
simultaneous localization and mapping (SLAM), exploration, and navigation.}
\end{IEEEbiography}
\begin{IEEEbiography}
[{\includegraphics[width=0.76in,height=0.95in,clip,keepaspectratio]{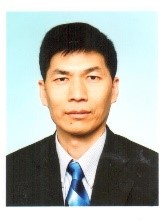}}]{Youfu Li}
\footnotesize{received the PhD degree in robotics from the Department of Engineering Science, University of Oxford in 1993. From 1993 to 1995 he was a research staff in the Department of Computer Science at the University of Wales, Aberystwyth, UK. He joined City University of Hong Kong in 1995 and is currently professor in the Department of Mechanical Engineering. His research interests include robot sensing, robot vision, and visual tracking. In these areas, he has published over 400 papers including over 180 SCI listed journal papers. Dr Li has received many awards in robot sensing and vision including IEEE Sensors Journal Best Paper Award by IEEE Sensors Council, Second Prize of Natural Science Research Award by the Ministry of Education, 1st Prize of Natural Science Research Award of Hubei Province, 1st Prize of Natural Science Research Award of Zhejiang Province, China.  He was on Top 2\% of the world’s most highly cited scientists by Stanford University, 2020, 2021 and Career Long. He has served as an Associate Editor for \textit{IEEE Transactions on Automation Science and Engineering (T-ASE)}, Associate Editor and Guest Editor for \textit{IEEE  Robotics and Automation Magazine (RAM)}, and Editor for CEB, IEEE International Conference on Robotics and Automation (ICRA). He is a Fellow of the IEEE.}
\end{IEEEbiography}
\begin{IEEEbiography}
[{\includegraphics[width=0.76in,height=0.95in,clip,keepaspectratio]{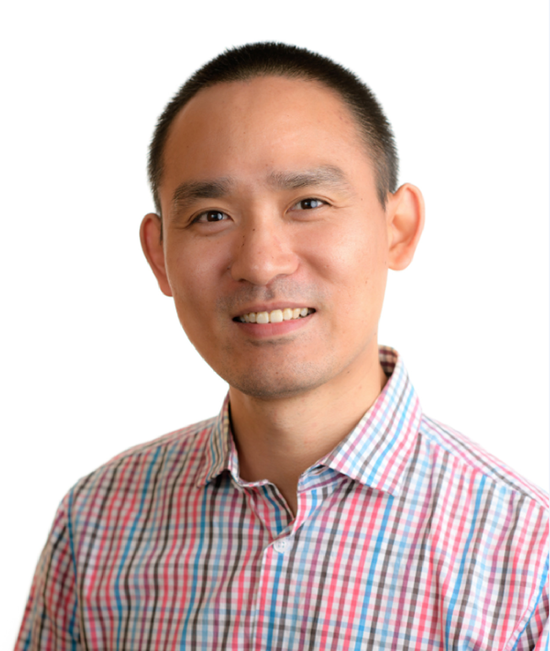}}]{He Kong} \footnotesize{received the Bachelor’s degree in Electrical Engineering from China University of Mining and
Technology, Xuzhou, China, Master’s degree in Control Science and Engineering from Harbin Institute of Technology, Harbin, China, and the Ph.D. degree in Electrical Engineering from the University of Newcastle, Australia, respectively. He was a Research Fellow at the Australian Centre for Field Robotics, the University of Sydney, Australia, during 2016–2021. In early 2022, he joined the Southern University of Science and Technology, Shenzhen, China, where he is currently an Associate Professor. His research interests include active multi-modal perception, robot audition, state estimation, and control applications. He is currently serving on the editorial board of \textit{IEEE Robotics and Automation Letters}, \textit{IEEE Robotics and Automation Magazine}, \textit{IEEE Sensors Letters}, \textit{International Journal of Adaptive Control and Signal Processing}. He has also served as an Associate Editor
for several international conferences in robotics and automation, including the
IEEE ICRA, IEEE/RSJ IROS, IEEE CASE, etc.}
\end{IEEEbiography}

\end{document}